%% file: double.tex
\documentclass[11pt, journal,final,twocolumn]{IEEEtran}

\usepackage{graphicx} 
\usepackage{multirow}
\usepackage{pgf}
\usepackage{color}
\input{rgb}
\usepackage{subfigure}
\usepackage{url}
\usepackage{stfloats}
\usepackage{amssymb}
\usepackage{amsfonts}
\usepackage{amsmath}   
\usepackage{pifont}
\usepackage{algorithmicx}
\usepackage{algorithm}
\usepackage{algpseudocode}
\usepackage{gensymb}
\usepackage{lineno,hyperref}

\begin{document}
%
\title{Shape Registration with Directional Data}
%
%
\author{Mair\'ead Grogan \& Rozenn Dahyot\\
School of Computer Science and Statistics\\
Trinity College Dublin, Ireland\\  
}

\maketitle




\begin{abstract}
We propose  several cost functions for  registration of shapes encoded with  Euclidean and/or non-Euclidean information (unit vectors). Our framework is assessed for estimation of both rigid and non-rigid transformations between the target and model shapes corresponding to 2D contours and 3D surfaces. 
The experimental results obtained confirm that using the combination of a point's position and unit normal vector in a cost function can enhance the registration results compared to state of the art methods. 
\end{abstract}

\section{Introduction}

Directions, axes or rotations are described as unit vectors in $\mathbb{R}^d$ and are known collectively as directional data. In computer vision this type of data is often processed  and includes surface normals and tangent vectors, orientations of image gradients, the direction of sound sources and GPS coordinate information \cite{Dalal2005,IROS2012,MLSP2014}. Directional data can be viewed as points on the surface of a hypersphere $\mathbb{S}^d$, with angular directions observed in the real world frequently visualized on the circle or sphere.       
A lot of research has been concerned with successfully modelling and analysing this type of data, with distributions proposed by von Mises, Fisher and Watson \cite{Fisher1953} used in a range of applications including data clustering, segmentation and texture mapping\cite{PhDHasnat2014}. 

In this paper we propose to use von Mises-Fisher kernels to model the normal vectors of the shape (i.e. 2D contours, and 3D surface). Registration is then performed by minimizing a distance between two Kernel Density Estimates encoding the target and model shapes.
Section \ref{sec:soa} reviews the related work and in Section \ref{sec:L2:directional}, we propose   several new cost functions for registration using normal information. 
Section \ref{sec:shape:implemDet} outlines some of the implementation details of our method and experimental results (Section \ref{sec:expRes}) compare our approach to leading techniques for registration   \cite{Jian2011}  \cite{CPD2010}  \cite{GoICP2013}.

\section{Related works}
\label{sec:soa}

\subsection{Euclidean distance between GMMs}
\label{sec:L2:GMM}

Considering $p_1(x)$ and $p_2(x)$, two probability density functions (pdf) for the random vector $x \in \mathbb{R}^d$, the Euclidean $\mathcal{L}_2$ distance between $p_1$ and $p_2$ is defined as:
\begin{equation}
\mathcal{L}_2(p_1,p_2)=\int \lbrack p_1(x)-p_2(x) \rbrack^2 \ dx =\|p_1-p_2\|^2
\end{equation}
Many divergences have been defined for p.d.f. but $\mathcal{L}_2$ has the advantage of being explicit in the case of Gaussian Mixtures Models (GMM) and also robust to outliers when these GMMs are kernel density estimates (KDE) with Gaussian Kernels  \cite{ScottTechnometrics2001,Jian2011}.
$\mathcal{L}_2$ can be rewritten as: 
\begin{equation}
\mathcal{L}_2(p_1,p_2)= \|p_1-p_2\|^2=\|p_1\|^2+\|p_2\|^2-2 \langle p_1 |p_2 \rangle
\end{equation}
and simplified to 
\begin{equation}
\mathcal{L}_2E(p_1,p_2)= \|p_1\|^2-2 \langle p_1 |p_2 \rangle
\end{equation}
when $p_2$ is chosen as the empirical p.d.f. (i.e. KDE with Dirac Kernels) fitted on a target point set  \cite{ScottTechnometrics2001,Jian2011}.
Jian et al. applied $\mathcal{L}_2$ for shape registration in $\mathbb{R}^2$ and $\mathbb{R}^3$, where a parameterized GMM $p_1(x|\theta)$ is fitted to a target shape model represented by GMM $p_2$ \cite{Jian2011}: $\theta$ controls an affine or non rigid transformation (e.g. Thin Plate Spline) and the solution  $\hat{\theta}$ is computed such that it minimizes the $\mathcal{L}_2$ distance. While Jian et al encode  shapes as GMMs fitted to point clouds of contours ($\mathbb{R}^2$) or surfaces ($\mathbb{R}^3$),
Arellano et al. extended this shape registration approach into a multiple instance shape (ellipses) detection scheme, and also use directional information about the contour   \cite{ArellanoPR2015}. Similarly, in this paper we propose to consider directional information and we consider specific kernels suited for directional data as opposed to the Gaussian Kernel used by Arellano et al.
Another application for $\mathcal{L}_2$ registration in the image domain is to compute the colour transfer function to change the colour feel of images and videos \cite{Grogan2015,Grogan2015b,GroganIP2017}.
In this context,  $\mathcal{L}_2$ registration is shown to outperform  other popular schemes including these designed on the Optimal Transport framework \cite{GroganIP2017}.

Other registration techniques include maximum likelihood techniques, in which one point set is represented by a GMM and the other by a mixture of delta functions, which is equivalent to minimising the KL divergence between the two mixtures. In \cite{CPD2010}, Myronenko et al. also impose that the GMM centroids move coherently, preserving the structure of the point clouds. The Iterative Closest Point algorithm is another popular registration technique which alternates between estimating closest-point correspondences and a rigid transformation.  However, it can become trapped in local minima and requires a good initialisation, and many methods have been proposed as improvements \cite{Rusinkiewicz2001}. Yang et al. \cite{GoICP2013} propose to combine ICP with a branch-and-bound scheme which efficiently searches the rigid 3D motion space. They also derive novel upper and lower bounds for the ICP error function and provide a globally optimal solution to the 3D rigid registration problem. Rather than using fuzzy correspondences \cite{CPD2010,Jian2011} other techniques estimate explicit correspondences between the point sets using shape descriptors. Belongie et al. use shape context to register point sets, while other techniques use the local neighbourhood structure of the shapes \cite{Zheng2006,Lee2011}. Yang et al. \cite{GLMD2015} propose to combine global and local structural differences in a global and local mixture distance (GLMD) based method for non-rigid registration. Their iterative two step process alternately estimates the correspondences and computes the transformation. They define a distance which combines both global and local feature differences and use it to estimate point correspondences, and use an annealing scheme which ensures that the defined distance gradually changes from a local distance to a global distance.

\subsection{Directional data}
\label{sec:kernelsForDD}

We consider the $d$-dimensional unit random vector $u$ such that $\|u\|=1$ ($u\in \mathbb{S}^{d-1}$ with $\mathbb{S}^{d-1}$ the hypersphere in $\mathbb{R}^{d}$ ). Several distributions exist for random unit vectors and some are presented in this section. Applications of such distributions include  RGB-D image segmentation  \cite{PhDHasnat2014} and  structure from motion in 360 video \cite{Guan17} amongst others \cite{Dalal2005,IROS2012,MLSP2014}.

\subsubsection{von Mises-Fisher kernel}
The von Mises-Fisher probability density function for a random unit vector $u\in\mathbb{S}^{d-1}$ is defined as:
\begin{equation}
vMF(u;\mu,\kappa)= C_d(\kappa) \ \exp\left(\kappa \ \mu^{T} u\right)
\end{equation}
with parameters $\kappa\geq 0$ and $\|\mu\|=1$ and the normalising constant $C_d$ is
defined as:
\begin{equation}
C_d(\kappa)=\frac{1}{\int_{\mathbb{S}^{d-1}} \exp\left(\kappa \
    \mu^{T} u\right) \ du}
=\frac{\kappa^{\frac{d}{2}-1}}{(2\pi)^{\frac{d}{2}}\
  \mathcal{I}_{\frac{d}{2}-1}(\kappa)}
\label{eq:Cd}
\end{equation}
with $\mathcal{I}_{\frac{d}{2}-1}$ the modified Bessel function of order $\frac{d}{2}-1$.
The von Mises-Fisher distribution with parameters $\kappa$ and $\mu$
is noted $vMF(\mu,\kappa)$ for simplification.
For dimension $d=3$, $u$ is a unit vector in $\mathbb{R}^3$ and belongs
to the sphere $\mathbb{S}^2$ ,  and the
normalising constant in the von Mises-Fisher distribution is \cite{PhDHasnat2014}:
\begin{equation}
C_{3}(\kappa)=\frac{\kappa}{4\pi \sinh(\kappa)} 
\label{eq:C3}
\end{equation}

For $d\neq3$, the value $C_{d}(\kappa)$ is not directly available but can be computed using numerical integration. 
The value of the parameter $\kappa$ determines the shape of the distribution, with high values of $\kappa$ creating a distribution highly concentrated about the mean direction $\mu$, and low values of $\kappa$ creating an almost uniform distribution on $\mathbb{S}^{d-1}$.


\subsubsection{Watson distribution} The Watson distribution is also used to model axially symmetric directional data and is defined as follows:
\begin{equation}
    W_d(u;\mu,\kappa)=M_d(\kappa) \exp\left(\kappa \ (\mu^{T} u)^2\right)
\end{equation}
with the normalising constant:
\begin{equation}
M_d(\kappa)=\frac{1}{\int_{\mathbb{S}^{d-1}} \exp\left(\kappa \ (\mu^{T}u)^2\right) \ du }    
\end{equation}
This can be computed as $M_d(\kappa) = M(\frac{1}{2}, \frac{d}{2},\kappa)$, the confluent hyper-geometric function also known as the Kummer function, which is not directly available but can be approximated. Like the von Mises-Fisher distribution, the value of $\kappa$ determines the shape of the distribution.%




\section{$\mathcal{L}_2$ with Directional Data}
\label{sec:L2:directional}

Given two sets of  observations  $S_1 = \lbrace (x_1^{(i)},u_1^{(i)})\rbrace_{i=1,\cdots,n_1}$ and  $S_2 = \lbrace (x_2^{(j)},u_2^{(j)})\rbrace_{j=1,\cdots,n_2}$ for the random vectors $(x,u)\in\mathbb{R}^{d_x}\times \mathbb{S}^{d_u}$, 
we encode the model and target shapes using KDEs and register them by minimising the $\mathcal{L}_2$ distance between them. Here we assume that the shapes differ by some transformation $\phi$, controlled by the parameter $\theta$, which registers $S_1$ to $S_2$ and creates a new shape with observations $\tilde{S}_1=\{(\tilde{x}_1^{(i)}, \tilde{u}_1^{(i)})\}_{i=1,\cdots,n_1}$ that maps to $S_2$.

Probability density functions are modelled using sets $\tilde{S}_1$ and $S_2$ providing two possible distributions denoted $p_1$ and $p_2$ for the random vector $x\in \mathbb{R}^{d_x}$, ${u}\in \mathbb{S}^{d_u}$, and $(x,u)\in \mathbb{R}^{d_x}\times \mathbb{S}^{d_u}$.
 The $\mathcal{L}_2$ distance between $p_1$ and $p_2$ can then be computed as $\mathcal{L}_2(p_1,p_2)=\|p_1-p_2\|^2=\|p_1\|^2+\|p_2\|^2-2\langle p_1|p_2\rangle$ which is equivalent to maximizing the scalar product $\langle p_1|p_2\rangle$ when $\phi$ is a  rigid mapping \cite{Jian2011}. 

\subsection{Pdf modelling for $x\in \mathbb{R}^{d_x}$}

Jian et al used a KDE with a Gaussian kernel $\mathcal{N}(x; \tilde{x}_1^{(i)}, h)$ fitted to each point $\tilde{x}_1^{(i)}$ in $\tilde{S}_1$ \cite{Jian2011}:
\begin{equation}
p_1(x) = \frac{1}{n_1}\sum_{i=1}^{n_1} \mathcal{N}(x; \tilde{x}_1^{(i)}, h_1)
\end{equation}
Likewise the second set of points $\{x_2^{(j)}\}$ is modelled
using the Gaussian kernels:
\begin{equation}
p_2(x) = \frac{1}{n_2}\sum_{j=1}^{n_2}\mathcal{N}(x; x_2^{(i)}, h_2)
\end{equation}
 All Gaussians are chosen spherical in shape and have uniform scale and using the scalar product between two Gaussian  kernels (cf. Table \ref{tab:scalarProducts}), the following cost function is proposed to estimate $\theta$:
\begin{multline}
     \mathcal{C}^x  = \frac{1}{n_1 n_1}\sum_{j=1}^{n_1}\sum_{i=1}^{n_1}\ \frac{1}{\sqrt{4h_1^2\pi}} \exp\left( \frac{- \|\tilde{x}_1^{(j)}-\tilde{x}_1^{(i)}\|^2}{4h_1^2}  \right)
    \\ -\frac{2}{n_1 n_2}  \sum_{j=1}^{n_2}\sum_{i=1}^{n_1}\ \frac{1}{\sqrt{2(h_1^2 +h_2^2)\pi}} \exp\left( \frac{- \|x_2^{(j)}-\tilde{x}_1^{(i)}\|^2}{2(h_1^2 + h_2^2)}  \right) 
\end{multline}
This cost function is equivalent to that proposed for shape registration by Jian et al.\cite{Jian2011}. 
Note that if we had instead fitted a Dirac kernel to each of the points in $\{x_2^{(j)}\}$ we would have obtained the following cost function:
\begin{multline}
 \mathcal{C}_\delta^x  =\frac{1}{n_1 n_1} \sum_{j=1}^{n_1}\sum_{i=1}^{n_1}\ \frac{1}{\sqrt{4h_1^2\pi}} \exp\left( \frac{- \|\tilde{x}_1^{(j)}-\tilde{x}_1^{(i)}\|^2}{4h_1^2}  \right)\\
 -\frac{2}{n_1 n_2} \sum_{j=1}^{n_2}\sum_{i=1}^{n_1}\ \frac{1}{\sqrt{2h_1^2\pi}} \exp\left( \frac{- \|x_2^{(j)}-\tilde{x}_1^{(i)}\|^2}{2h_1^2}  \right)  
\end{multline}
which is equivalent to  $\mathcal{C}^x$ when $h_2 = 0$.

\subsection{Pdf modelling for $u \in \mathbb{S}^{d_u}$}
\label{sec:Cu}

To model the first set of normals $\{u_1^{(i)}\}$ we propose a KDE with a von Mises-Fisher kernel $vMF(u; \tilde{u}_1^{(i)}, \kappa)$ fitted to each normal $\tilde{u}_1^{(i)}$ in $\tilde{S}_1$:
\begin{equation}
p_1(u) = \frac{1}{n_1}\sum_{i=1}^{n_1} vMF(u; \tilde{u}_1^{(i)}, \kappa_1)
\end{equation}
We propose to model the second set of normal vectors $\{u_2^{(j)}\}$
using either the empirical distribution:
\begin{equation}
p_2(u) = \frac{1}{n_2}\sum_{j=1}^{n_2}\delta(u-u_2^{(j)})
\label{eq:norm:p1:dirac}
\end{equation}
or using the von Mises-Fisher distribution:
\begin{equation}
p_2(u) = \frac{1}{n_2}\sum_{j=1}^{n_2}vMF(u; u_2^{(j)}, \kappa_2).
\label{eq:norm:p1:vMF}
\end{equation}


Using the definitions for $\langle p_1|p_2\rangle$ as given in Table \ref{tab:scalarProducts}, two cost functions used to estimate $\theta$ by minimising $\|p_1\|^2-2\langle p_1|p_2\rangle$ can then be defined as follows:
\begin{multline}
 \mathcal{C}_{\delta}^u  = \left(\frac{C_{d_u}(\kappa_1)}{n_1} \right)^2 \sum_{i=1}^{n_1}\sum_{j=1}^{n_1}  
C_{d_u}^{-1}\left(\|\kappa_1 \tilde{u}_1^{(i)}+\kappa_1 \tilde{u}_1^{(j)}\|\right)\\
 -\frac{2C_{d_u}(\kappa_1)}{n_1 n_2}  \sum_{i=1}^{n_1}\sum_{j=1}^{n_2}  
\exp(\kappa_1 \tilde{u}_1^{{(i)}T}u_2^{(j)}),
\end{multline}
based on the modelling for $p_2$ defined Equation (\ref{eq:norm:p1:dirac}) and 
\begin{multline}
 \mathcal{C}^u  = 
 \left(\frac{C_{d_u}(\kappa_1)}{n_1 } \right)^2 \sum_{i=1}^{n_1}\sum_{j=1}^{n_1}  
C_{d_u}^{-1}\left(\|\kappa_1 \tilde{u}_1^{(i)}+\kappa_1 \tilde{u}_1^{(j)}\|\right)\\ -\frac{2\ C_{d_u}(\kappa_1)C_{d_u}(\kappa_2)}{n_1 n_2}  \sum_{i=1}^{n_1}\sum_{j=1}^{n_2}  
C_{d_u}^{-1}\left(\|\kappa_1 \tilde{u}_1^{(i)}+\kappa_2 u_2^{(j)}\|\right)
\end{multline}
based on the modelling for $p_2$ defined in Equation (\ref{eq:norm:p1:vMF}).

Since both terms in $\mathcal{C}^u$ depend on the normalising constant $C_{d_u}(\kappa)$, the computation of $\mathcal{C}^u$ requires numerical integration when $d_u \neq 3$. On the other hand, when $\phi_{\theta}$ is a rigid transformation $\mathcal{C}_{\delta}^u$ can be simplified as 
\begin{equation}
\widehat{\theta} = \arg\max_{\theta}\bigg\{\mathcal{C}_{\delta}^u =  \sum_{i=1}^{n_1}\sum_{j=1}^{n_2}  
\exp(\kappa_1 \tilde{u}_1^{(i)T}u_2^{(j)})\bigg\}
\end{equation}
and therefore can be easily computed  $\forall d_u$. 
This is one of the main advantages of using the Dirac distribution to model one set of normal vectors.

\subsection{Pdf modelling for  $(x,u)\in \mathbb{R}^{d_x}\times \mathbb{S}^{d_u}$}

We investigate in this section a cost function which accounts for both the normal vectors and point positions of the shapes in the modelling.
For the transformed observations  a KDE with Gaussian kernels fitted to each point $\tilde{x}_1^{(i)}$ and a $vMF$ kernel fitted to each normal vector $\tilde{u}_1^{(i)}$ is modelled as follows:
\begin{equation}
p_1(x,u) = \frac{1}{n_1}\sum_{i=1}^{n_1} vMF(u;\tilde{u}_1^{(i)}, \kappa_1)\  \mathcal{N}(x; \tilde{x}_1^{(i)}, h_1)
\label{eq:entangle:p1}
\end{equation}
For the second set of observations we again propose two methods for modelling the point and normal vectors. First, we propose to fit a dirac Delta kernel to each normal vector $u_2^{(j)}$ and a Gaussian kernel to each point $x_2^{(j)}$ to create a KDE of the form:
\begin{equation}
p_2(x,u) = \frac{1}{n_2}\sum_{j=1}^{n_2} \delta(u - u_2^{(
j)})\ \mathcal{N}(x; x_2^{(j)}, h_2).
\label{eq:entangle:dirac}
\end{equation}
We also propose an alternate KDE with vMF kernels fitted to the normal vectors $\{u_2^{(j)}\}$ as in Equation \ref{eq:entangle:p1}:
\begin{equation}
p_2(x,u) = \frac{1}{n_2}\sum_{j=1}^{n_2} vMF(u ; u_2^{(
j)}, \kappa_2) \  \mathcal{N}(x; x_2^{(j)}, h_2).
\label{eq:entangle:vMF}
\end{equation}
Then the parameter $\theta$ is estimated by minimizing one of the following cost functions:
\begin{multline}
\mathcal{C}_\delta^{x,u} = 
    \frac{1}{n_1 n_1}\sum_{j=1}^{n_1}\sum_{i=1}^{n_1} 
    \langle  vMF(\tilde{u}_1^{(i)}, \kappa_1)| vMF(\tilde{u}_1^{(j)}, \kappa_1) \rangle \\
    \times \langle   \mathcal{N}(\tilde{x}_1^{(i)}, h_1)| \mathcal{N}( \tilde{x}_1^{(j)}, h_1) \rangle\\
    -\frac{2}{n_1 n_2} \sum_{j=1}^{n_2}\sum_{i=1}^{n_1} \langle  vMF(\tilde{u}_1^{(i)}, \kappa_1)| \delta(u_2^{(j)})\rangle \\
    \times \langle \mathcal{N}(\tilde{x}_1^{(i)}, h_1)| \mathcal{N}(x_2^{(j)}, h_2) \rangle 
\end{multline}
based on the modelling proposed in Equation (\ref{eq:entangle:dirac}), or 
\begin{multline}
\mathcal{C}^{x,u} = 
\frac{1}{n_1 n_1} \sum_{j=1}^{n_1}\sum_{i=1}^{n_1} 
    \langle  vMF(\tilde{u}_1^{(i)}, \kappa_1)| vMF(\tilde{u}_1^{(j)}, \kappa_1) \rangle \\
    \times \langle   \mathcal{N}(\tilde{x}_1^{(i)}, h_1)| \mathcal{N}( \tilde{x}_1^{(j)}, h_1) \rangle\\
    -\frac{2}{n_1 n_2} \sum_{j=1}^{n_2}\sum_{i=1}^{n_1} \langle  vMF(\tilde{u}_1^{(i)}, \kappa_1)| vMF(u_2^{(j)}, \kappa_2)\rangle \\
    \times \langle \mathcal{N}(\tilde{x}_1^{(i)}, h_1)| \mathcal{N}(x_2^{(j)}, h_2) \rangle 
\end{multline}
based on the modelling proposed in Equation (\ref{eq:entangle:vMF}). 
Using the appropriate scalar product definitions given in Table \ref{tab:scalarProducts}, the proposed cost functions can be written explicitly as:
\begin{multline}
  \mathcal{C}_\delta^{x,u} =
  \frac{C_{d}(\kappa_1)C_{d}(\kappa_1)}{n_1 n_1 \sqrt{4h_1^2\pi}}\times
   \sum_{i=1}^{n_1}\sum_{j=1}^{n_1}\\
   C_d^{-1}\left(\|\kappa_1 \tilde{u}_1^{(i)}+\kappa_1 \tilde{u}_1^{(j)}\|\right)\ 
   \exp\left( \frac{- \|\tilde{x}_1^{(j)}-\tilde{x}_1^{(i)}\|^2}{4h_1^2}  \right) \\
-\frac{2 C_d(\kappa_1)}{n_1 n_2\sqrt{2(h_1^2 + h_2^2)\pi}}\times  \sum_{i=1}^{n_1}\sum_{j=1}^{n_2}\\ \exp(\kappa_1 \  {u_2^{(j)}}^T\tilde{u}_1^{(i)})\ \exp\left( \frac{- \|x_2^{(j)}-\tilde{x}_1^{(i)}\|^2}{2(h_1^2 + h_2^2)}  \right) 
 \end{multline}
and 
\begin{multline}
\mathcal{C}^{x,u} =
  \frac{C_{d_u}(\kappa_1)C_{d_u}(\kappa_1)}{n_1 n_1\sqrt{4h_1^2\pi}} \times
   \sum_{i=1}^{n_1}\sum_{j=1}^{n_1} \\
   C_{d_u}^{-1}\left(\|\kappa_1 \tilde{u}_1^{(i)}+\kappa_1 \tilde{u}_1^{(j)}\|\right)
   \exp\left( \frac{- \|\tilde{x}_1^{(j)}-\tilde{x}_1^{(i)}\|^2}{2h_1^2}  \right) \\
-2 \frac{C_{d_u}(\kappa_1)C_{d_u}(\kappa_2)}{n_1 n_2 \sqrt{2(h_1^2 + h_2^2)\pi}} \times \sum_{i=1}^{n_1}\sum_{j=1}^{n_2}\\
C_{d_u}^{-1}\left(\|\kappa_1 \tilde{u}_1^{(i)}+\kappa_2 u_2^{(j)}\|\right) \  \exp\left( \frac{- \|x_2^{(j)}-\tilde{x}_1^{(i)}\|^2}{2(h_1^2 + h_2^2)}  \right) 
\end{multline}
Although both terms in $\mathcal{C}^{x,u}$ depend on the normalizing constant $C_{d_u}(\kappa)$, in $\mathcal{C}_\delta^{x,u}$ the term $\langle p_1 | p_2 \rangle $ is independent of this constant and can be computed for any dimension $d_u$. Therefore when $\phi$ is a rigid transformation, $\theta$ can be estimated for any dimension $d_u$ by maximizing the cost function:
\begin{multline}
 \mathcal{C}_{\delta}^{x,u} = \\ \sum_{i=1}^{n_1}\sum_{j=1}^{n_2} \exp(\kappa_1 \  {u_2^{(j)}}^T\tilde{u}_1^{(i)})\exp\left( \frac{- \|x_2^{(j)}-\tilde{x}_1^{(i)}\|^2}{2(h_1^2 + h_2^2)}  \right) 
  \end{multline}

\section{Implementation Details}

In this section we outline some of the implementation details of our algorithm when it was applied to registering two shapes $S_1$ and $S_2$ with point sets $\{x_1^{(i)}\}$ and $\{x_2^{(j)}\}$ and unit normal vectors $\{u_1^{(i)}\}$ and $\{u_2^{(j)}\}$ respectively. 
\label{sec:shape:implemDet}
\subsection{Transformation function $\phi$}
To test the proposed cost functions, we considered shapes differing by both a rigid and non-rigid transformation $\phi$. As a translation only affects the observations $\{x^{(i)}\}$ and not the normal vectors $\{u^{(i)}\}$, the cost functions $\mathcal{C}^{u}$ and $\mathcal{C}_{\delta}^{u}$ are invariant to translation. To ensure all cost functions are evaluated equally, when estimating a rigid transformation we omit a translation and only consider data differing by a rotation.

For shapes differing by a non-rigid deformation, we estimate a Thin Plate Spline transformation and do not include a regularisation term to control the non-linearlity of the transformation.  However this can be added by the user if necessary. The $N$ control points $c_j$ used to control the TPS transformations in our non-rigid experiments are chosen uniformly on a grid spanning the bounding box of the model shape.

\subsection{Algorithm}
\label{sec:shape:algor}

Given two point sets $\{x_1^{(i)}\}_{i = 1,..n_1}$ and $\{x_2^{(j)}\}_{j = 1,..n_2}$ representing the model and target shapes, our strategy for estimating the transformation $\phi(x, \theta)$ is summarised in Algorithm \ref{algo:estimation}. 
\begin{algorithm}[H]
\begin{algorithmic}
\Require $\hat{\theta}$ initialised so that $\phi(x, \hat{\theta})=x$ (identity function)
\Require  $\kappa_{init}$, $\kappa_{final}$ and $h_{init}$, $h_{final}$ for $\mathcal{C}^{x,u}$. 
\Require $h_{step}$, $\kappa_{step}$
\Require Computation of unit normal vectors $\{u_1^{(i)}\}_{i = 1,..n_1}$ and $\{u_2^{(j)}\}_{j = 1,..n_2}$ from $\{x_1^{(i)}\}_{i = 1,..n_1}$ and $\{x_2^{(j)}\}_{j = 1,..n_2}$.
\State Choose $m$ points $\{x_1^{(i)}\}_{i = 1,..m}$ and $\{x_2^{(j)}\}_{j = 1,..m}$ and their associated unit normal vectors $\{u_1^{(i)}\}_{i = 1,..m}$ and $\{u_2^{(j)}\}_{j = 1,..m}$ for processing.
\State Start $h=h_{init}$ and $\kappa = \kappa_{init}$
\Repeat 
\State $ \hat{\theta} \leftarrow\arg\min_{\theta} \mathcal{C}(\theta)$ 
\State $h\leftarrow h_{step} \times h$
\State $\kappa \leftarrow \kappa_{step} \times \kappa$  
\Until{Convergence $h<h_{final}$ and $\kappa>\kappa_{final}$}
\Return $\hat{\theta}$ 
\end{algorithmic}
\caption{Our strategy for estimating the transformation $\phi(x, \theta)$.}
\label{algo:estimation}
\end{algorithm}
 In all of our experiments we let $h_1 = h_2 = h$ and $\kappa_1 = \kappa_2 = \kappa$.  To avoid local minima, we implement a simulated annealing strategy by gradually decreasing $h$ and increasing $\kappa$. The values chosen for all parameters can be found in the supplementary material. As the computation time of our proposed algorithms are dependent on the number of points in $S_1$ and $S_2$, we reduce the number of points processed by choosing a sample of $m$ points and their associated unit normal vectors from both $S_1$ and $S_2$.
 
 \begin{figure}[!h]
\begin{center}
\begin{tabular}{c c c}

\multicolumn{3}{c}{\includegraphics[width = .25\linewidth]{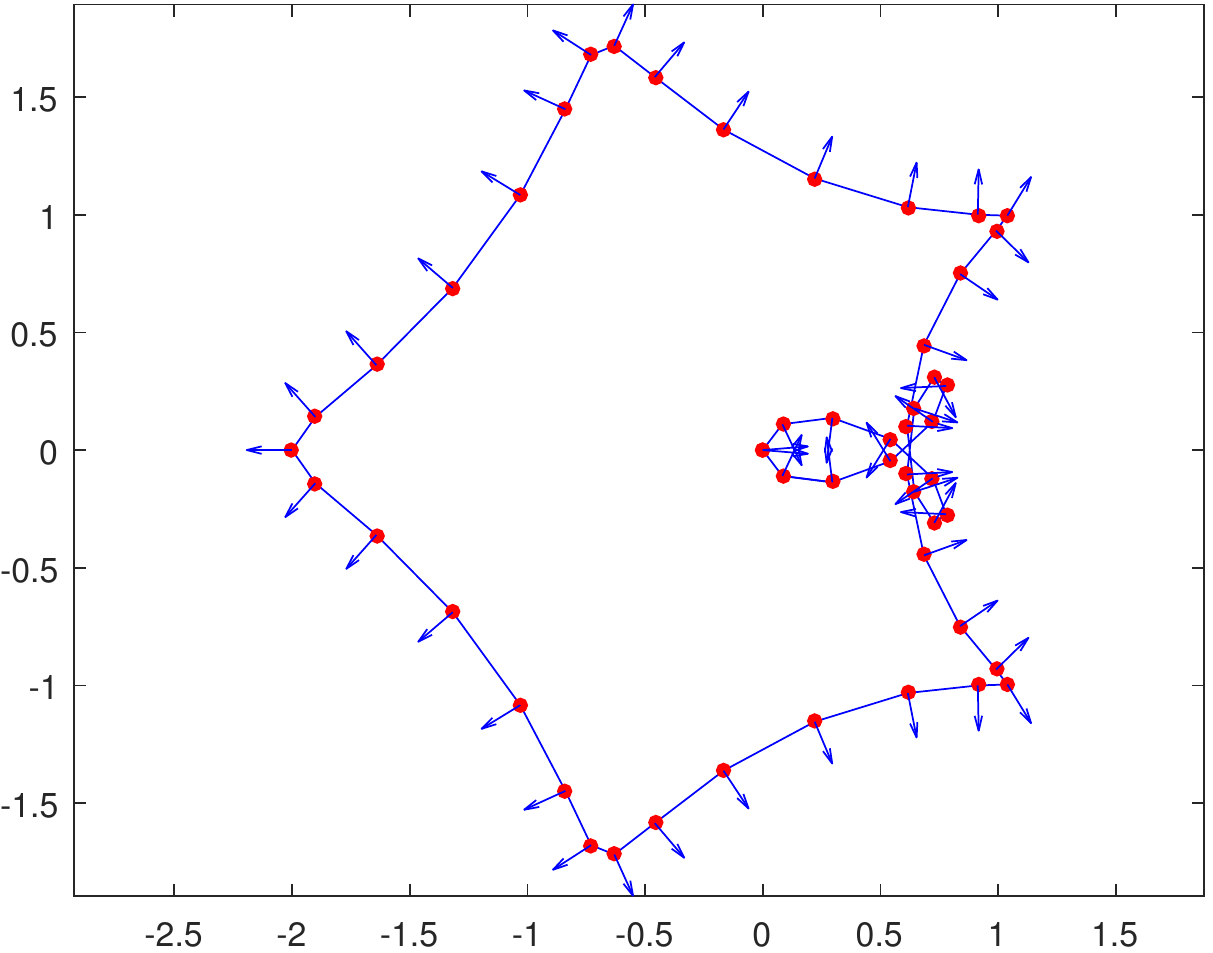}}\\\multicolumn{3}{c}{(a)}\\
\includegraphics[width = .25\linewidth]{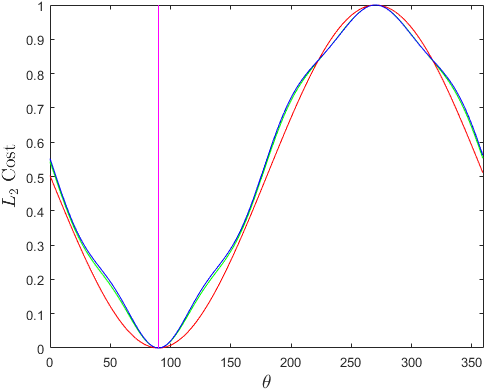}&
\includegraphics[width = .25\linewidth]{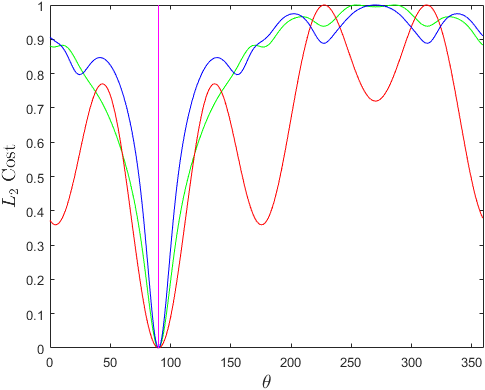}&
\includegraphics[width = .25\linewidth]{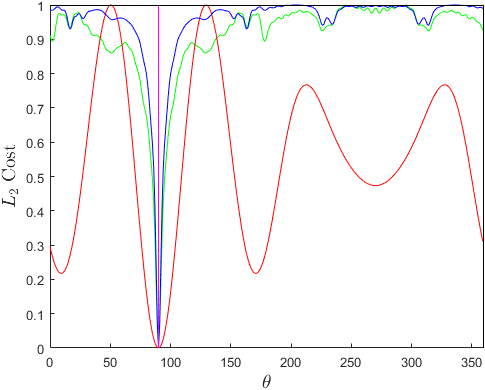}\\
 (b) & (c) & (d)\\
\footnotesize{$h = 4^{4}h_{final}$} & \footnotesize{$h = 4^{2}h_{final}$} & \footnotesize{$h = h_{final}$}\\

\footnotesize{$\kappa = \frac{1}{2^{4}}\kappa_{final}$} & \footnotesize{$\kappa = \frac{1}{2^2}\kappa_{final}$} & \footnotesize{$\kappa =\kappa_{final}$} \\
 \end{tabular}
\end{center}
 \caption{In (a) the parametric curve (blue) sampled at 50 locations (red) with normal vectors (shown in blue) is $S_1$, and was rotated by an angle of $\theta_{GT} = 90 \degree$ to generate $S_2$. In (b), (c) and (d) we show the effect that our simulated annealing strategy has on the cost functions computed using $S_1$ and $S_2$ as $\theta$ ranges from $1 \degree$ to $360 \degree$. To avoid local minima, $h$ is gradually decreased and $\kappa$ is gradually increased until $h = h_{final}$ and $\kappa = \kappa_{final}$. Green: $\mathcal{C}^{x}$; Red:  $\mathcal{C}^{u}$;  Blue: $\mathcal{C}^{x,u}$; Pink: $\theta_{GT}$.   }
\label{fig:SimAnn}
\end{figure}

\subsection{Normal Vector Computation} 
\label{sec:normComput}
We use several methods to compute the normal vectors $\{ u^{(i)}\}$ at the points $\{ x^{(i)}\}$. When testing our cost functions on 2D data, we use parametric curves and compute the normal vectors analytically. For 3D shapes in the form of meshes, we compute the normal vectors at a given vertex $x^{(i)}$ as the average of the normal vectors of each face connected to $x^{(i)}$. We also compute the normal vectors without exploiting the connectivity of a vertex, instead fitting a plane to it's nearest neighbours to compute the normal vector \cite{Hoffman1987,Hoppe1992}. 

\subsection{Computation Complexity}

Due to the double sum in all of the cost functions, their computational complexity depends on the number of points $n_1$  and $n_2$. When no point correspondences are chosen the computational complexity is of order $\mathcal{O}(n_1 \times n_2)$. Choosing to use $n$ point correspondences reduces this to $\mathcal{O}(n)$. The computation time needed by the gradient ascent technique also depends on the dimension of the latent space, which is determined by the transformation being estimated and the dimension $d_x$ of the space in which the shapes are defined. 
We do not provide analytical gradients to the gradient ascent algorithm when testing any of the proposed cost functions. Numerical methods are used instead to approximate the gradient at each iteration. While analytical gradients can be computed for $\mathcal{C}^{x}$ \cite{Jian2011}, and for all cost function when estimating a rotation, computing gradients for $\mathcal{C}^{x,u}$ or $\mathcal{C}_\delta^{x,u}$ when estimating a TPS transformation is not trivial, and we found that using numerical approximation was preferable. We used this approach for all cost functions to maintain consistency.  We use Matlab's optimisation function \textit{fmincon} when estimating a rotation transformation, and \textit{fminunc} when estimating a TPS transformation.

\subsection{Correspondences}
\label{sec:shape:corr}
In some cases, when estimating a non-rigid transformation, correspondences are used to reduce computational complexity and improve the registration result. When $n$ correspondences are chosen, the double sum $\sum_i^{n_1} \sum_j^{n_2}$ in all cost functions is reduced to $\sum_i^n$. To compute correspondences we used the method proposed by Yang et al.\cite{GLMD2015}. First, a global distance between the model and target point sets $\{ x_1^{(i)}\}$ and $\{ x_2^{(i)}\}$ is computed based on the squared Euclidean distance between each pair of points in the point sets. Then a local distance, which measures the difference in neighbourhood structure between each pair of points in the point sets, is computed. The local and global distances are combined in a cost matrix, and used to estimate a set of point correspondences. Yang et al also incorporate an annealing scheme which is designed to slowly change the cost minimisation from local to global. We also implemented this transition from local to global by incorporating it into our simulated annealing strategy.       

\subsection{Comparisons}
To evaluate our algorithm we compared our results to several state of the art registration techniques \cite{Jian2011,CPD2010,GoICP2013,GLMD2015}. The parameters chosen for each of these techniques is given in the supplementary material. To ensure that a fair comparison between Jian's cost function $\mathcal{C}^x$ and our proposed cost functions was presented, we altered some of the optimisation steps in the code provided by Jian et al.\footnote{https://github.com/bing-jian/gmmreg} so that they coincided with those implemented with our proposed cost functions. For example, we implemented the same simulated annealing framework for $\mathcal{C}^x$ as for $\mathcal{C}^u$, $\mathcal{C}_{\delta}^u$, $\mathcal{C}_{\delta}^{x,u}$ and $\mathcal{C}^{x,u}$. The optimisation changes made enhanced the results achievable by $C^x$ by allowing the function to avoid local minima and converge to a good solution.

\subsection{Evaluation}
In all experiments we chose the model and target point sets to be of equal size with $n_1 = n_2 = n$. The ground truth point correspondences between the  model and target shapes $\{ x_1^{(i)},x_2^{(i)}\}_{i = 1,..n} $ are also known and we evaluate the results of all algorithms by computed the mean square error (MSE) between corresponding points in the transformed model ($\{ \tilde{x}_1^{(i)}\}$) and target point sets as follows:
\begin{equation}
\frac{1}{n}\sum_{i=1}^{n} \| \tilde{x}_1^{(i)} - x_2^{(i)} \|
\end{equation}
Note that all $n$ points in the target shape and transformed model are used to compute the MSE, not just the subsample of size $m$ used to estimate $\phi(x, \hat{\theta})$.

\subsection{Experimental set up}

Our cost functions are evaluated experimentally (c.f. sections \ref{sec:expRes} and \ref{sec:3d:exp}) with the following settings:
\begin{itemize}
\item For rigid transformation,  the scalar product $\langle p_1 | p_2 \rangle$ is the cost function that is maximized as $\|p_1\|$ does not change in this case (In Sections \ref{sec:exp:2Drigid} and \ref{sec:exp:3Drigid}).
\item For non-rigid transformation,  $\|p_1 \| ^2 -2 \langle p_1 | p_2 \rangle$ is minimized to estimate $\theta$ (c.f.  Sections \ref{sec:2DnonRigid} and \ref{sec:3DnonRigid}).
\end{itemize}
 In Section \ref{sec:shape:compTime} we give details about the computational cost of our algorithm. 

\section{Experimental Results  2D}
\label{sec:expRes}

When considering the von Mises kernel as part of our cost functions, because its normalizing constant $C_3(\kappa)$ is explicitly available for $u\in \mathbb{S}^2$  while $C_2(\kappa)$  is not for 2D data ($u\in \mathbb{S}$), we  propose to  artificially define $u$ on $\mathbb{S}^2$ instead of $\mathbb{S}$ in this case  by adding a third dimensional coordinate to the normal vector (which is set to zero) to ease and speed up computation.

\subsection{2D Rotation Registration}
\label{sec:exp:2Drigid}


Curves $S_1$ and $S_2$ differ by a rotation $\phi$ which is defined as $\phi(x, \theta) = \mathrm{R}x$ (and $\phi(u, \theta) = \mathrm{R}u$ for the normal vector $u\in \mathbb{S}$, ignoring the zero value added to artificially extend $u\in \mathbb{S}^2$ ), where $\mathrm{R}$ is a 2D rotation matrix controlled by the angle $\theta$. 
We assess the estimation of the rotation angle $\theta$ using the cost functions $\mathcal{C}^{x}$ \cite{Jian2011}, $\mathcal{C}^{u}$, $\mathcal{C}_\delta^{u}$, $\mathcal{C}^{x,u}$ and $\mathcal{C}_{\delta}^{x,u}$. 


When testing our results we found that $\mathcal{C}^{u}$ and $\mathcal{C}_{\delta}^{u}$ as well as $\mathcal{C}^{x,u}$ and $\mathcal{C}_{\delta}^{x,u}$ are practically equivalent, so for ease of comparison we only present results for $\mathcal{C}^{u}$ and $\mathcal{C}^{x,u}$. \footnote{Results for $\mathcal{C}_{\delta}^{u}$ and $\mathcal{C}_{\delta}^{x,u}$ can be found in the supplementary material.}

\begin{figure}[!h]
\begin{center}
\begin{tabular}{c c }
\includegraphics[width = .45\linewidth]{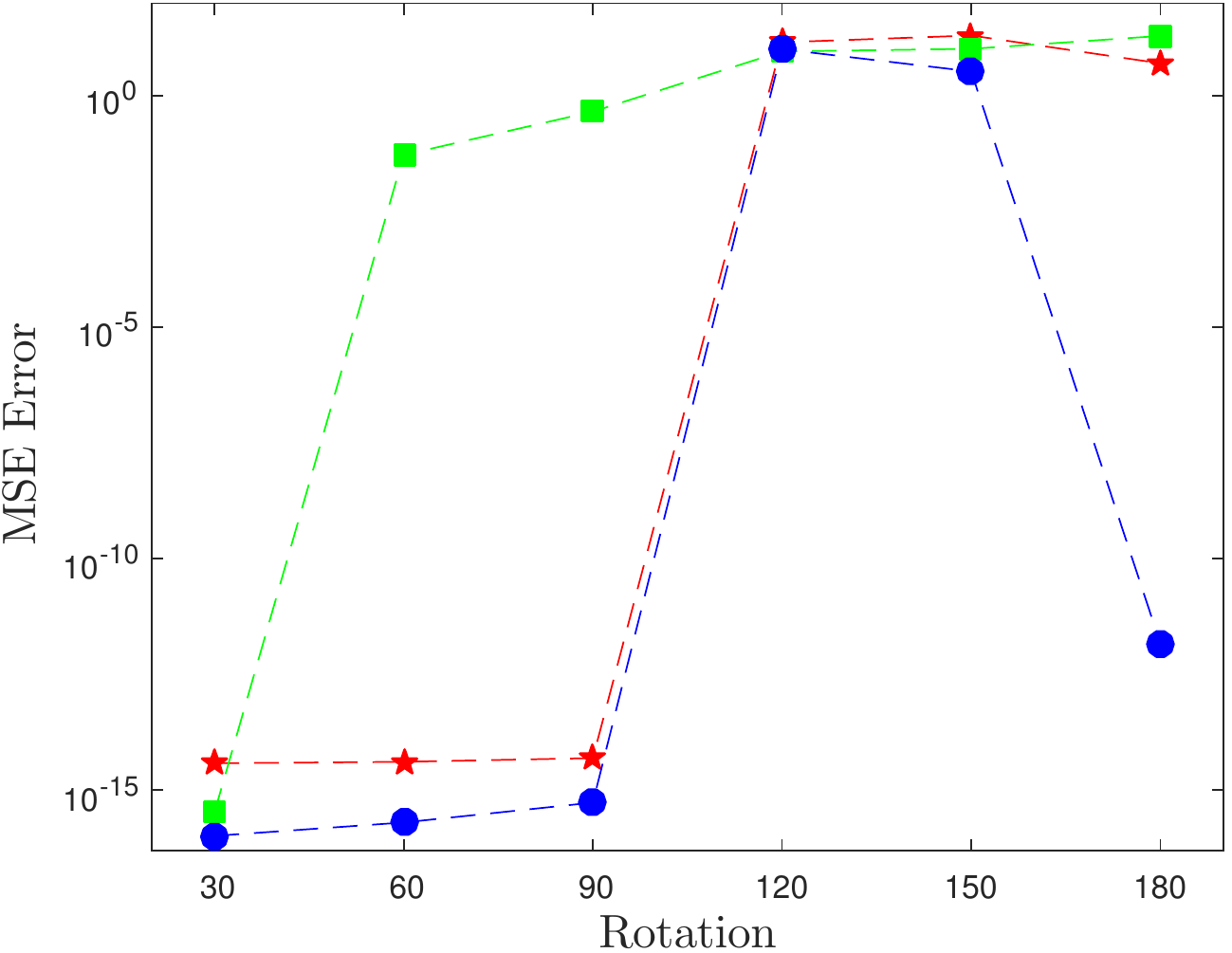}& 
\includegraphics[width = .45\linewidth]{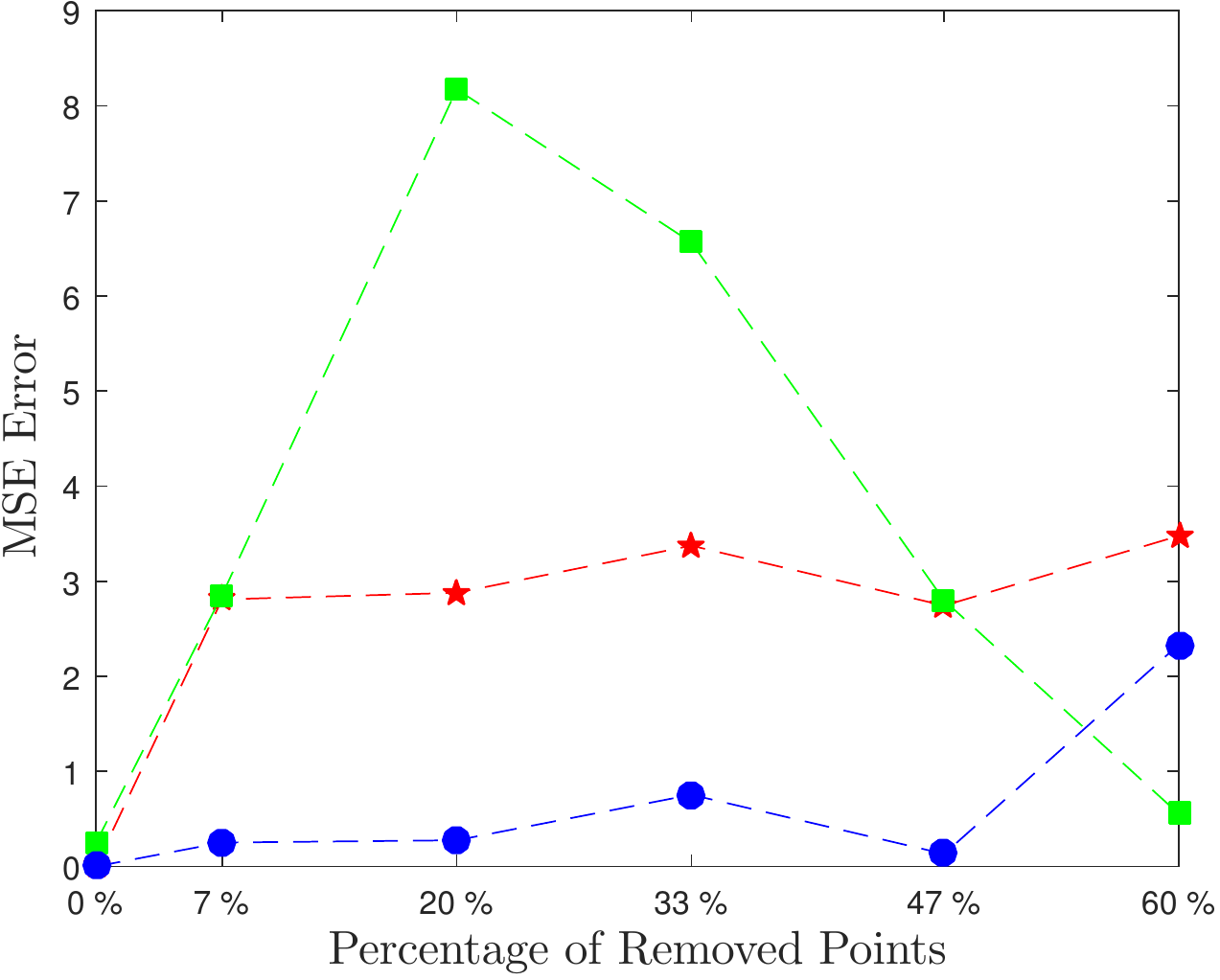}\\
\multicolumn{2}{c}{\includegraphics[width = .4\linewidth]{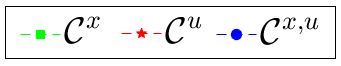}} \\
(a) & (b) \\
\end{tabular}
\end{center}
   \caption{MSE results comparing our cost functions on 2D data with rigid tranformation (rotation). In (a) the MSE value given at each rotation is the average over 10 curve registration results, as is the MSE value given at each percentage of removed points in (b).}
\label{fig:2DRigidErrors}
\end{figure}

\begin{enumerate}
\item  Figure \ref{fig:2DRigidErrors}(a)  presents the average MSE errors computed for $\mathcal{C}^x$ \cite{Jian2011}, $\mathcal{C}^u$ and $\mathcal{C}^{x,u}$ when considering rotation transformation between the two shapes to be registered. It shows that  $\mathcal{C}^{x,u}$ performs the best, followed by $\mathcal{C}^{u}$ and then $\mathcal{C}^{x}$.
Several values of $\theta$ were tested (reported in abscissa Fig. \ref{fig:2DRigidErrors}(a)), and for each value we created and registered 10 pairs of curves $S_1$ and $S_2$.

\item  Figure \ref{fig:2DRigidErrors}(b)  presents the average MSE errors computed for $\mathcal{C}^x$ \cite{Jian2011}, $\mathcal{C}^u$ and $\mathcal{C}^{x,u}$ when considering rotation transformation but with missing data between the two shapes to be registered. 
Again $\mathcal{C}^{x,u}$ performs the best, followed by $\mathcal{C}^{u}$ and then $\mathcal{C}^{x}$.
In this experiment a parametric curve $S_1$ is represented by  150 vertices $\{x_1^{(i)}\}_{i = 1,..150}$ with their corresponding normal vectors $\{u_1^{(i)}\}_{i = 1,..150}$ to create $S_1 = \lbrace (x_1^{(i)},u_1^{(i)})\rbrace_{i=1,..150}$. $S_2$  is created by rotating $S_1$ (with $\theta = 60$ degrees) and  a percentage of points (reported in abscissa Fig. \ref{fig:2DRigidErrors}(b)) are removed from $S_2$ before registration. For each percentage of points removed $(7\%, 20 \%, 33\%, 47\%, 60\%)$ we generating 10 pairs of curves $S_1$ and $S_2$ on which we tested the estimation of $\theta$.
As the number of removed points increases to $60\%$ or more, $\mathcal{C}^{x,u}$ had a higher tendency to fall into local minima and the error increases as a result. Similar results were found for other values of $\theta$ tested.
\end{enumerate}

\subsection{2D Non-rigid Registration}
\label{sec:2DnonRigid}


Shapes $S_1$ and $S_2$ differ now by a non-rigid deformation (defined as a TPS transformation with varying degrees of deformation). The estimated TPS transformation is controlled by $N = 12$ control points and our latent space of parameters to estimate is of dimension $(12 \times 2) + 6 = 30$.
Cost functions $\mathcal{C}^{x}$ \cite{Jian2011} and $\mathcal{C}^{x,u}$ are assessed, and $\mathcal{C}^{u}$ and $\mathcal{C}_\delta^{u}$ are omitted as normal information alone is not sufficient when estimating a non-rigid transformation.  $\mathcal{C}_\delta^{x,u}$  generates similar results to $\mathcal{C}^{x,u}$ and is not reported either.
As well as comparing $\mathcal{C}^{x}$ \cite{Jian2011} and $\mathcal{C}^{x,u}$, we also compare to other state of the art non-rigid registration  techniques namely CPD \cite{CPD2010} and GLMD \cite{GLMD2015}. For comparison we implement a similar experimental framework as that presented by Yang et al.\cite{GLMD2015} and for this reason we also normalize all curves so that they lie within $[0,1] \times [0,1]$.

\begin{figure*}[t]
\begin{center}
\begin{tabular}{ccc}
\includegraphics[width = .3\linewidth]{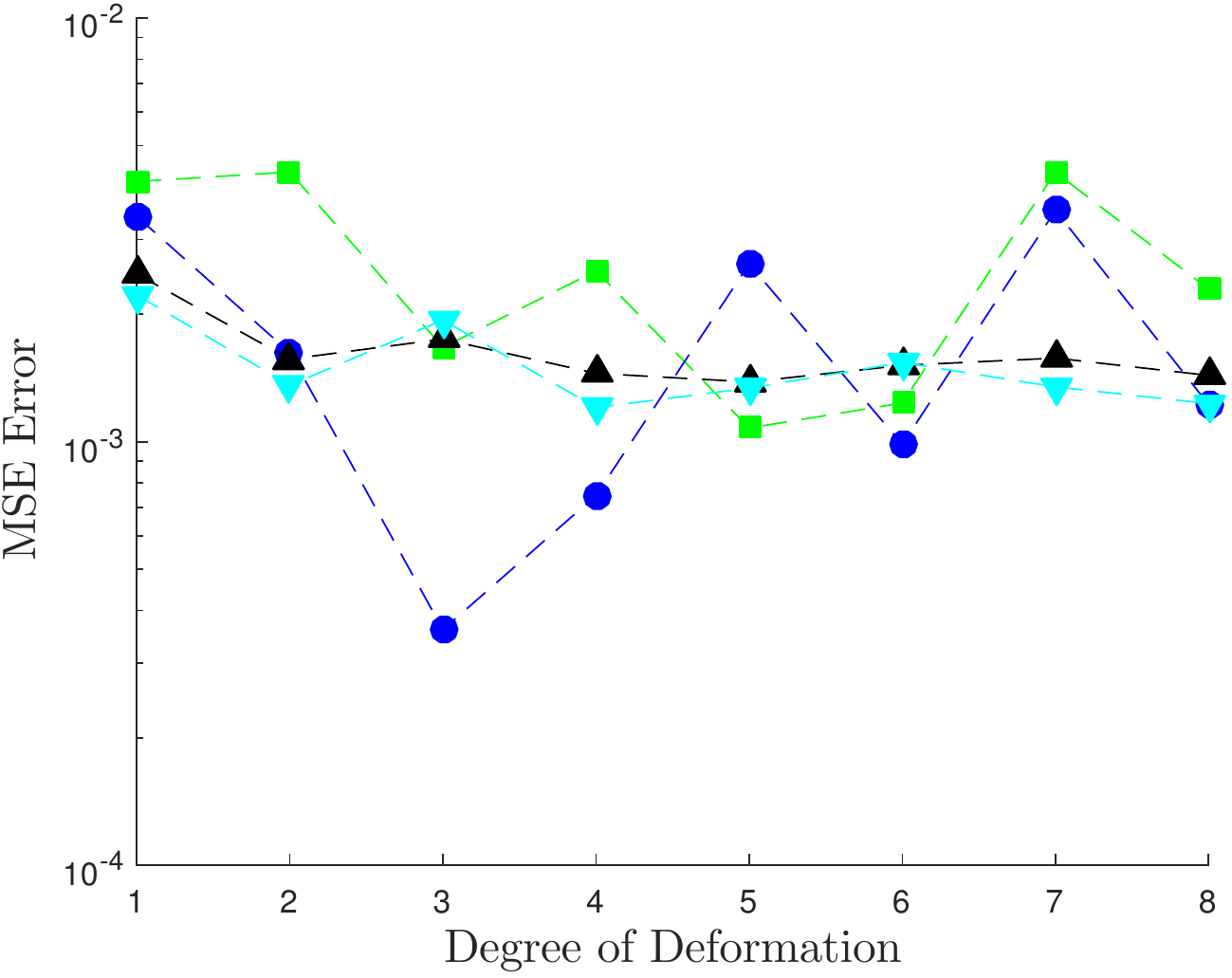}&
 \includegraphics[width = .3\linewidth]{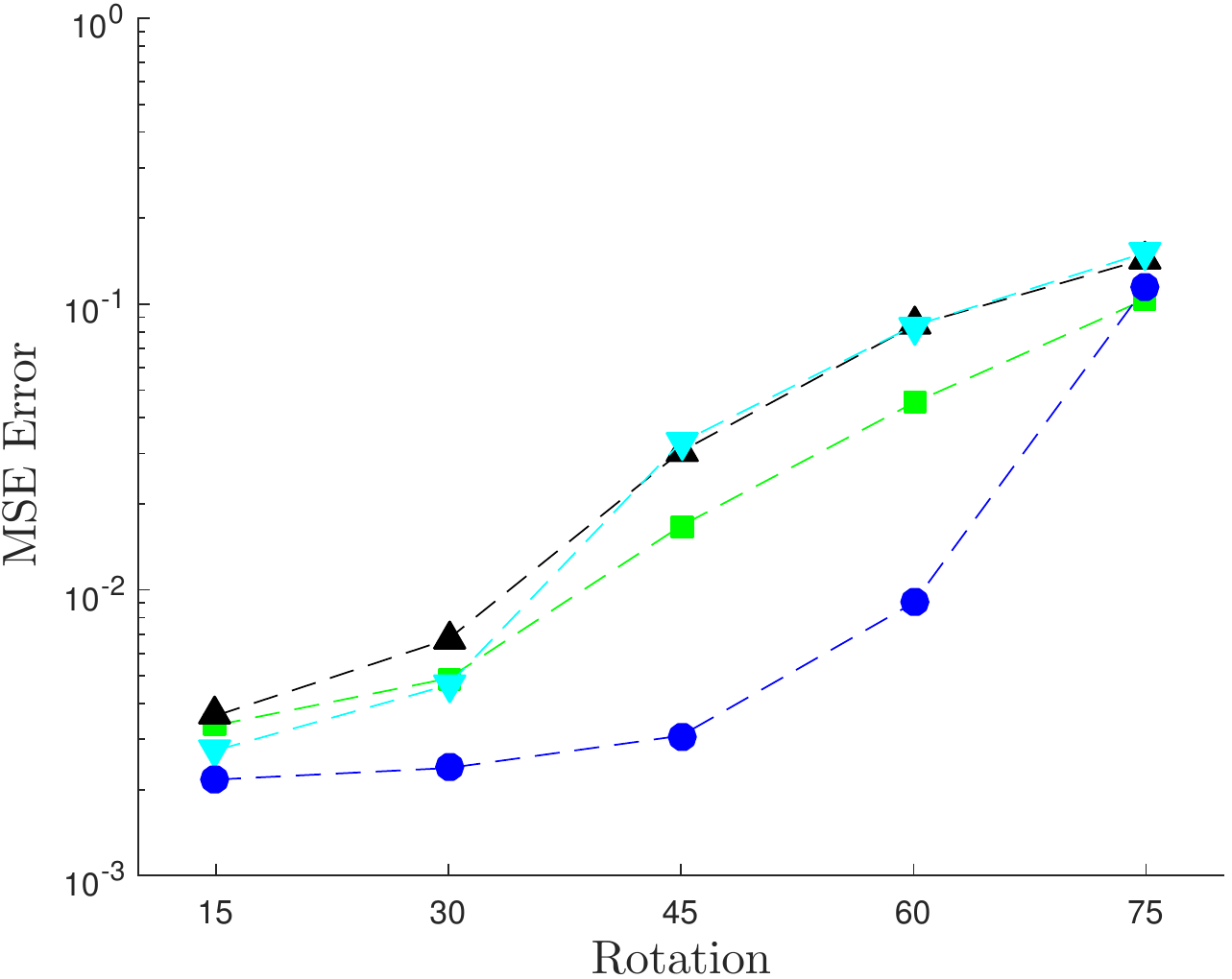}&
\includegraphics[width = .3\linewidth]{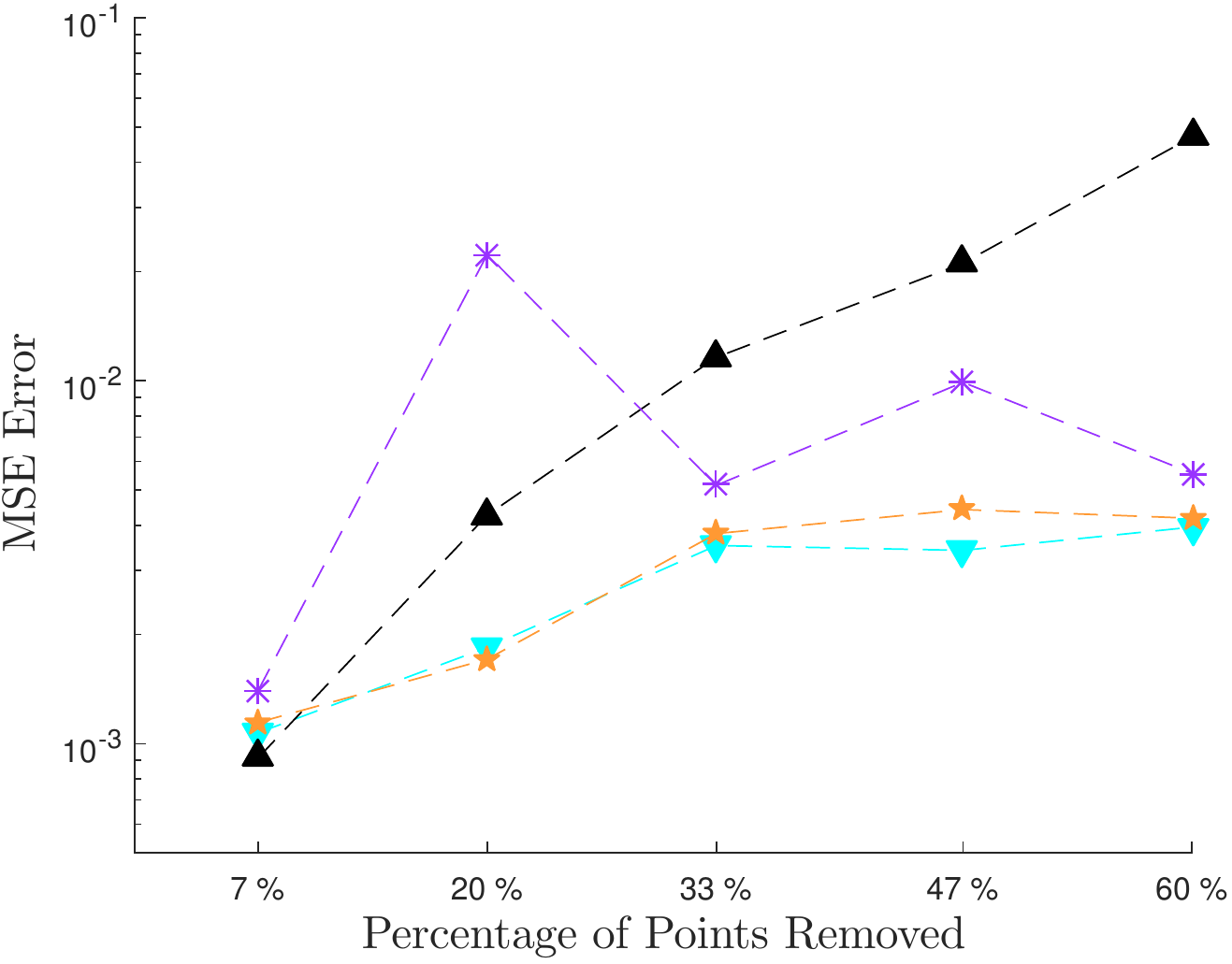} \\
\multicolumn{3}{c}{\includegraphics[width = .6\linewidth]{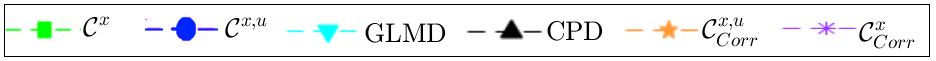}} \\
(a) & (b) & (c) \\
\end{tabular}
\end{center}
   \caption{MSE  results for non-rigid registration with 2D data. (a) Deformation estimation results with degree of deformation varying from 1 to 8; (b) Deformation and rotation estimation, with degree of deformation 4 and rotation varying from $15\degree$ to $75 \degree$; (c) Deformation estimation with missing data. Our methods $\mathcal{C}^x$ \cite{Jian2011},  $\mathcal{C}^{x,u}$  $\mathcal{C}^x_{Corr}$ and  $\mathcal{C}^{x,u}_{Corr}$ Vs GLMD \cite{GLMD2015}, CPD\cite{CPD2010}.}
\label{fig:2DNonRigid:errorss}
\end{figure*}


\begin{enumerate}
\item 
For each level of non rigid deformation we register 120 pairs of shapes $S_1$ and $S_2$ and present the average MSE results in Figure \ref{fig:2DNonRigid:errorss}(a). 
 We found that in general $\mathcal{C}^{x,u}$ performs well, but fails on occasion  skewing the average MSE (i.e. at deformations of degree $5$ and $7$). Similar spikes appear in the results for $\mathcal{C}^{x}$\cite{Jian2011}. Both CPD\cite{CPD2010} and GLMD\cite{GLMD2015}  generate consistent results over all deformations.

\item 
Setting the degree of deformation to $4$,  rotations of $\pm 15 \degree, \pm 30 \degree, \pm 45 \degree , \pm 60 \degree $ and $\pm 75 \degree$ is added so that both rotation and non rigid parameters now need to be estimated to register $S_1$ and $S_2$.
At each rotation value we registered 240 pairs of deformed curves for each method and the mean square errors computed can be seen in Figure \ref{fig:2DNonRigid:errorss} (b). 
  $\mathcal{C}^{x,u}$ performs best, followed by $\mathcal{C}^{x}$, GLMD\cite{GLMD2015} and CPD\cite{CPD2010}. The addition of the normal information in the cost function ensured that in general $\mathcal{C}^{x,u}$ estimated the correct rotation and deformation, while in the case of the other cost functions, the non-rigid deformation parameters were often used to attempt to account for the rotation difference.    

\item 
Setting the degree of deformation to $4$ without rotation, a percentage of points is randomly removed from ${S_1}$ before registration on ${S_2}$. 
Figure \ref{fig:2DNonRigid:errorss}(c) shows that $\mathcal{C}_{corr}^{x,u}$ performed as well as GLMD\cite{GLMD2015}, followed by $\mathcal{C}_{corr}^{x}$ and CPD\cite{CPD2010}. Without correspondences we found that both $\mathcal{C}^{x,u}$ and $\mathcal{C}^{x}$ (not reported) tried to maximize the amount of overlap between the curves and rarely estimated the correct parameters. 
 For this experiment correspondences were estimated using Yang et al's technique, as described in Section \ref{sec:shape:corr}, and were used when optimizing  $\mathcal{C}_{corr}^{x,u}$ and $\mathcal{C}_{corr}^{x}$.  120 pairs of curves were registered at each level of missing data  (reported in abscissa)
 and the average MSE is reported in Figure \ref{fig:2DNonRigid:errorss}(c).

 \end{enumerate}

\section{Experimental Results in 3D}
\label{sec:3d:exp}

\subsection{3D Rotation Registration}
\label{sec:exp:3Drigid}

We now consider two 3D shapes $S_1 = \lbrace (x_1^{(i)},u_1^{(i)})\rbrace_{i=1,\cdots,n}$ and  $S_2 = \lbrace (x_2^{(j)},u_2^{(j)})\rbrace_{j=1,\cdots,n}$ which are represented by their point locations $\{x^{(i)}\} \in \mathbb{R}^3$ and normal vectors $\{ u ^{(i)} \} \in \mathbb{S}^2$. Shapes $S_1$ and $S_2$ differ by a rotation $\phi$ which is defined as $\phi(x, \theta) = \mathrm{R}x$ with $\theta = \mathrm{R}$. In this case our latent space is of dimension $9$. We compared our results to those obtained using Jian et al's method $\mathcal{C}^x$ \cite{Jian2011}, CPD \cite{CPD2010} and Go-ICP   \cite{GoICP2013}. The shapes used in this experiment are the Stanford Bunny, Dragon and Buddha meshes provided by the Stanford University Computer Graphics Laboratory \footnote{\url{http://graphics.stanford.edu/data/3Dscanrep/}}, and the Horse mesh provided by Sumner et al. \cite{Sumner2004}. 
Each of these shapes is stored in .ply format with both vertex and edge information available, from which normal vectors are easily calculated, as described in Section \ref{sec:normComput}. 
These meshes have between 5000 and 40000 vertices each, and a sub-sample of vertices and their corresponding normal vectors are used  in all of our experiments.

When testing our results we found that $\mathcal{C}^{u}$ and $\mathcal{C}_{\delta}^{u}$ as well as $\mathcal{C}^{x,u}$ and $\mathcal{C}_{\delta}^{x,u}$ are practically equivalent, so for ease of comparison we only present results for $\mathcal{C}_{\delta}^{u}$ and $\mathcal{C}_{\delta}^{x,u}$ in the following section. Further comparisons with $\mathcal{C}^{u}$ and $\mathcal{C}^{x,u}$ can be found in the supplementary material.



\begin{figure}[t]
\begin{center}
\begin{tabular}{ c c c c }
 Same Sampling & Different Sampling & Added Noise & \\

\includegraphics[width = .3\linewidth, height = .27\linewidth]{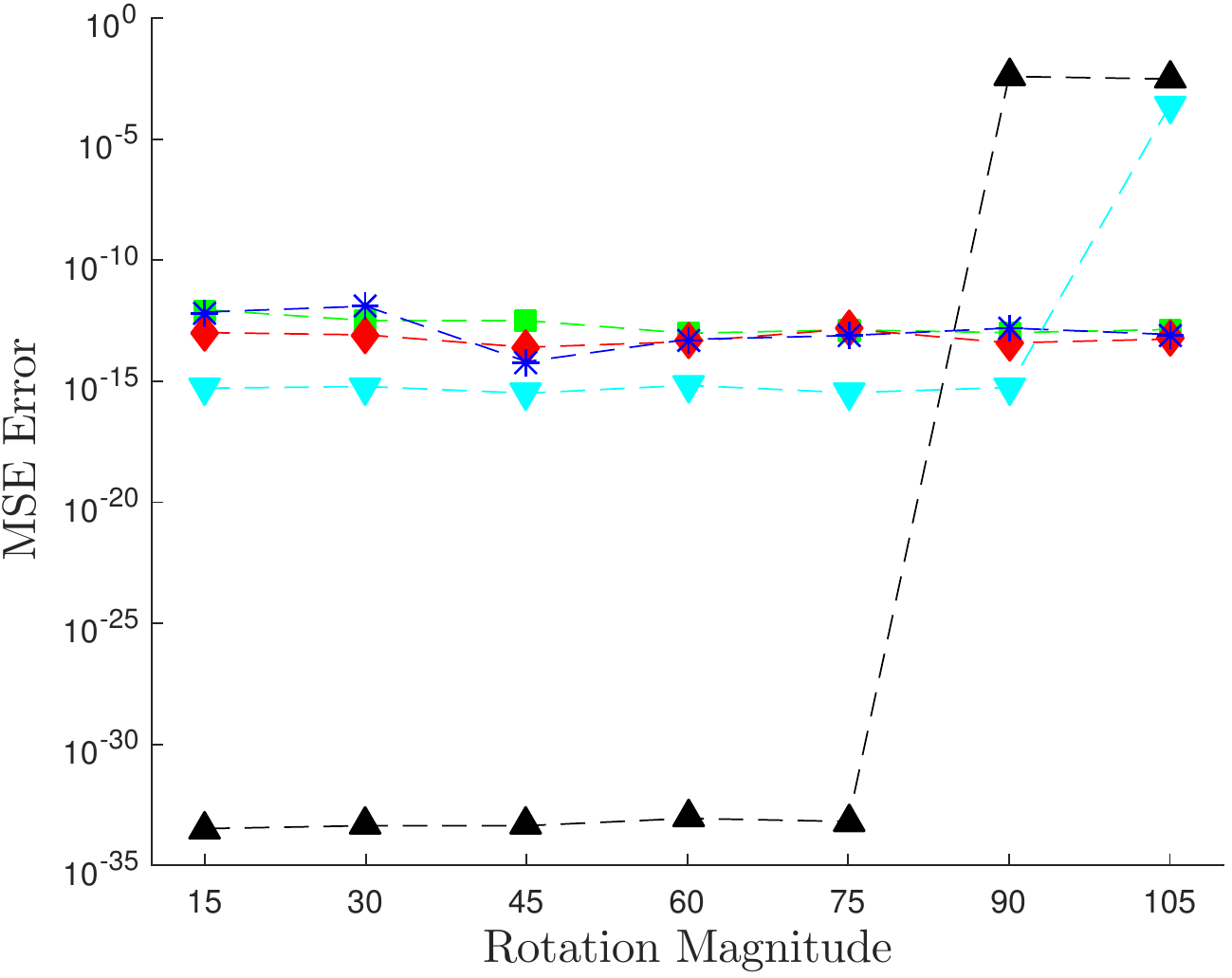}&

\includegraphics[width = .3\linewidth, height = .27\linewidth]{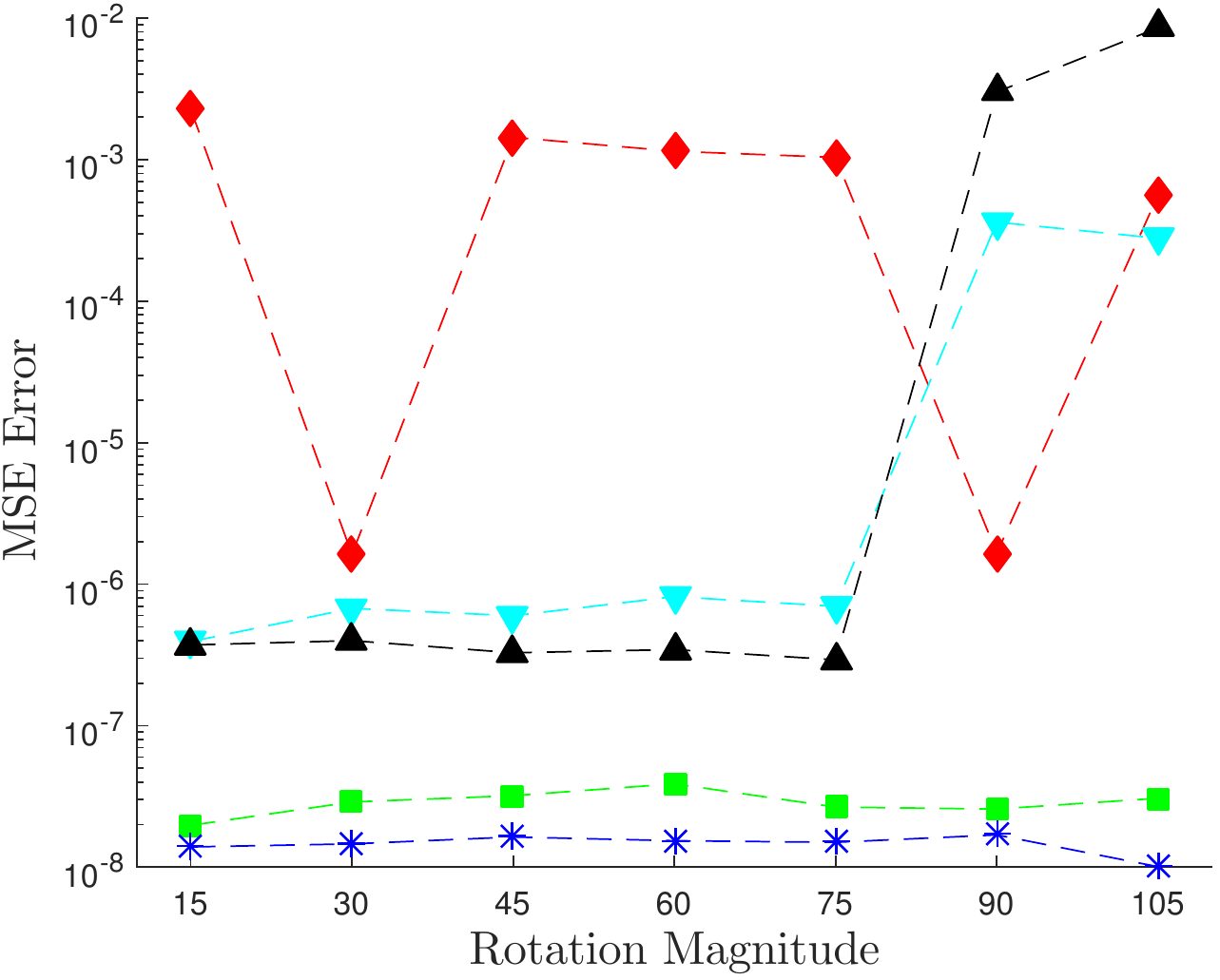}&
\includegraphics[width = .3\linewidth, height = .27\linewidth]{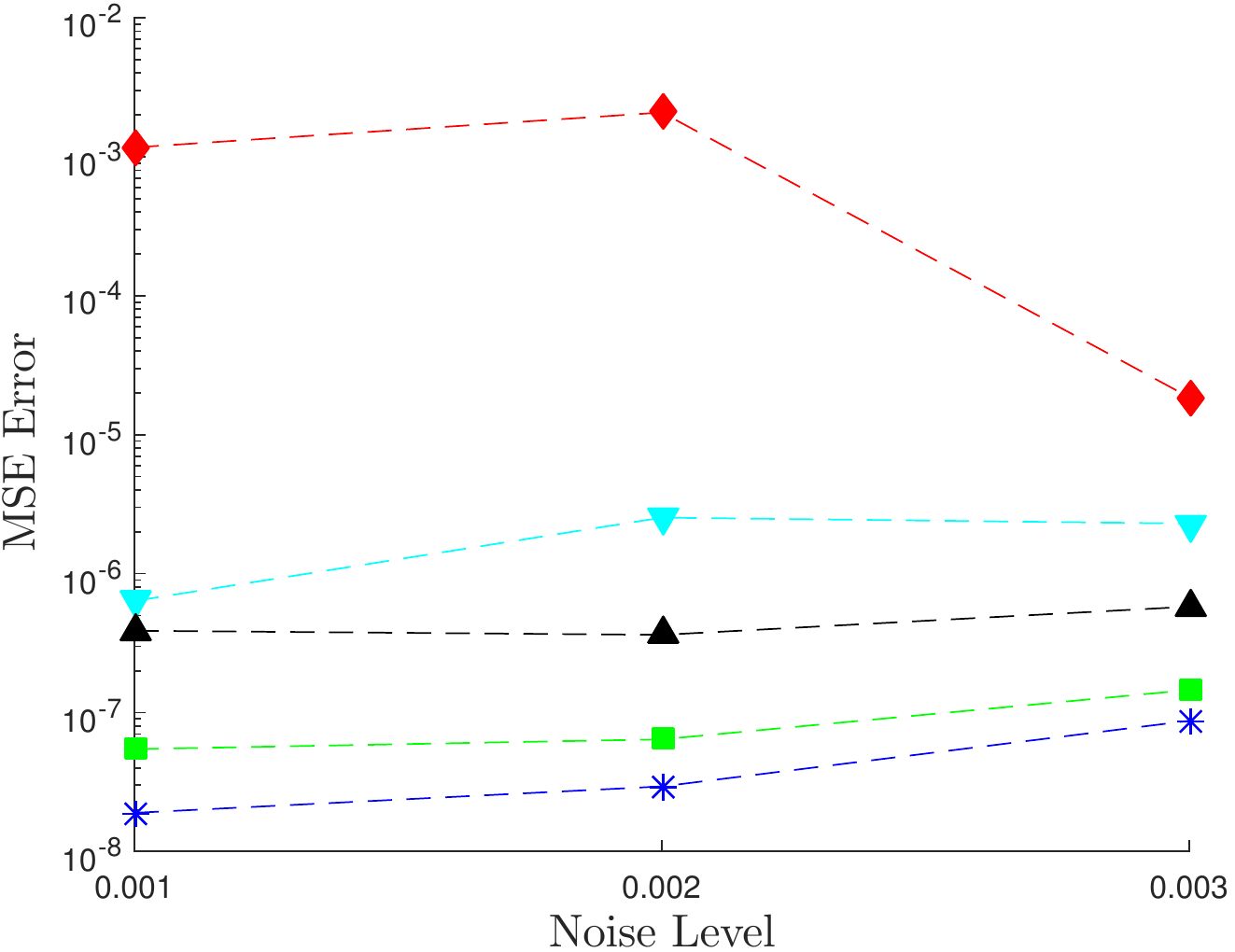}  \\
&Bunny &&\\

\includegraphics[width = .3\linewidth, height = .27\linewidth]{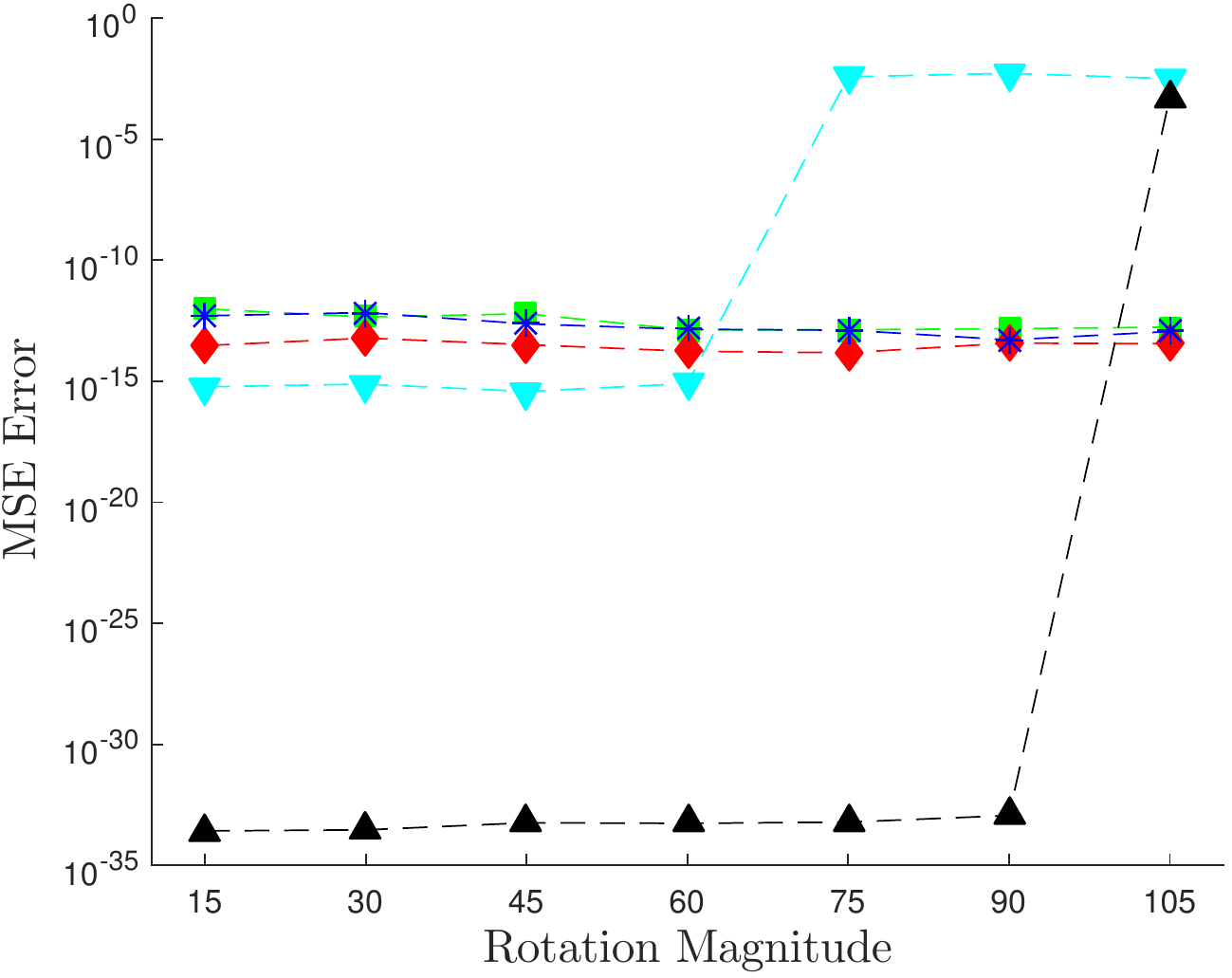}&
\includegraphics[width = .3\linewidth, height = .27\linewidth]{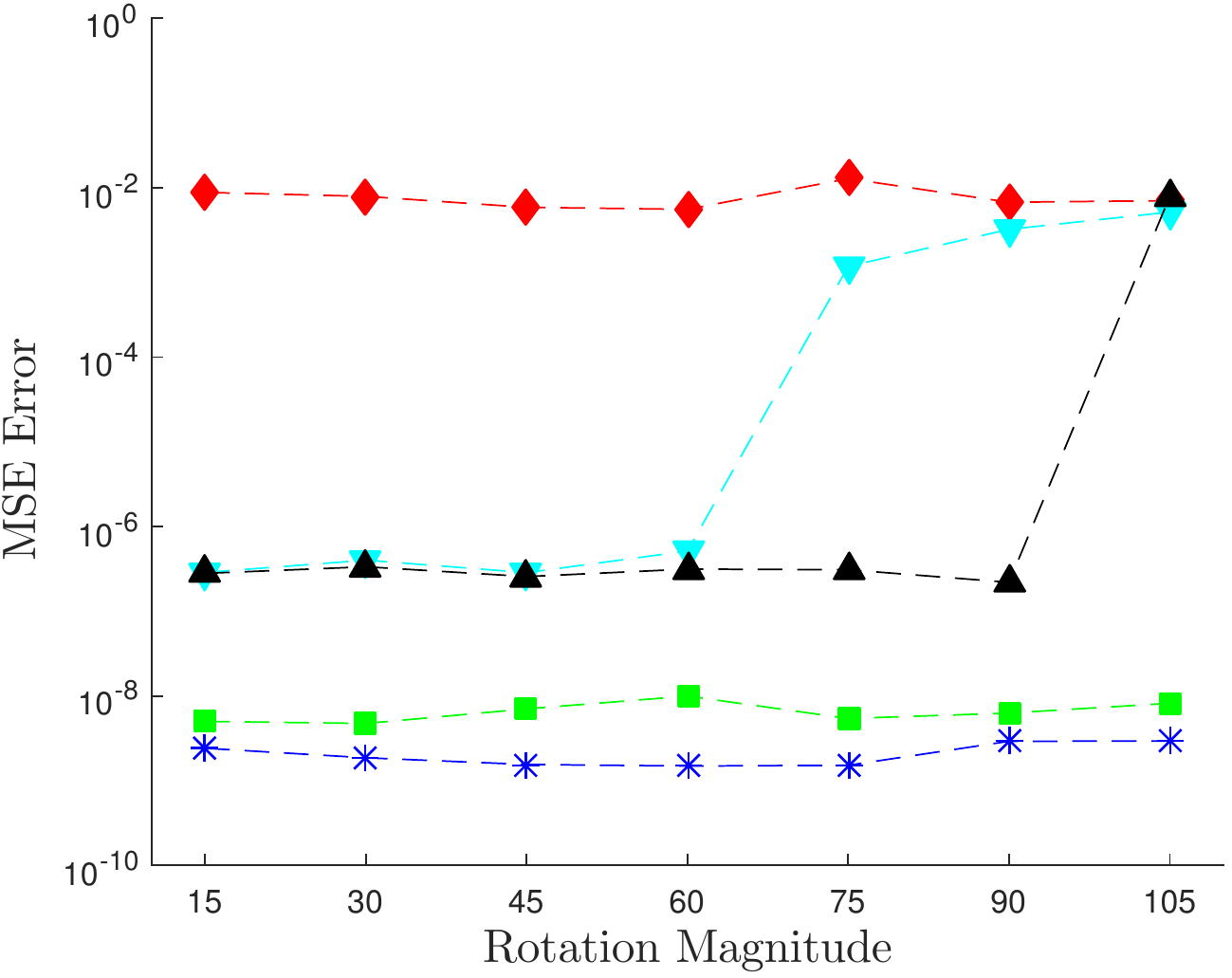}&
\includegraphics[width = .3\linewidth, height = .27\linewidth]{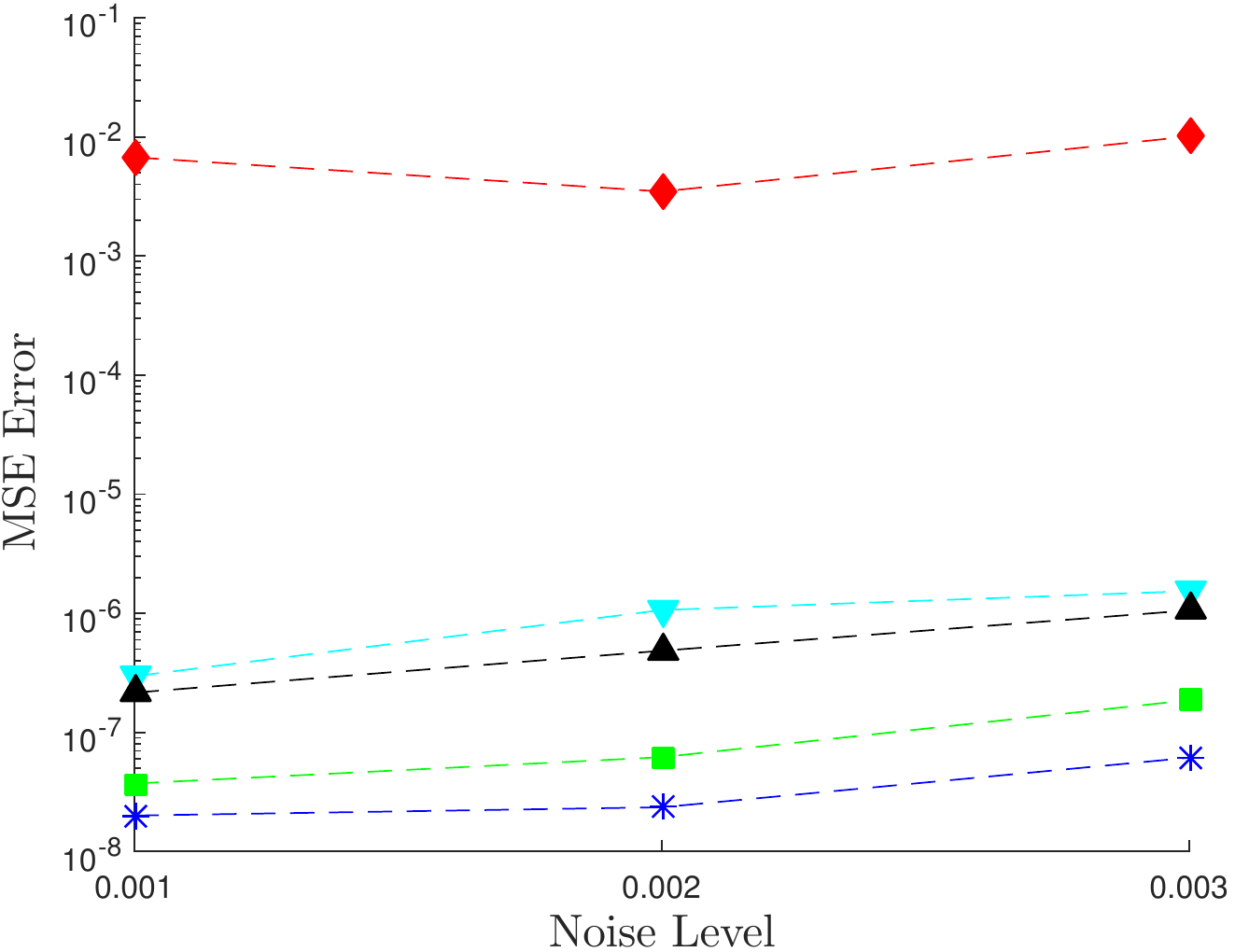} & \\

&Dragon& &\\

\includegraphics[width = .3\linewidth, height = .27\linewidth]{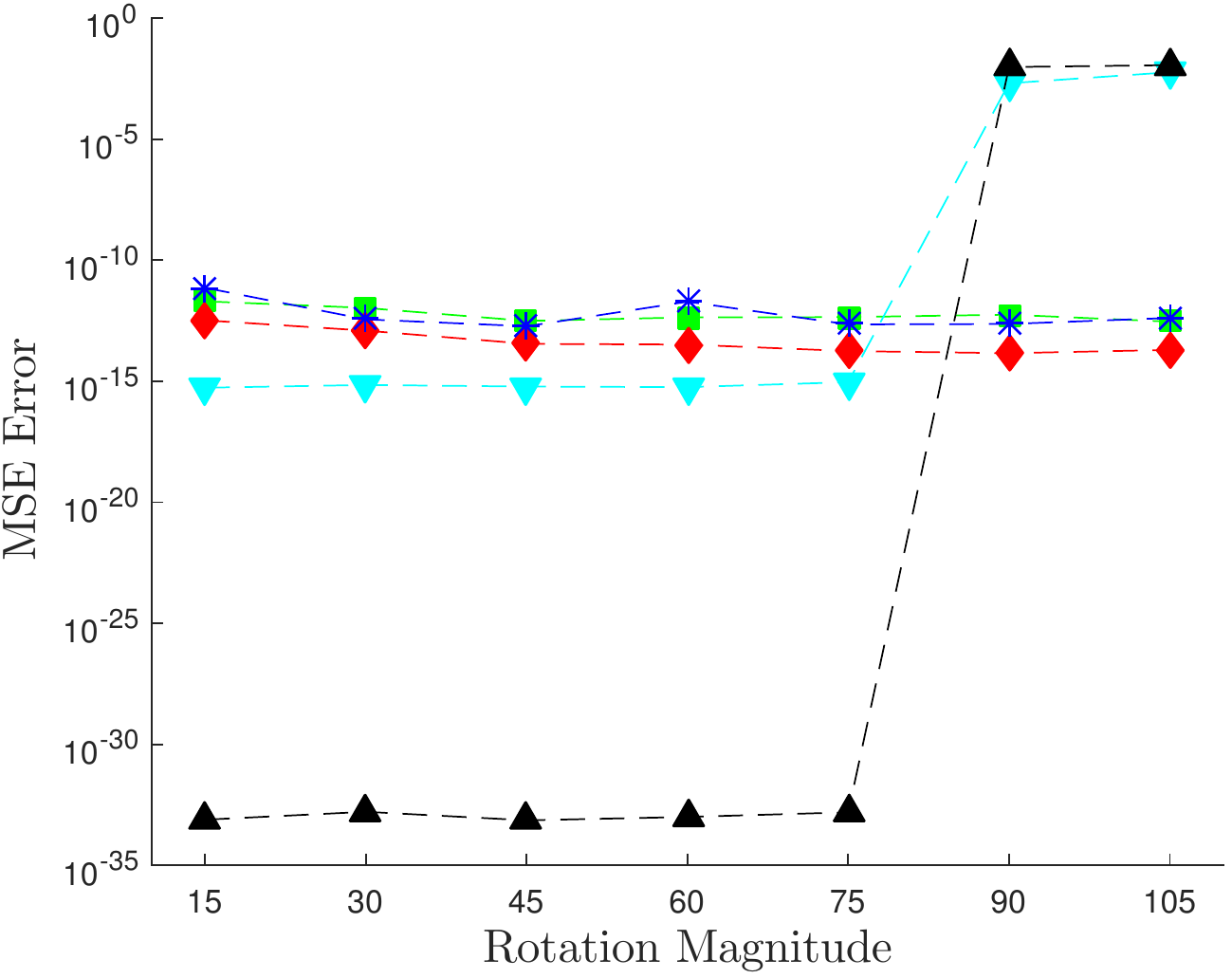} &
\includegraphics[width = .3\linewidth, height = .27\linewidth]{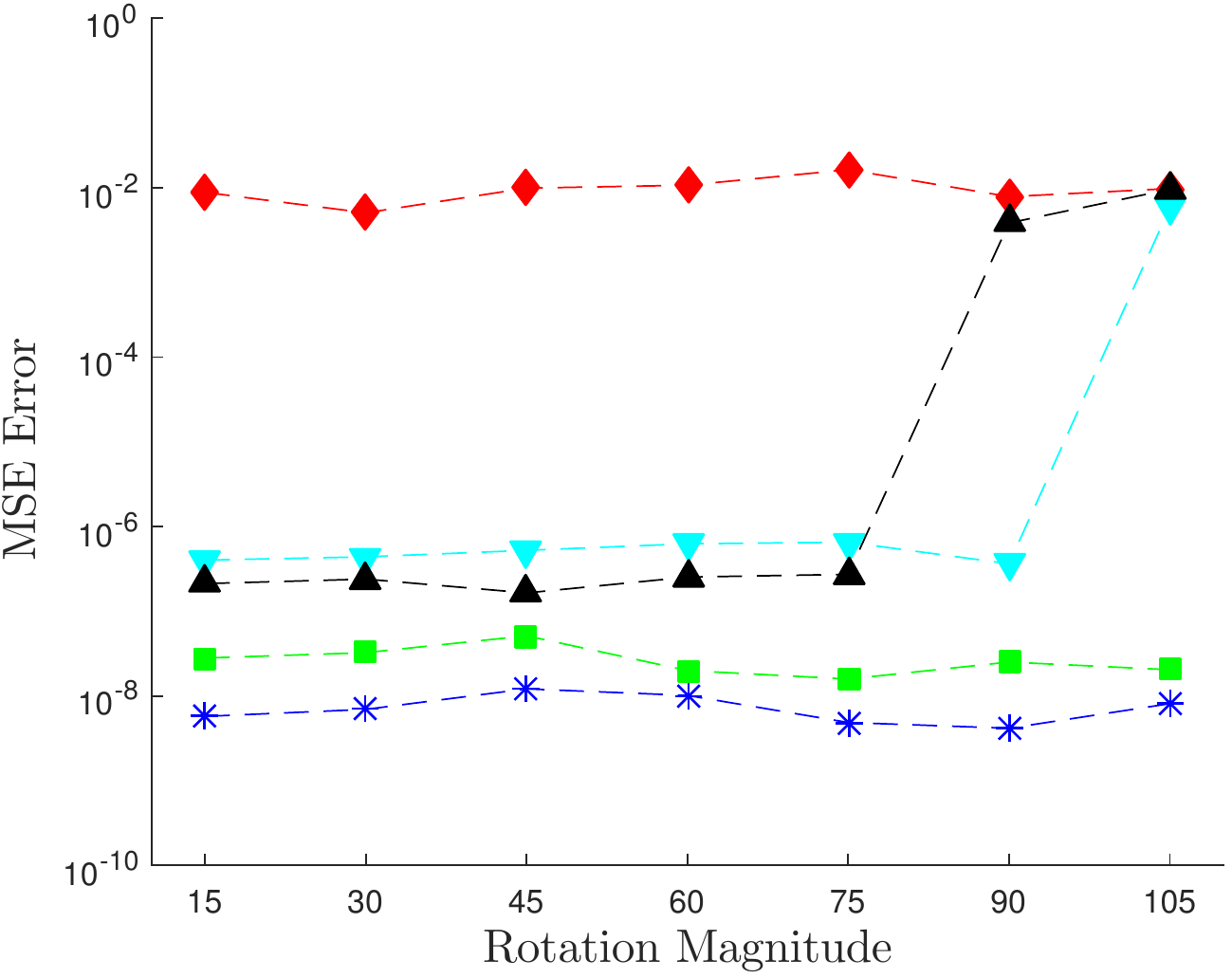}&
\includegraphics[width = .3\linewidth, height = .27\linewidth]{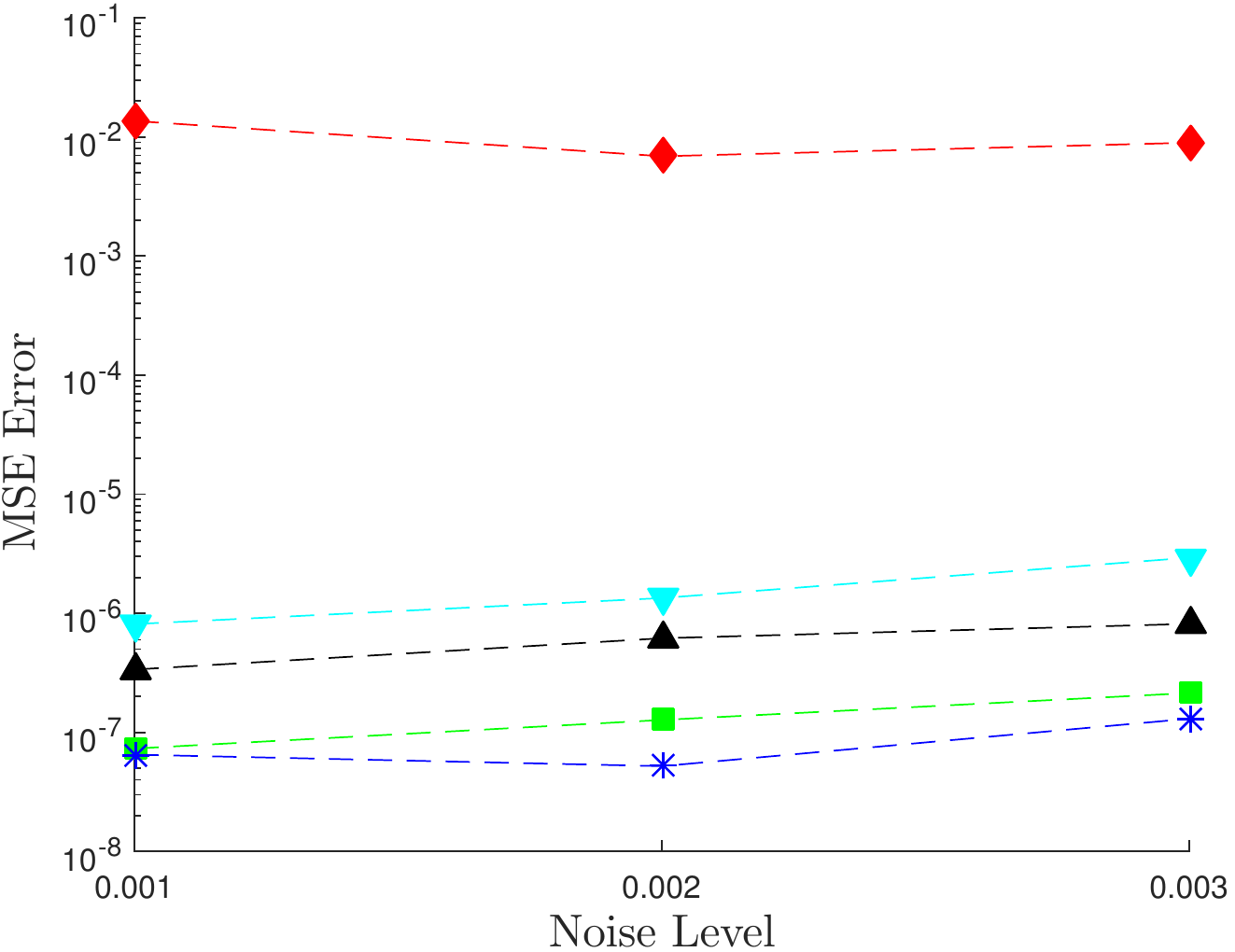} & \\
&Buddha &&\\

 \includegraphics[width = .3\linewidth, height = .27\linewidth]{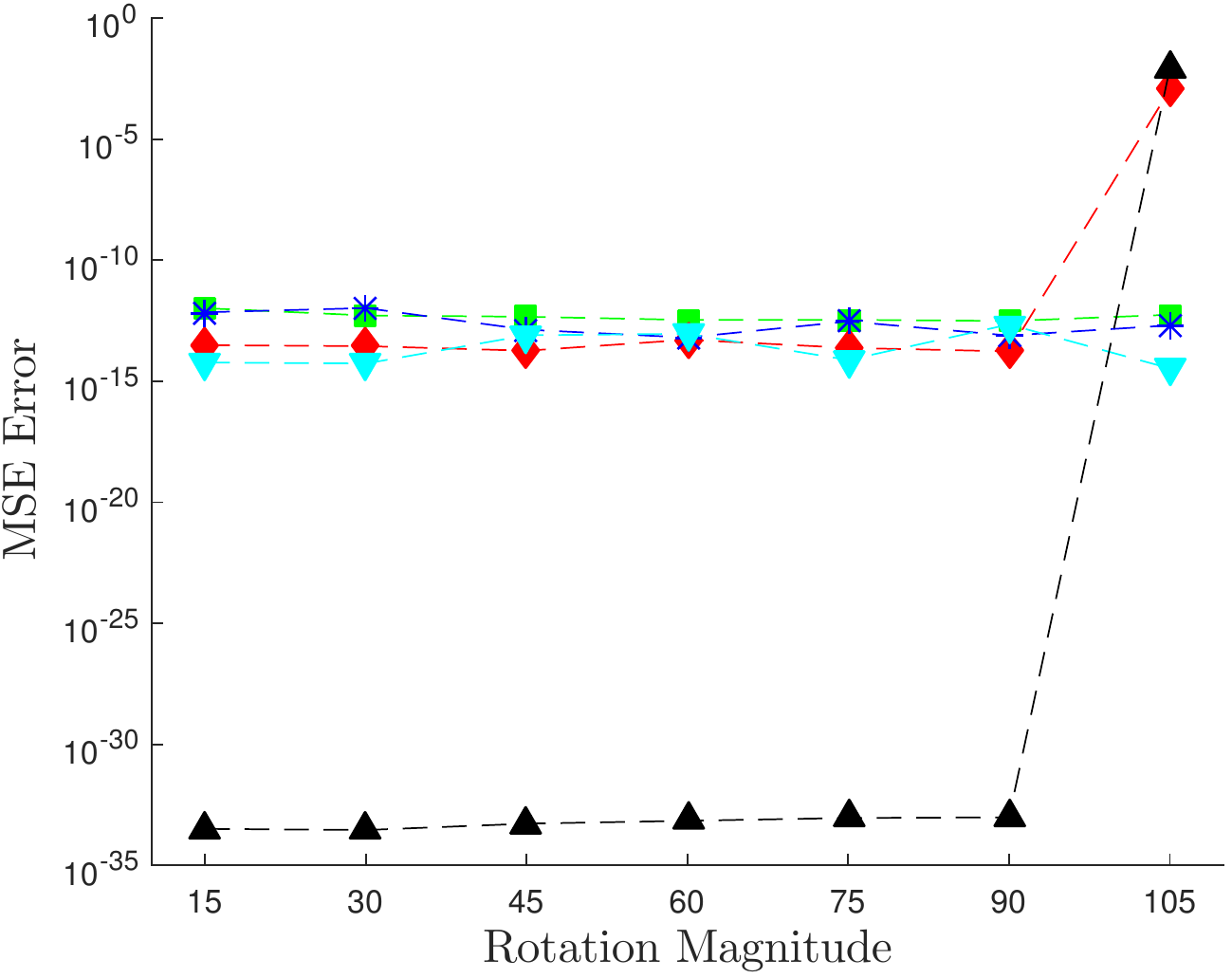}&
\includegraphics[width = .3\linewidth, height = .27\linewidth]{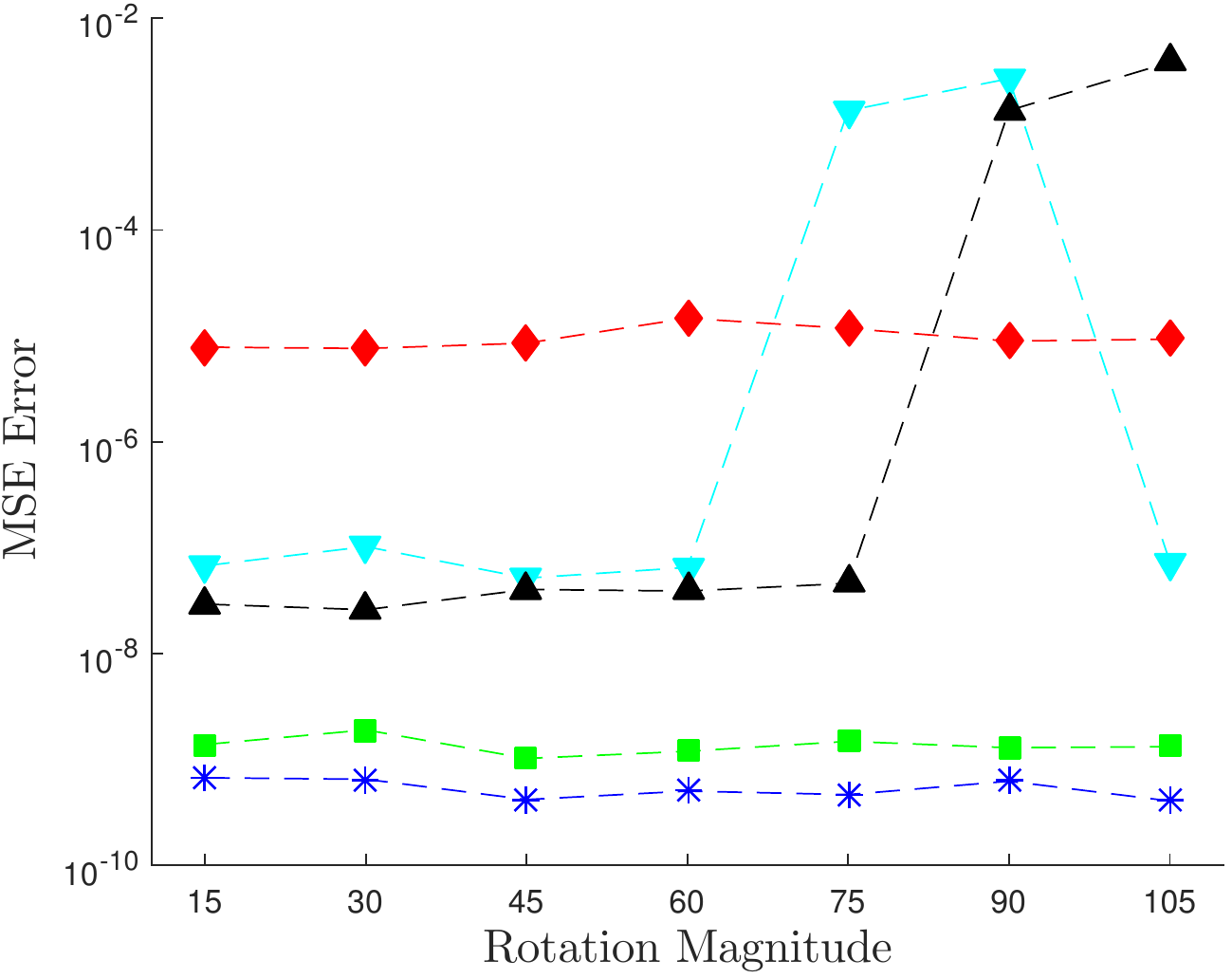}&
\includegraphics[width = .3\linewidth, height = .27\linewidth]{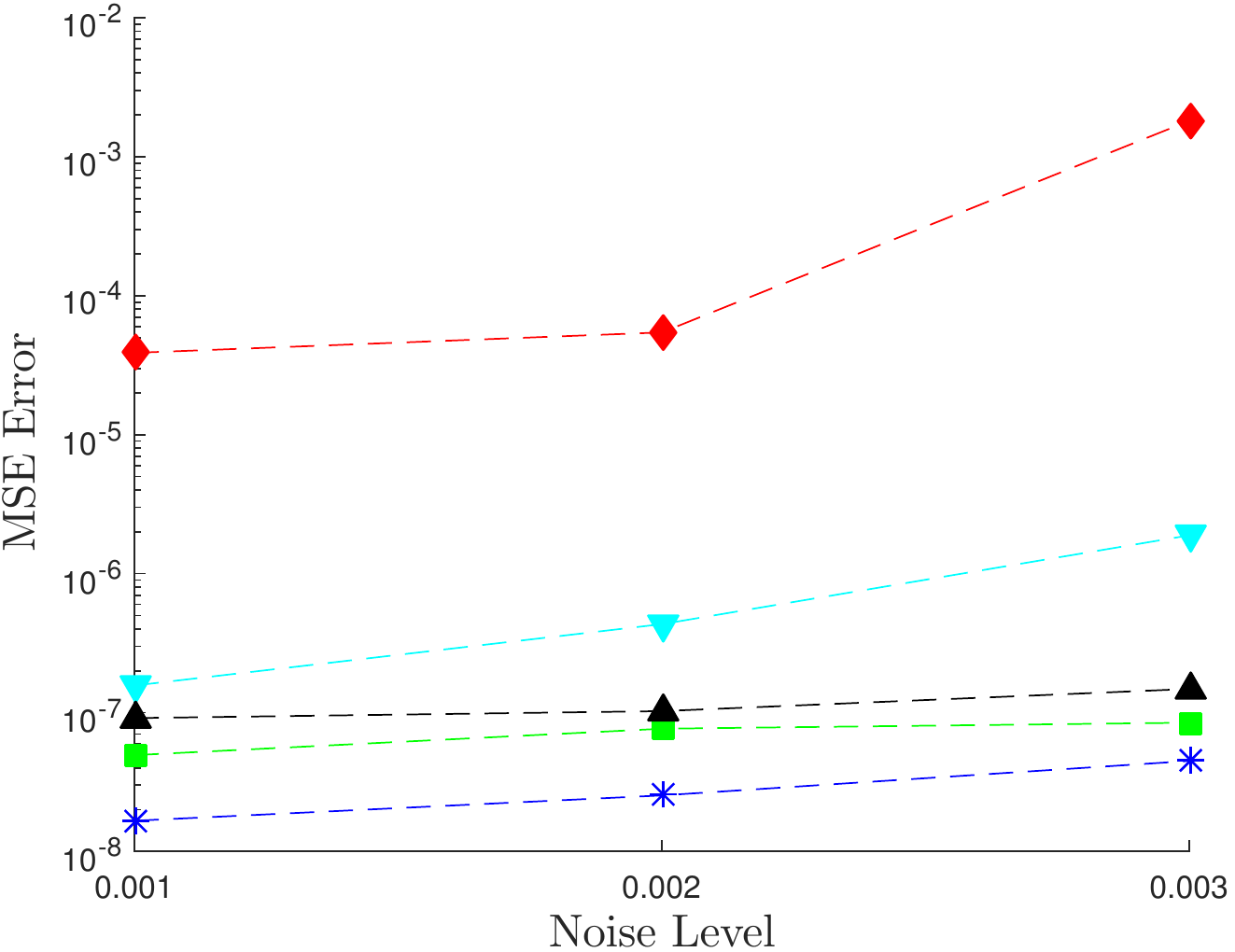}& \\
&Horse & &\\
\multicolumn{4}{c}{\includegraphics[width = .6\linewidth]{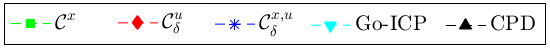}}

\end{tabular}
\end{center}
   \caption{MSE results for rigid registration with 3D data: same sampling (column 1), different sampling (column 2) and with added noise (column 3). Rows 1-4 give the error results for the Bunny, Dragon, Buddha and Horse meshes respectively.}
\label{fig:Rigid3d:ErrorResults}
\end{figure}

The MSE errors for the following  experiments are presented in Figure \ref{fig:Rigid3d:ErrorResults}. 
\begin{enumerate}
\item 


 The first column of Figure \ref{fig:Rigid3d:ErrorResults} presents the results of rigid registration when target and  model meshes have the same sampling w.r.t. different levels of rotation magnitude (reported in abscissa). At each level of rotation magnitude, 15 different pairs of shapes $S_1$ and $S_2$ were registered from which MSE is calculated. Correspondences are not used to enhance the registration process in this case and  overall CPD performs best, followed by Go-ICP and $\mathcal{C}_{\delta}^{u}$, while $\mathcal{C}_{\delta}^{x, u}$  and Jian et al's method $\mathcal{C}^{x}$ seem to generate similar results. Since there is a one to one correspondence between the samples from each shape, all methods  perform very well with an average MSE of around $10^{-34}$ for CPD and $10^{-12}$ in all other cases. Both CPD and Go-ICP have a tendency to fall into local minima as the rotation increases while $\mathcal{C}_{\delta}^{u}$, $\mathcal{C}_{\delta}^{x}$ and $\mathcal{C}_{\delta}^{x,u}$ continue to estimate good solutions.

\item 
Similarly the second experiment (reported in the second column of Figure \ref{fig:Rigid3d:ErrorResults}) 
considers the case  where target and model meshes do not have the same set of vertices but instead different samples of 1000 points were chosen from $S_1$ and $S_2$, along with their corresponding normal vectors, so that no one to one correspondence exist between the subsampled target and model shapes. 
 $\mathcal{C}_{\delta}^{x, u}$ performs the best in this case, followed by Jian et al's method $\mathcal{C}^{x}$. Here $\mathcal{C}_{\delta}^{u}$ performed the worst as unlike vertices, normal vectors represent the first derivative of the surface and are more sensitive to noise, thus varying more when they are not sampled at exactly the same locations on $S_1$ and $S_2$. Again both CPD and Go-ICP fall into local solutions as the rotation magnitude increases.

\item 
Our cost functions are assessed   when noise is present in the data (third column of Figure \ref{fig:Rigid3d:ErrorResults})) with  three levels  of Gaussian noise (mean zero and standard deviation varying from 0.001 to 0.003 reported in abscissa) applied to vertices of $S_2$, which differs from $S_1$ by a rotation of magnitude $30\degree$.  
  When computing the normal vectors of the noisy shape $S_2$ we used the nearest neighbours approach implemented by Meshlab, as described in Section \ref{sec:normComput}. We found that when noise is present in the point positions, this gives a better estimate of the normal vector than using the vertex connectivity. As the noise on the points $\{x_2^{(i)} \}$ increases we also increase the number of nearest neighbours ($N_k$) used to compute the normal vectors, for example we set $N_k = 40, 60$ and $120$ for noise levels $0.001, 0.002$ and $0.003$ respectively when registering two Bunny shapes. The normal vectors associated with the noise free shape $S_1$ were computed using the vertex and edge information provided in the .ply.
   Different samples of 1000 points, along with their associated normal vectors, were then chosen from $S_1$ and the noisy $S_2$, so that no one to one point correspondences exist between the target and model point clouds. The registration process was  repeated 15 times for each noise level, and  in all cases, $\mathcal{C}_{\delta}^{x,u}$ performs the best, followed by $\mathcal{C}^{x}$, CPD and Go-ICP. The additional smoothed normal vector information used by $\mathcal{C}_{\delta}^{x,u}$ allows it to converge to a more accurate solution, even when a large degree of noise is added to the points in the shape $S_1$. Again $\mathcal{C}_{\delta}^{u}$ does not perform as well as the other approaches. 
\end{enumerate}

\subsection{3D Non-rigid Registration}
\label{sec:3DnonRigid}
Finally we consider two 3D shapes $S_1$ and $S_2$ differ by a non-rigid deformation. We register these shapes by estimating a non-rigid TPS transformation. We choose the number of control points as $N = 125$ so our latent space has $(125 \times 3) + 12 = 387$ dimensions. Point correspondences are used here in the cost functions $\mathcal{C}^x$ and  $\mathcal{C}^{x,u}$, notated as $\mathcal{C}_{corr}^x$ and  $\mathcal{C}_{corr}^{x,u}$. Again we omit $\mathcal{C}^u$ and $\mathcal{C}_{\delta}^u$ as we found that they did not perform well when estimating a non-rigid transformation. We also omit $\mathcal{C}_{\delta}^{x,u}$ as it has previously been shown to perform similarly to $\mathcal{C}^{x,u}$.

 We present two sets of experiments in this section. In the first set we compare how $\mathcal{C}_{corr}^x$ and  $\mathcal{C}_{corr}^{x,u}$ perform when registering shapes with known correspondences and in the second we compare  $\mathcal{C}_{corr}^x$, $\mathcal{C}_{corr}^{x,u}$, CPD\cite{CPD2010} and GLMD\cite{GLMD2015} when registering shapes with unknown correspondences that must be estimated.

\begin{enumerate}

\item 
In this first experiment we use the dataset of shapes provided by Sumner et al. \cite{Sumner2004} which contains meshes of several different types of animal in different poses, including a cat, lion and horse. Each mesh of the same animal has an equal number of vertices and exact point correspondences. 
We use the ground truth point correspondences when computing $\mathcal{C}_{corr}^x$ and $\mathcal{C}_{corr}^{x,u}$ to reduce computational complexity.
  Choosing two meshes of the same type of animal, we let the vertices of each mesh be the points $\{ x_1^{(i)} \}$ and $\{ x_2^{(i)} \}$ and compute the corresponding normal vectors $\{ u_1^{(i)} \}$ and $\{ u_2^{(i)} \}$ using the edge information provided in the mesh. We then apply a rotation to $S_1$ so that the shapes differ by both a rotation and non-rigid deformation. For each level of rotation tested we register 10 pairs of shapes $S_1$ and $S_2$.
    Figure \ref{fig:3DnonRigid:MSE}(a) reports MSE comparing $\mathcal{C}_{corr}^x$ and $\mathcal{C}_{corr}^{x,u}$: due to the large dimension of the latent space (387 dimensions), the gradient ascent technique required a large number of iterations to register the shape $S_1$ to $S_2$. For each cost function, to reduce computation time the  limit of on the number of function evaluations computed during optimization is set to 50,000  (at each simulated annealing step). 
  Very little difference  is observed between the cost functions and  even with 50,000 functions evaluations, both $\mathcal{C}_{corr}^x$ and $\mathcal{C}_{corr}^{x,u}$ failed to converge to a good solution.

\item  

\begingroup
\setlength{\tabcolsep}{9pt} 
\begin{figure*}[!t]
\begin{center}
\begin{tabular}{c c c}

\includegraphics[width = .3\linewidth]{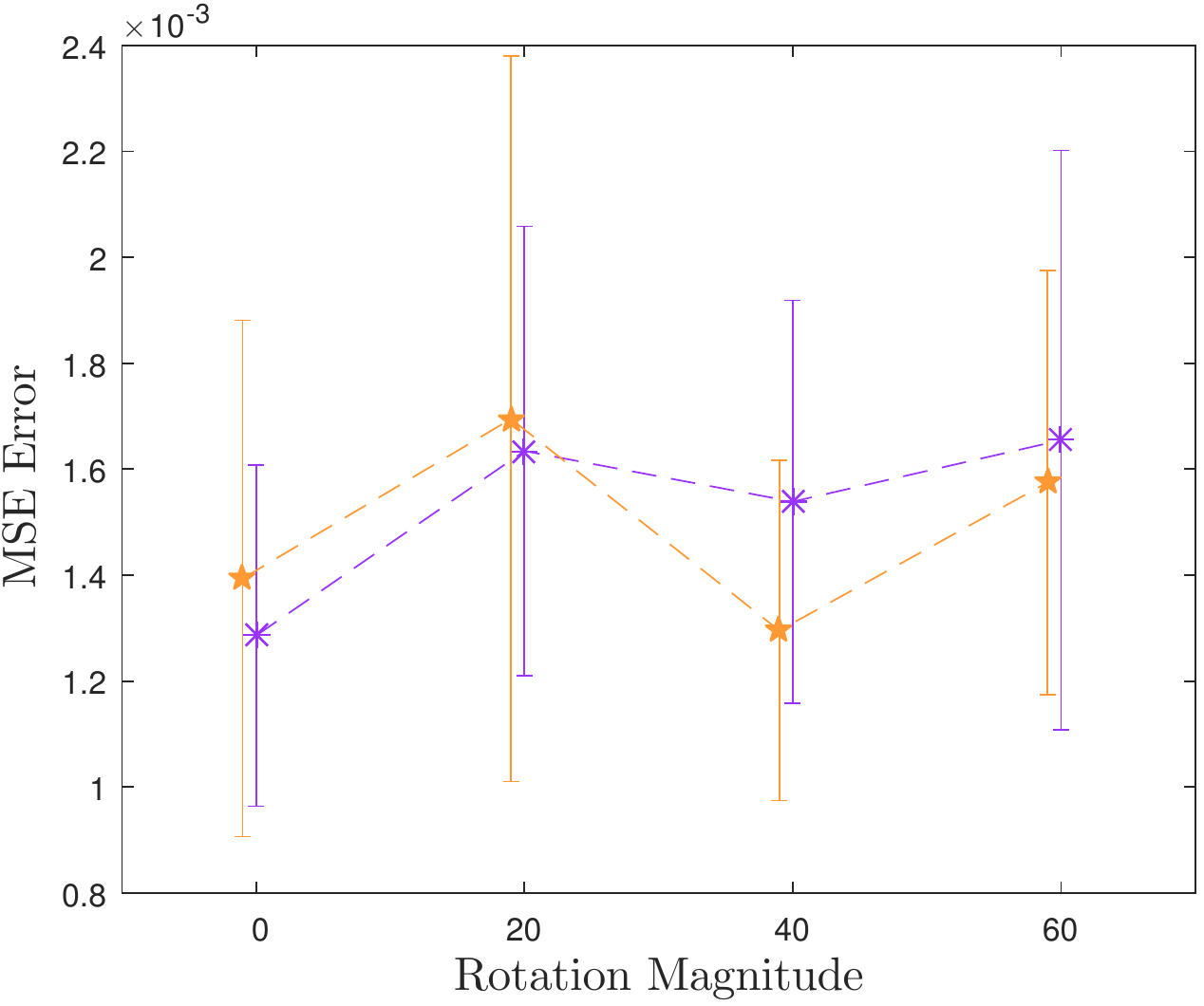}&
\includegraphics[width = .3\linewidth]{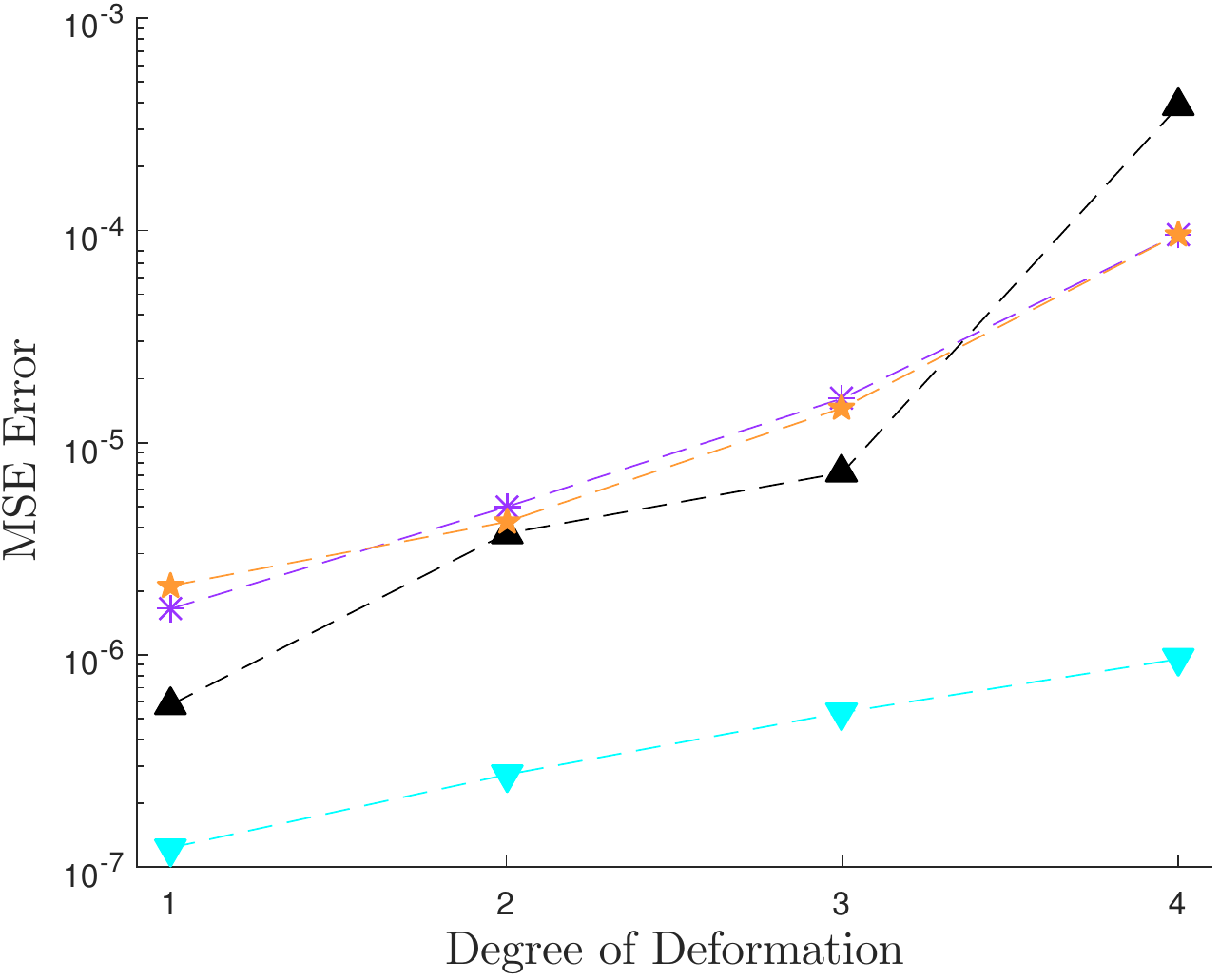}&
\includegraphics[width = .3\linewidth]{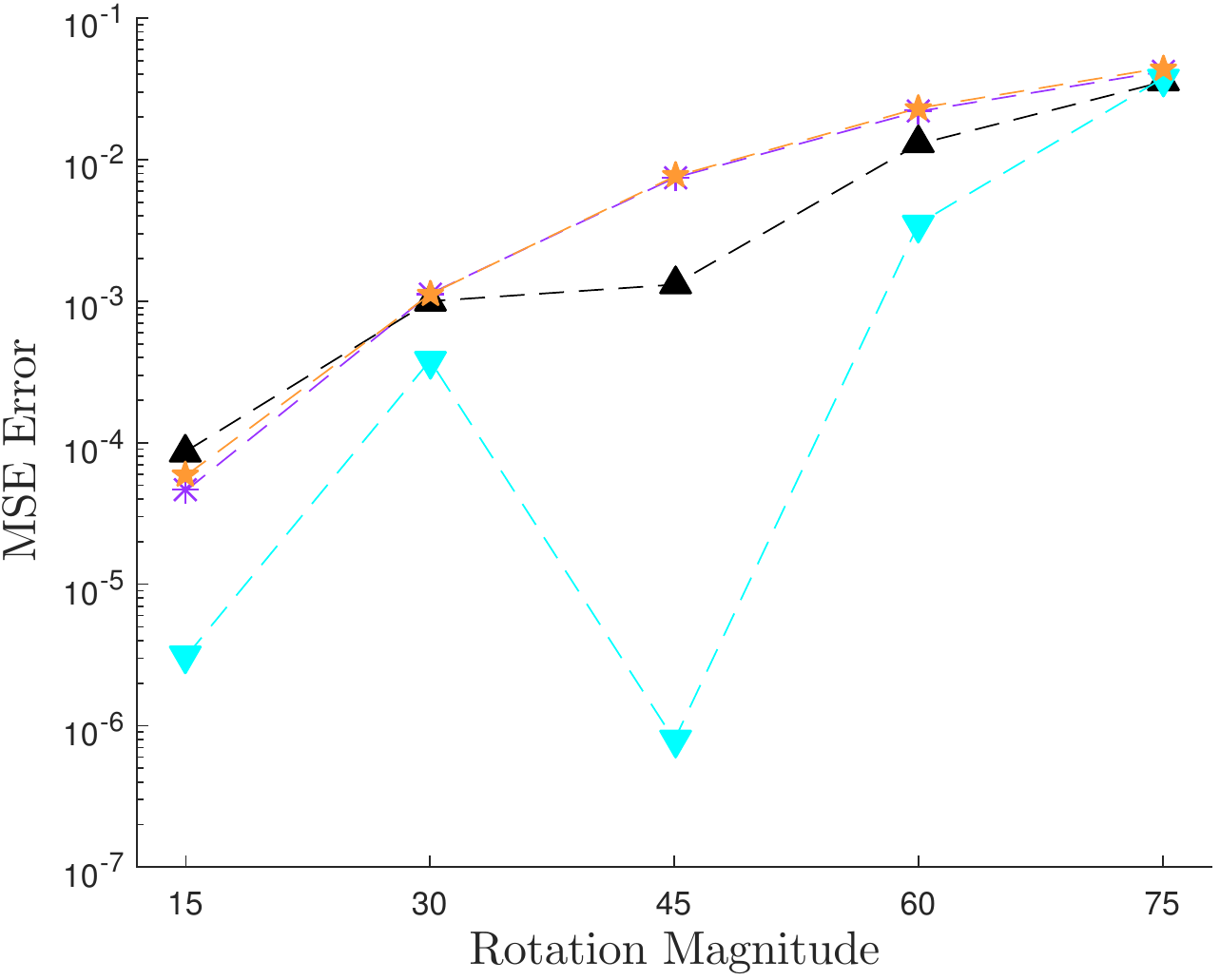}\\
\multicolumn{3}{c}{\includegraphics[width = .5\linewidth]{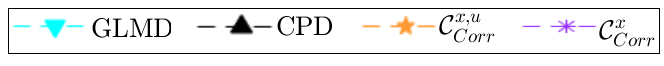}} \\
(a)& (b) & (c) \\

\end{tabular}
\end{center}
   \caption{MSE results for non-rigid registration with 3D data. (a) Comparison between $\mathcal{C}_{corr}^x$ and $\mathcal{C}_{corr}^{x,u}$ when registering meshes with exact correspondences. The meshes differ by a deformation and rotation varying from $0 \degree$ to $60 \degree$. The standard error bars are included and emphasise the similarity between the cost functions; (b) Non-rigid transformation estimation between bunny shapes differing by a deformation varying from degree 1 to 4; (c) Non-rigid transformation estimation when two bunny shapes differ by a deformation of degree 3 and rotation varying from $15 \degree$ to $75 \degree$.}
\label{fig:3DnonRigid:MSE}
\end{figure*}
\endgroup
In this experiment a scan taken of the Stanford Bunny with 1000 points is used  to generate $S_1$ and $S_2$. Taking the points of the scan to be $\{ x_2^{(i)} \}$, we computed the normals vectors $\{ u_2^{(i)}\}$ using the nearest neighbour approach discussed in Section \ref{sec:normComput}. Then using the same deformation technique proposed by Yang et al. \cite{GLMD2015} and implemented in Section \ref{sec:normComput}, we used $9$ control points on the boundary of the points $\{ x_2^{(i)} \}$ to deform them, generating the points $\{x_1^{(i)}\}$. Again the normal vectors $\{ u_1^{(i)} \}$ were computed using the nearest neighbour approach. 

For cost functions $\mathcal{C}_{corr}^x$ and $\mathcal{C}_{corr}^{x,u}$,  we estimate the point correspondences using the method proposed by Yang et al.\cite{GLMD2015} and detailed in Section \ref{sec:shape:corr}. We test 4 levels of deformation and register 15 pairs of shapes at each level. We also test the case in which $S_1$ and $S_2$ differ by a rotation and non-rigid deformation by applying a rotation to the shape $S_1$. We set the level of deformation to $3$ and test 5 levels of rotation $(15 \degree, 30 \degree, 45 \degree, 60 \degree, 75 \degree)$, with 15 pairs of shapes registered for each rotation. 

 The MSE results can be seen in Figure \ref{fig:3DnonRigid:MSE}(b) and \ref{fig:3DnonRigid:MSE}(c).
In Figure\ref{fig:3DnonRigid:MSE}(b), for all degrees of deformation applied to the model shape $S_1$, GLMD performs the best, while $\mathcal{C}_{corr}^x$ and $\mathcal{C}_{corr}^{x,u}$ perform similarly. Although we found that the correspondences estimated by Yang et al's technique and used by $\mathcal{C}_{corr}^x$ and $\mathcal{C}_{corr}^{x,u}$ were accurate, using only $125$ control points for the estimated TPS transformation limited the accuracy of both $\mathcal{C}_{corr}^x$ and $\mathcal{C}_{corr}^{x,u}$ in comparison to GLMD, which uses all 1000 points in $S_1$ as control points. However, increasing the number of control points used by $\mathcal{C}_{corr}^x$ and $\mathcal{C}_{corr}^{x,u}$ also increases the dimension of the latent space, requiring a larger number of iterations to converge to a good solution. 
  
 The results of registering Bunny shapes differed by both a non rigid deformation and rotation can be seen in Figure \ref{fig:3DnonRigid:MSE}(c). In this case we found that the correspondences estimated by Yang et al's technique had some errors due to the rotation difference between the shapes. This decreased the accuracy of both GLMD and the cost functions $\mathcal{C}_{corr}^x$ and $\mathcal{C}_{corr}^{x,u}$, although GLMD still performed the best. Although $\mathcal{C}_{corr}^{x,u}$ typically performs well when the shapes differ by a rotation, when the wrong point correspondences are used the accuracy of $\mathcal{C}_{corr}^{x,u}$ is reduced. Again we found that using only $125$ control points also reduced the accuracy achievable by $\mathcal{C}_{corr}^x$ and $\mathcal{C}_{corr}^{x,u}$. 
\end{enumerate}

\subsection{Computation time}
\label{sec:shape:compTime}

In Table \ref{tab:Shape:compTime100Iter} we present the computation times needed by the proposed cost functions to carry out 10 iterations of the gradient ascent algorithm used to register two shapes $S_1 = \lbrace (x_1^{(i)},u_1^{(i)})\rbrace_{i=1,\cdots,100}$ and  $S_2 = \lbrace (x_2^{(j)},u_2^{(j)})\rbrace_{j=1,\cdots,100}$, each with $100$ points and unit normal vectors. In Table \ref{tab:Shape:numIter} we give the average number of iterations needed by each cost function to converge to the correct solution. These figures were computed when using our full annealing strategy, with the number of annealing steps used given in column 5 of Table \ref{tab:Shape:numIter}. In Table \ref{tab:Shape:compTimesSOA} we also present the computation times needed by the CPD, Go ICP and GLMD algorithms to register shapes $S_1$ and $S_2$. 

\begin{table*}[!t]
\begin{center}
 \begin{tabular}{|c|c|c|c|c|c|c|c|c|c|}
 \hline
 & ctrl pts& dim & $n_1$ & $n_2$ & $\mathcal{C}^x$ & $\mathcal{C}^u$& $\mathcal{C}_{\delta}^{u}$& $\mathcal{C}^{x,u}$& $\mathcal{C}_{\delta}^{x,u}$ \\ \hline
 2D Rotation & \ding{55}  & 1 & 100 & 100 & 0.20s  & 0.20s & 0.16s & 0.21s  & 0.20s  \\
  3D Rotation & \ding{55}  & 9 & 100 & 100 & 0.22s & 0.27s & 0.29s & 0.30s & .4695 \\
   2D TPS & 12 & 30 & 100 & 100 & 2.2s & \ding{55} &  \ding{55}  & 2.9s   &  \ding{55}  \\
    3D TPS & 125 & 387 & 100 & 100 & 16s &  \ding{55}  &  \ding{55}  & 30s &  \ding{55}  \\
 \hline
 \end{tabular}
 \caption{The time taken by each of the cost functions to compute 100 iterations of the gradient ascent algorithm. In each case the shapes $S_1$ and $S_2$ have 100 points each. The number of control points used by the TPS functions is shown in column 2 and column 3 gives the dimension of the latent space in each case.}
    \label{tab:Shape:compTime100Iter}
\end{center}
\end{table*}

\begin{table*}[!h]
\begin{center}
 \begin{tabular}{|c|c|c|c|c|c|c|c|c|c|}
 \hline
 & dim & $n_1$ & $n_2$ & Ann Steps &  $\mathcal{C}^x$ & $\mathcal{C}^u$& $\mathcal{C}_{\delta}^{u}$& $\mathcal{C}^{x,u}$& $\mathcal{C}_{\delta}^{x,u}$ \\ \hline
 2D Rotation & 1 & 100 & 100 & 6 & 50  & 43 & 50 & 40  & 48  \\
  3D Rotation & 9  & 100 & 100 & 8 & 220 & 275 & 240 & 390 & 400 \\
   2D TPS & 30 & 100 & 100 &5 & 1370& \ding{55} &  \ding{55}  & 1500   &  \ding{55}  \\
    3D TPS & 387 & 100 & 100 & 8 & 880* &  \ding{55}  &  \ding{55}  & 880* &  \ding{55}  \\
 \hline
 \end{tabular}
 \caption{ The number of iterations typically taken by each algorithm to register two pointclouds with 100 points each. These figures are computed using our full simulated annealing strategy, with the number of simulated annealling steps used given in column 5. *Note that due to the high dimension of the latent space, we limited the number of function evaluations in this case, thus limiting the number of iterations allowed. Although a good solution was reached after this many iterations, the cost functions still had not fully converged. }
    \label{tab:Shape:numIter}
\end{center}
\end{table*}

\begin{table*}[!h]
\begin{center}
 \begin{tabular}{|c|c|c|c|c|c|}
 \hline
 & $n_1$ & $n_2$ & Go ICP & CPD & GLMD \\ \hline
  3D Rotation & 100 & 100 & 0.78s & 32s & \ding{55}\\
   2D TPS & 100 & 100 & \ding{55} & 0.09s  & 0.13s \\
    3D TPS& 100 & 100 & \ding{55} & 0.05s & 0.12s \\
 \hline
 \end{tabular}
 \caption{The time taken, on average, by the Go ICP, CPD and GLMD methods to converge to the correct solution. For CPD, the MSE tolerance chosen for 3D rotation, 2D TPS and 3D TPD registration was the same as that used in the demo code provided by authors. It was set to $e^{-8}$ for 3D rotation, $e^{-8}$ for 2D TPS and $e^{-3}$ for 3D TPS, hence the difference in computation times.   }
    \label{tab:Shape:compTimesSOA}
\end{center}
\end{table*}

\section{Shape Registration and Interpolation (Qualitative Experiments)}

\begin{figure}[h]
\begin{center}
\begin{tabular}{ c c c c}
Model & Target & $\mathcal{C}^{x}$ &$\mathcal{C}^{x,u}$ \\
\hline
\\[-10pt]


\includegraphics[width = .2\linewidth,height = .12\linewidth]{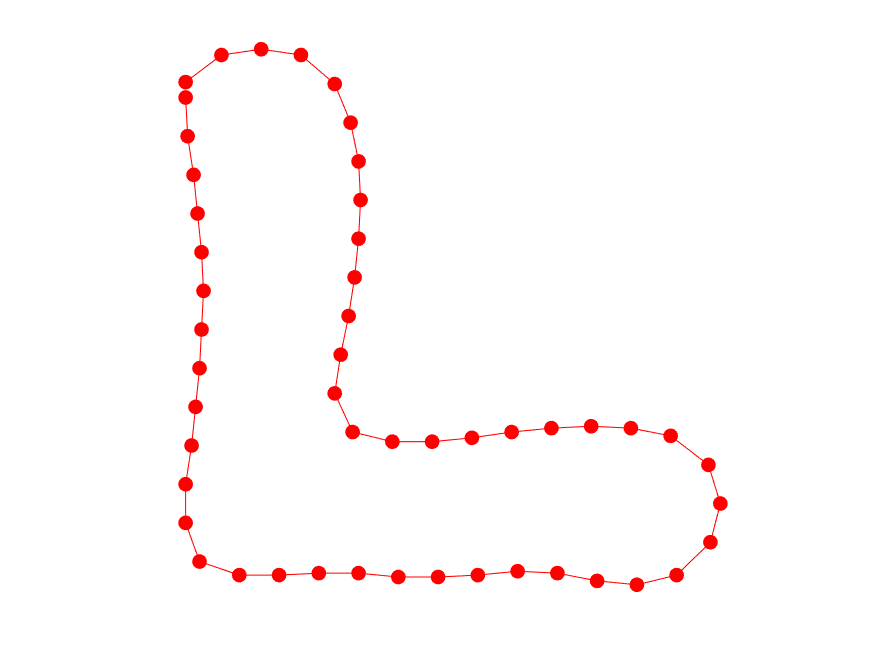}  & 

\includegraphics[width = .2\linewidth,height = .12\linewidth]{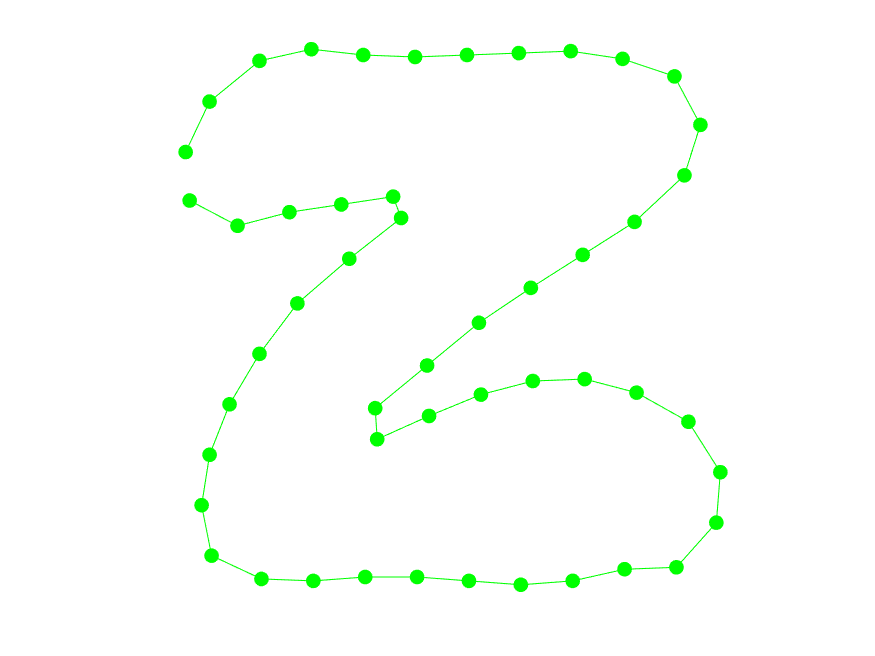}  & 

\includegraphics[width = .2\linewidth,height = .12\linewidth]{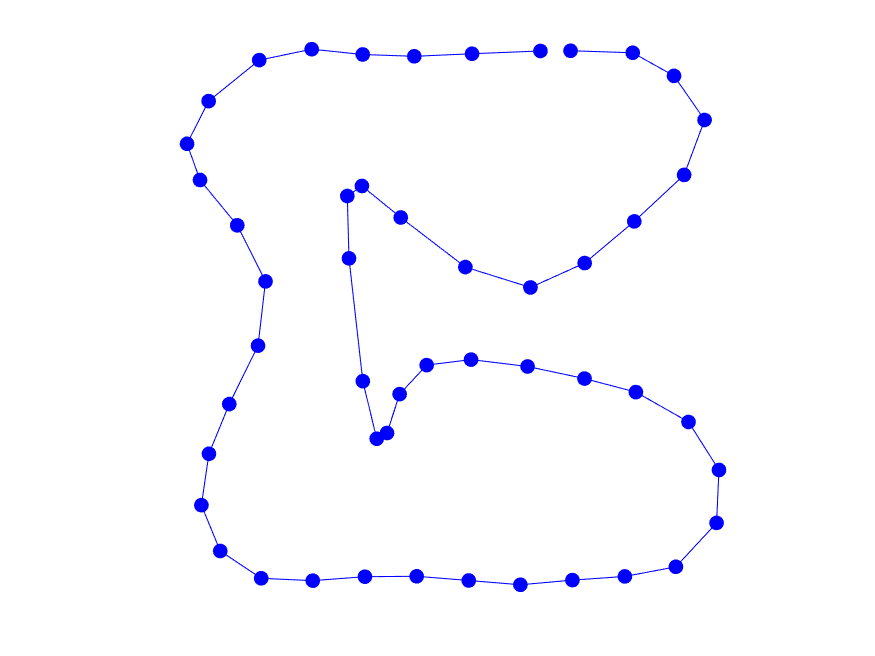}  & 

\includegraphics[width = .2\linewidth,height = .12\linewidth]{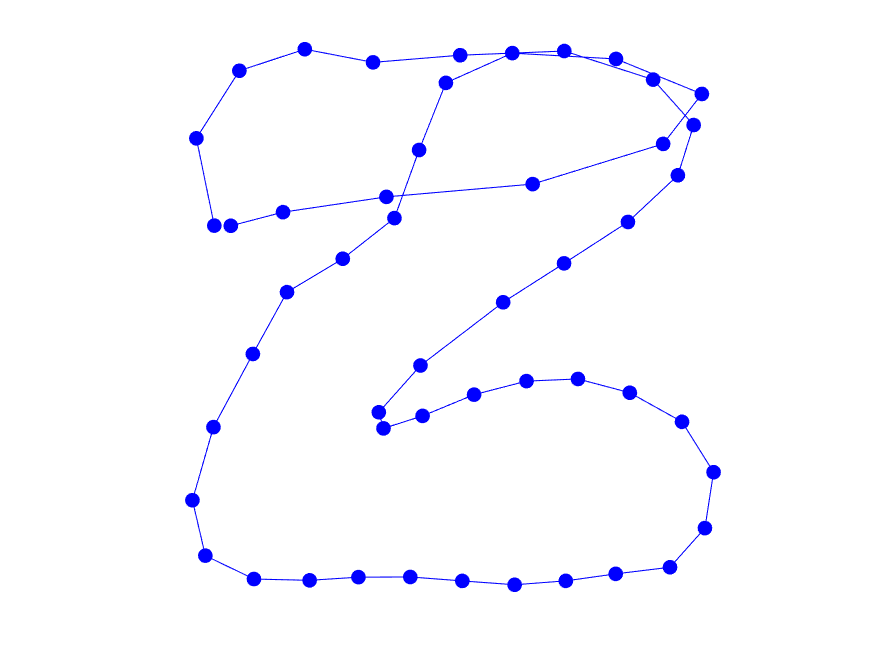} \\

\includegraphics[width = .15\linewidth,height = .12\linewidth]{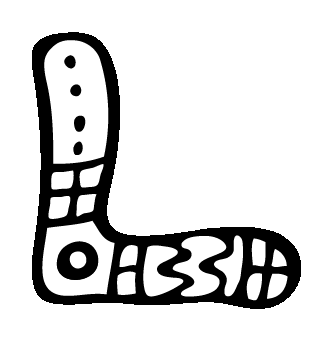}  &

\includegraphics[width = .15\linewidth,height = .12\linewidth]{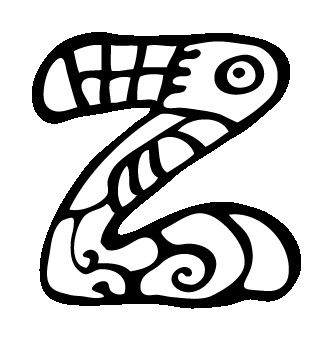}  &

\includegraphics[width = .2\linewidth,height = .12\linewidth]{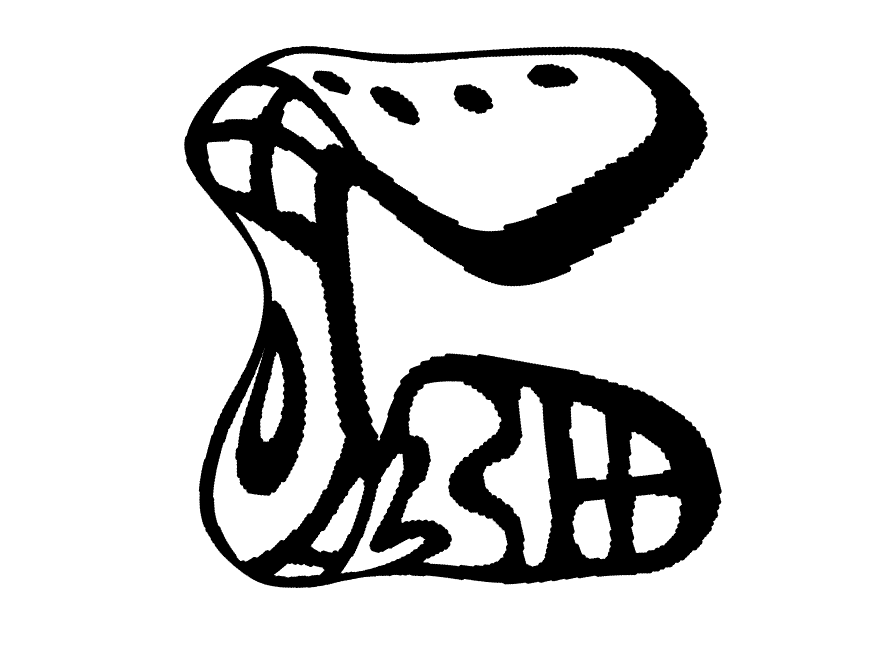}  & 

\includegraphics[width = .2\linewidth,height = .12\linewidth]{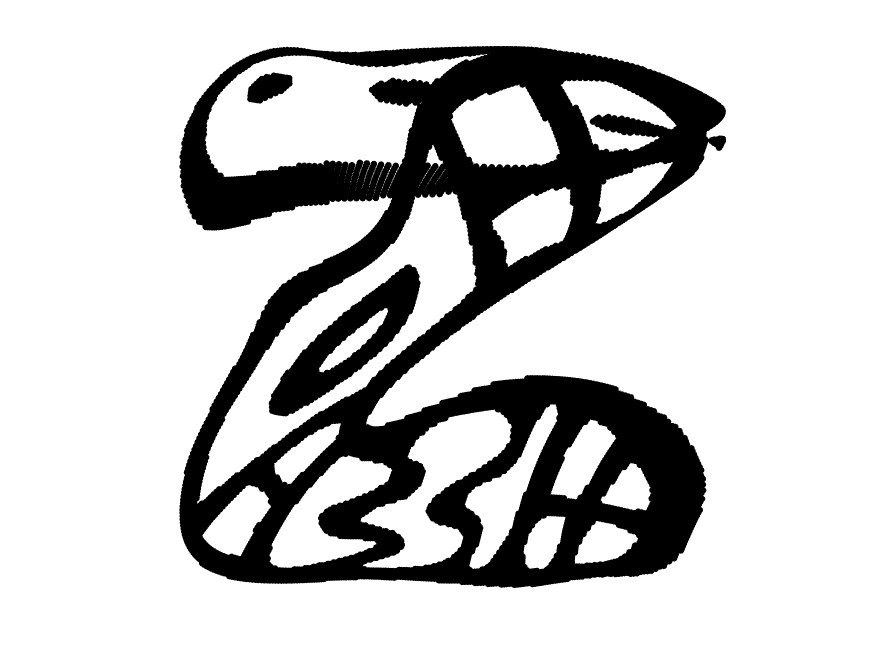} \\

\hline
\\[-10pt]

\includegraphics[width = .2\linewidth,height = .12\linewidth]{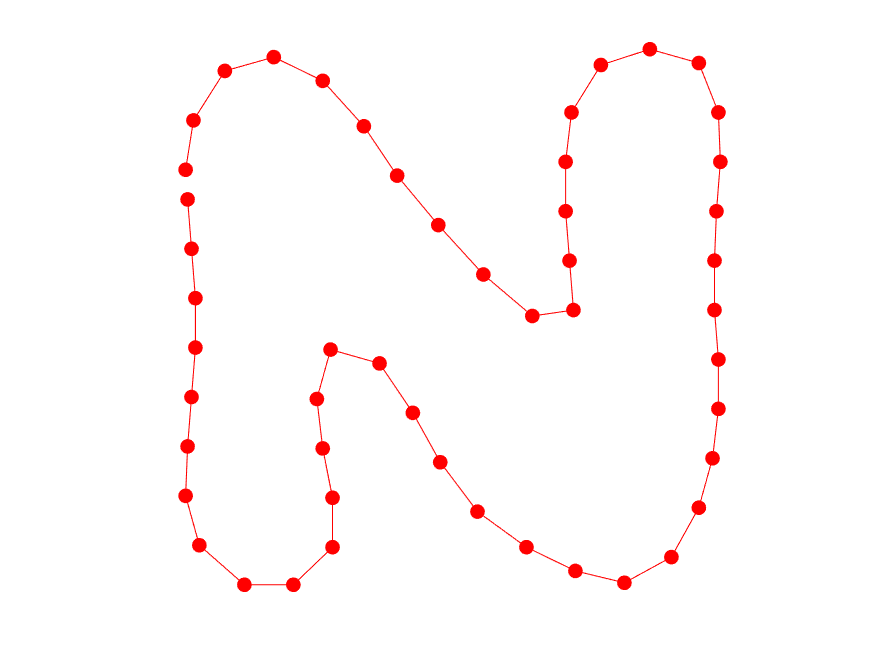}  & 

\includegraphics[width = .2\linewidth,height = .12\linewidth]{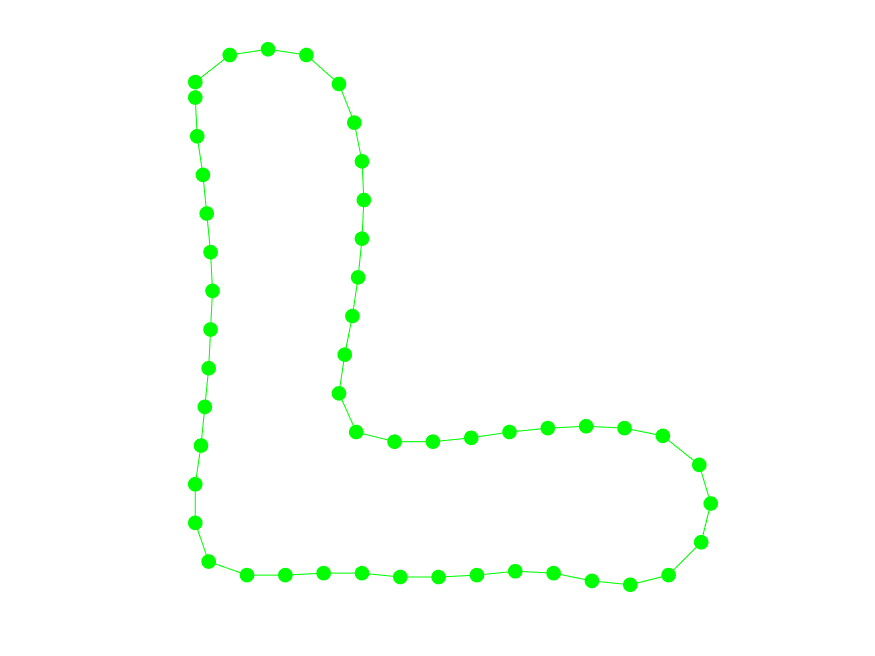}  & 

\includegraphics[width = .2\linewidth,height = .12\linewidth]{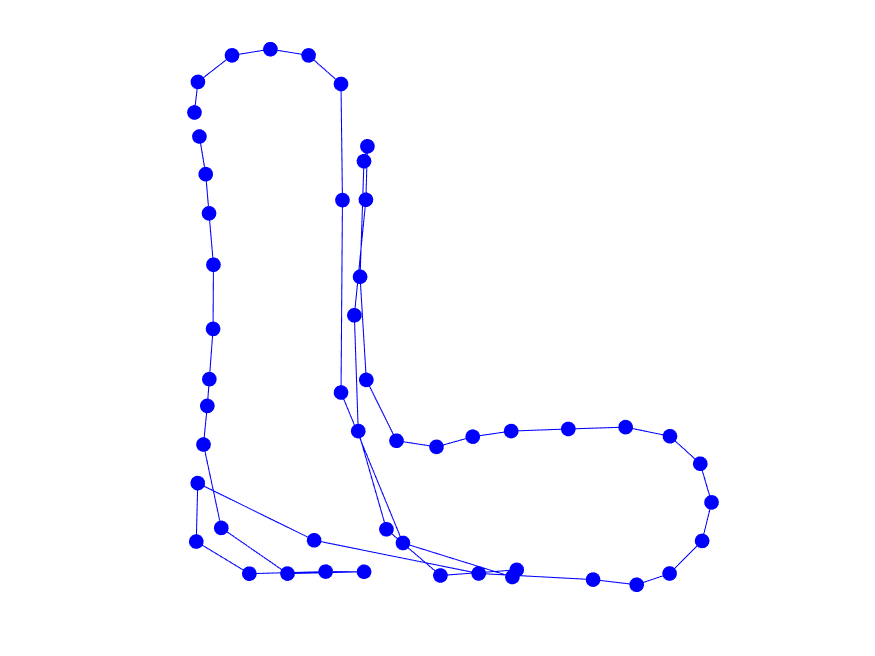}  & 

\includegraphics[width = .2\linewidth,height = .12\linewidth]{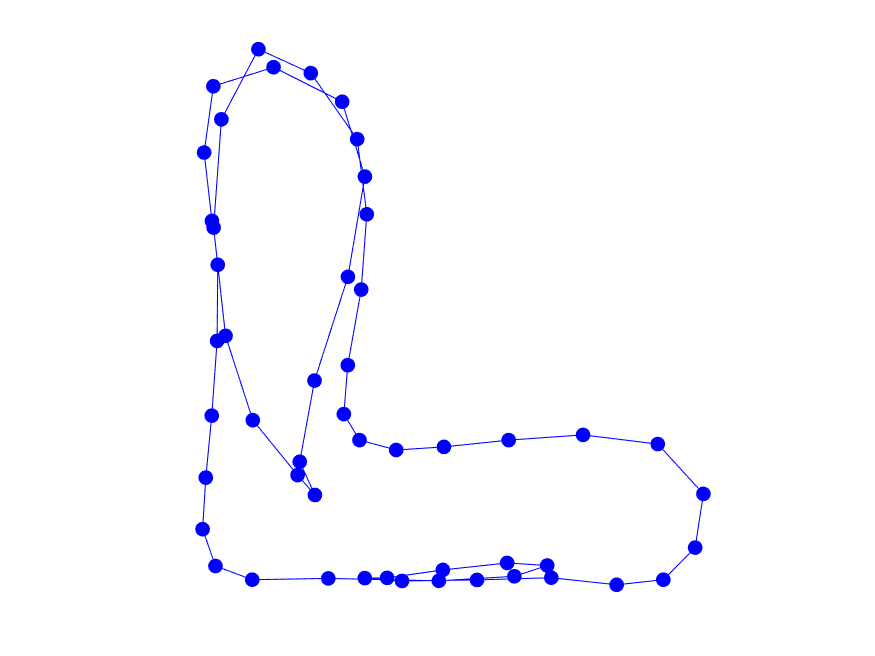} \\

\includegraphics[width = .15\linewidth,height = .12\linewidth]{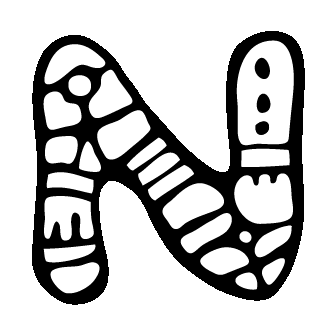}  & 

\includegraphics[width = .15\linewidth,height = .12\linewidth]{NewRegistration/Aboriginal/original/aboriginal_stroke07.png}  &

\includegraphics[width = .2\linewidth,height = .12\linewidth]{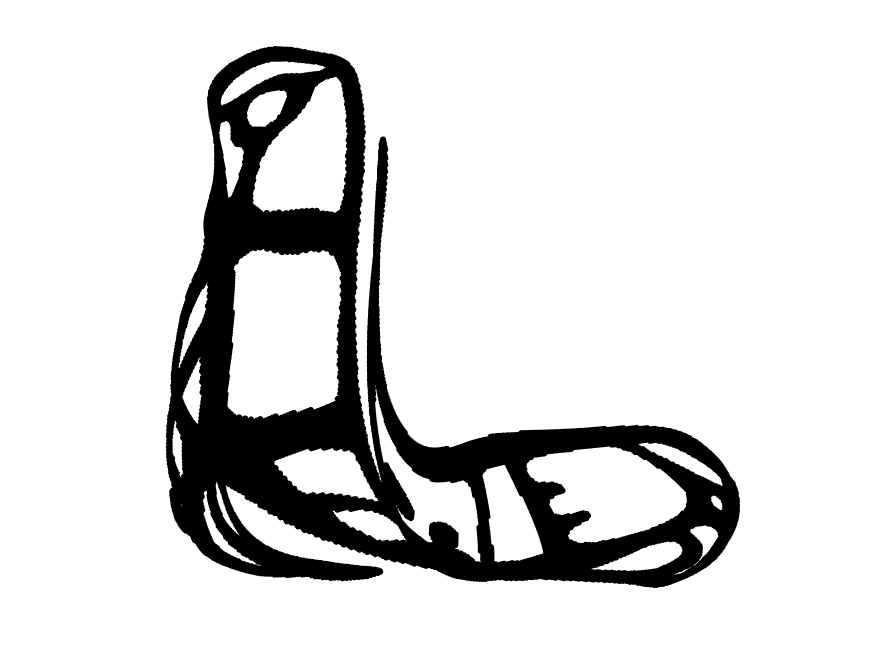}  & 

\includegraphics[width = .2\linewidth,height = .12\linewidth]{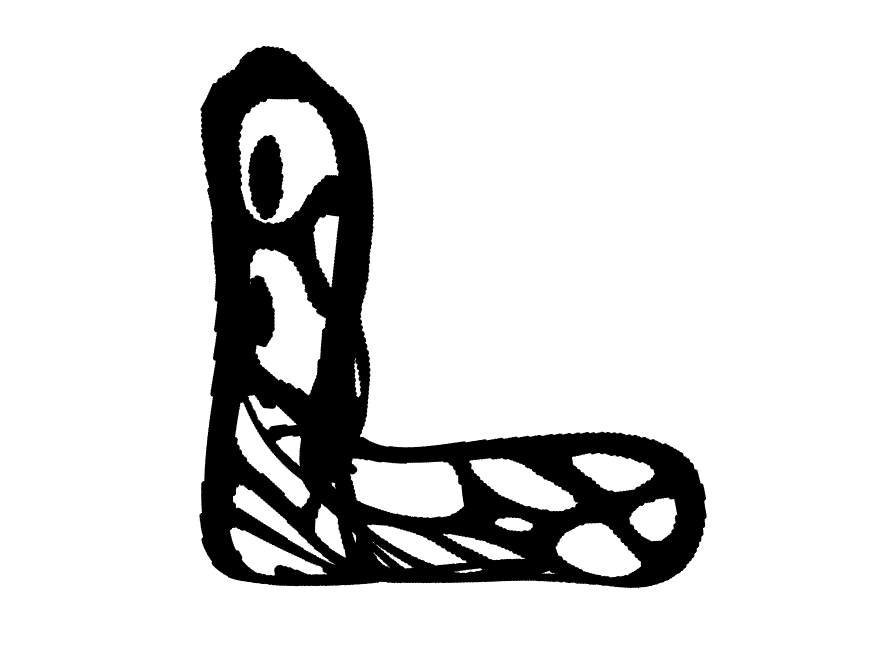} \\


\hline
\\[-10pt]

\includegraphics[width = .2\linewidth,height = .12\linewidth]{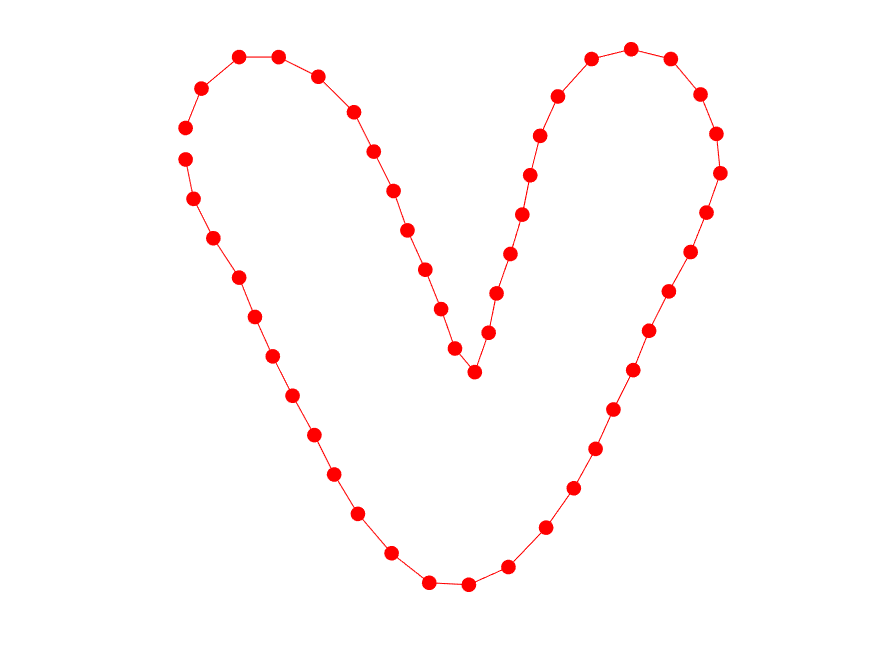}  & 

\includegraphics[width = .2\linewidth,height = .12\linewidth]{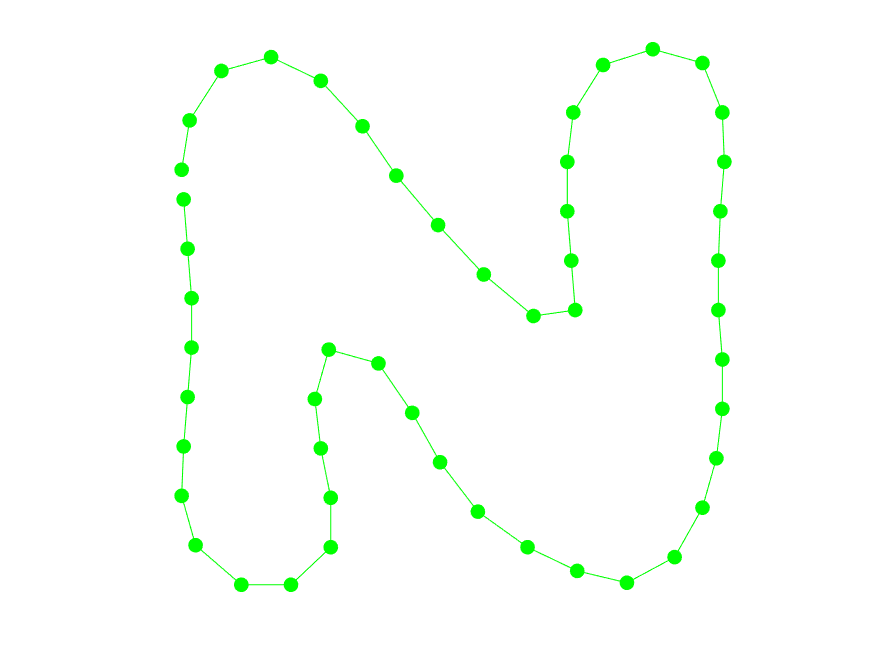}  & 

\includegraphics[width = .2\linewidth,height = .12\linewidth]{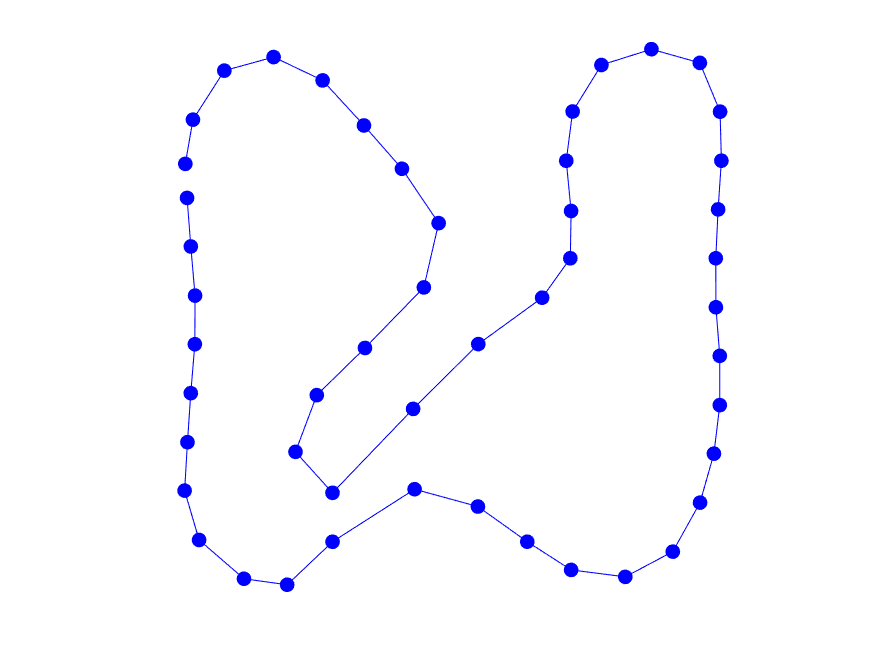}  & 

\includegraphics[width = .2\linewidth,height = .12\linewidth]{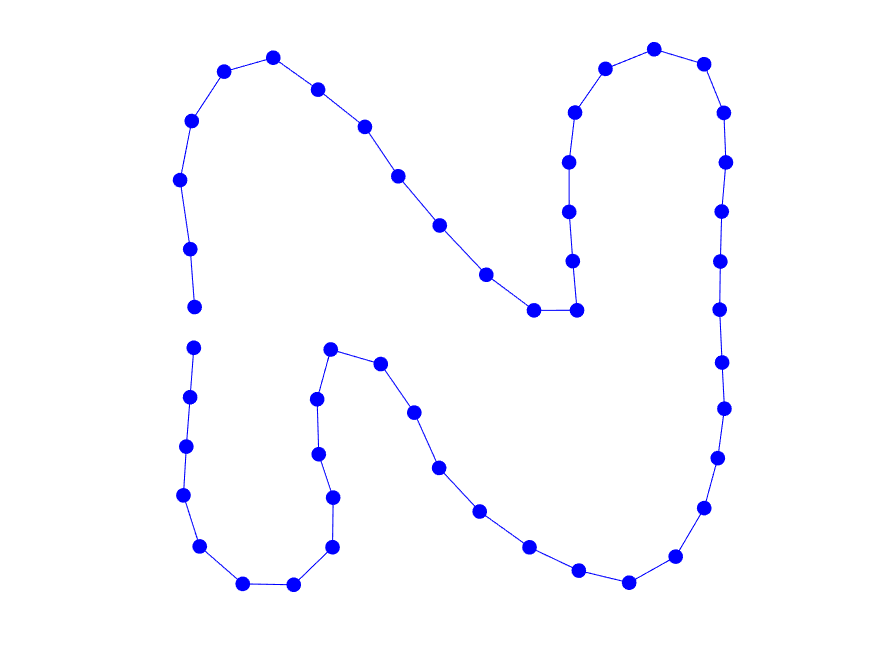} \\

\includegraphics[width = .15\linewidth,height = .12\linewidth]{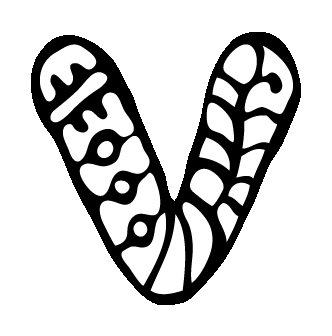}  & 

\includegraphics[width = .15\linewidth,height = .12\linewidth]{NewRegistration/Aboriginal/original/aboriginal_stroke04.png}  & 

\includegraphics[width = .2\linewidth,height = .12\linewidth]{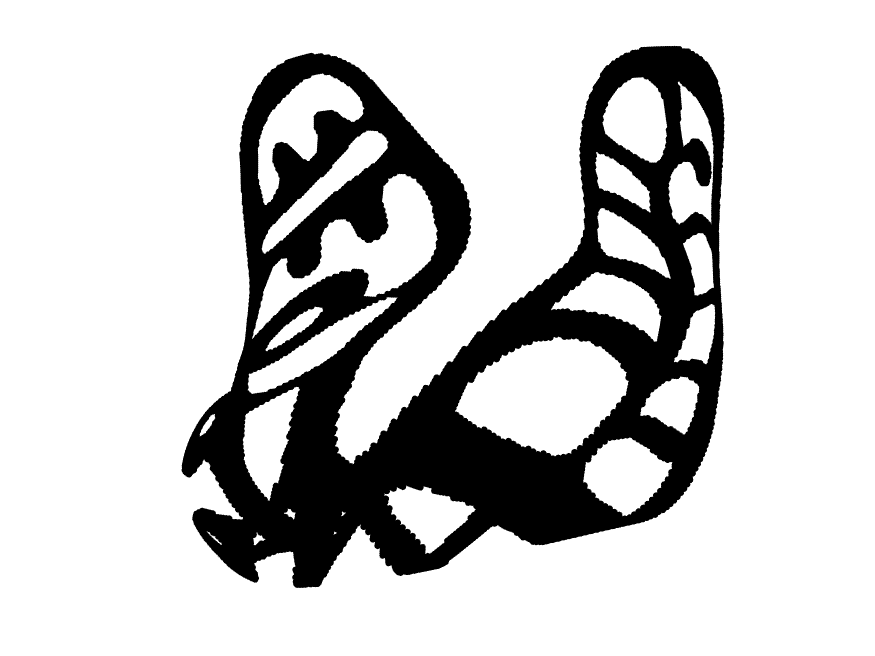}  & 

\includegraphics[width = .2\linewidth,height = .12\linewidth]{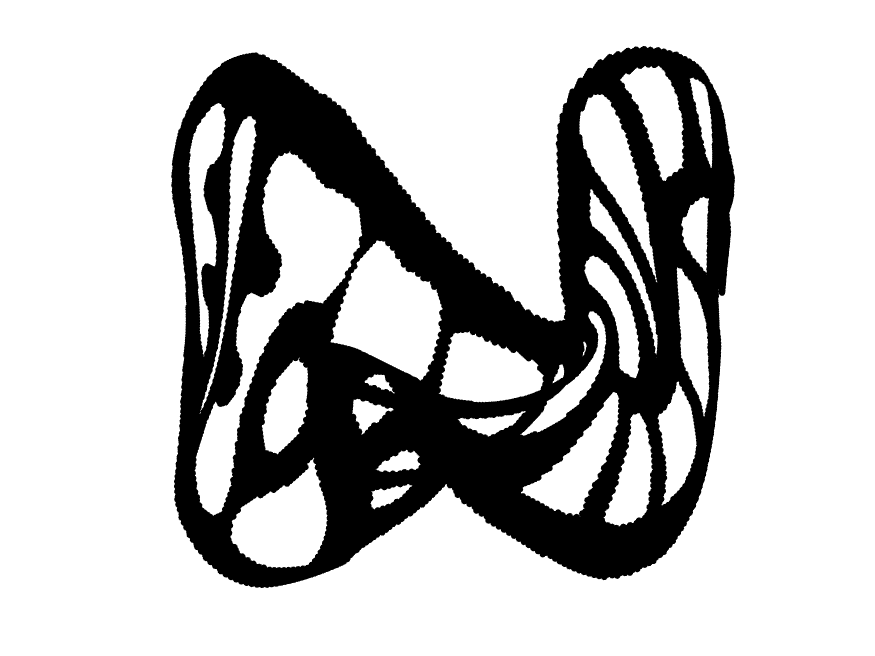} \\

\hline
\\[-10pt]

\includegraphics[width = .2\linewidth,height = .12\linewidth]{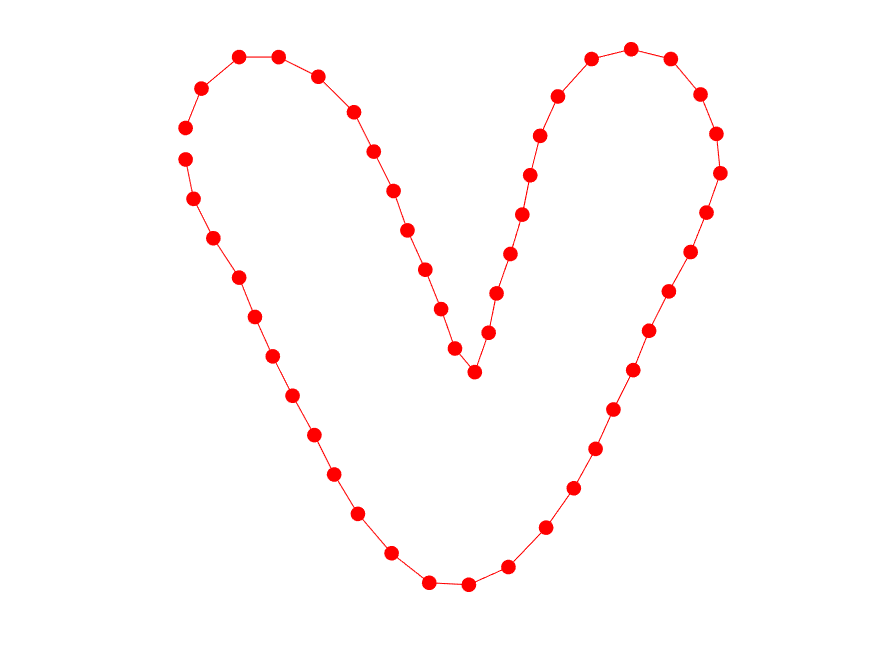}  & 

\includegraphics[width = .2\linewidth,height = .12\linewidth]{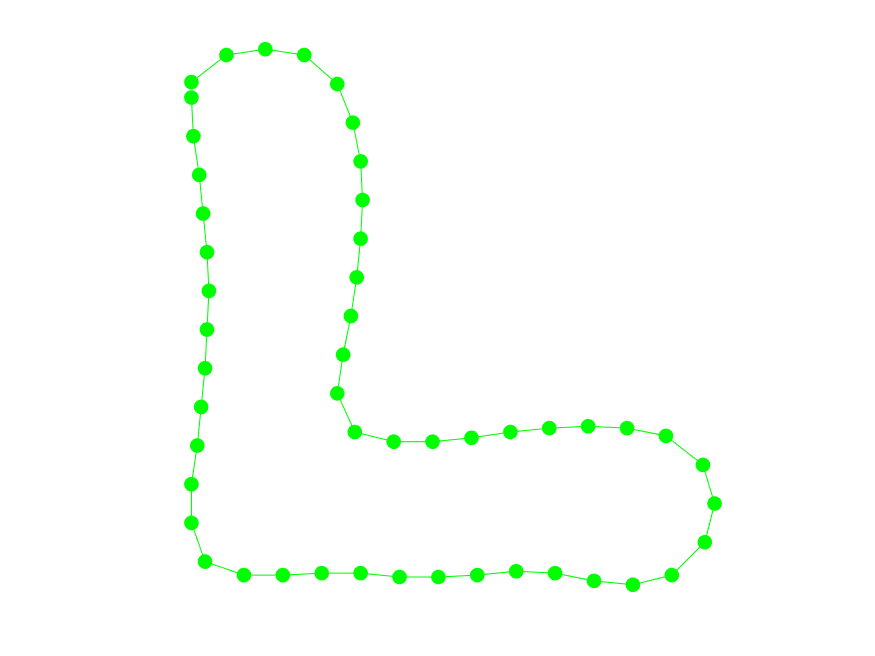}  & 

\includegraphics[width = .2\linewidth,height = .12\linewidth]{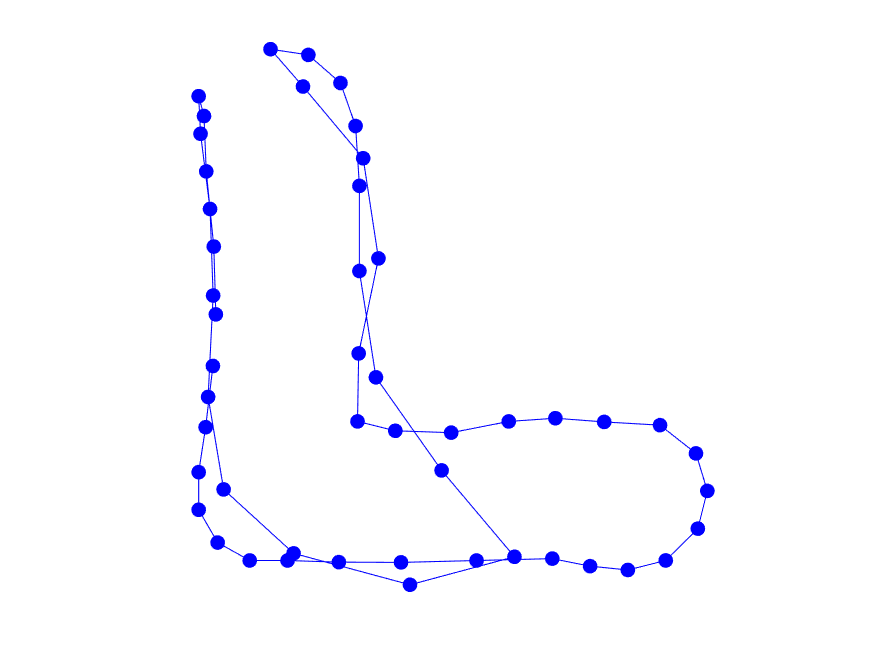}  & 

\includegraphics[width = .2\linewidth,height = .12\linewidth]{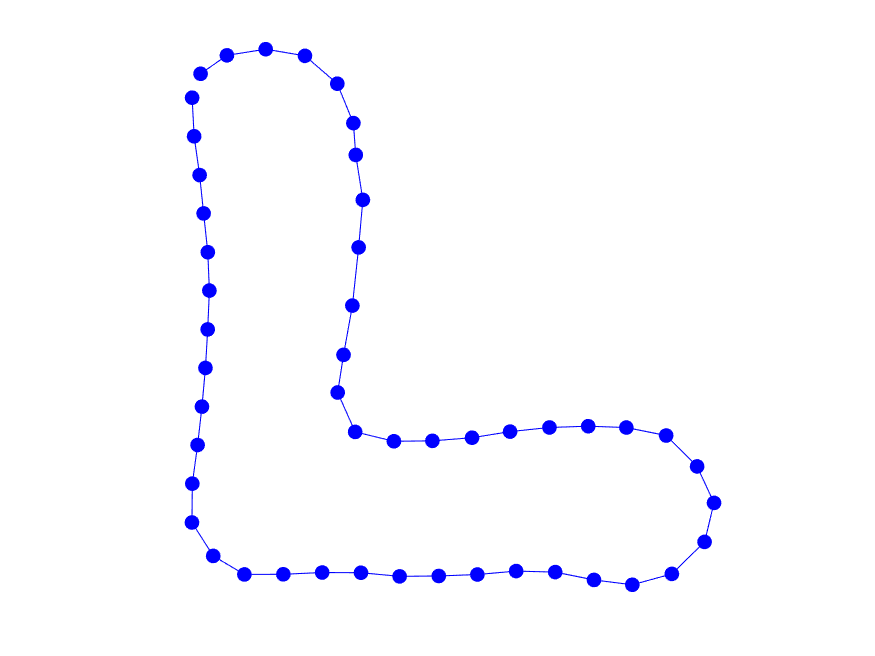} \\

\includegraphics[width = .15\linewidth,height = .12\linewidth]{NewRegistration/Aboriginal/original/aboriginal_stroke00.png}  &

\includegraphics[width = .15\linewidth,height = .12\linewidth]{NewRegistration/Aboriginal/original/aboriginal_stroke07.png}  &

\includegraphics[width = .2\linewidth,height = .12\linewidth]{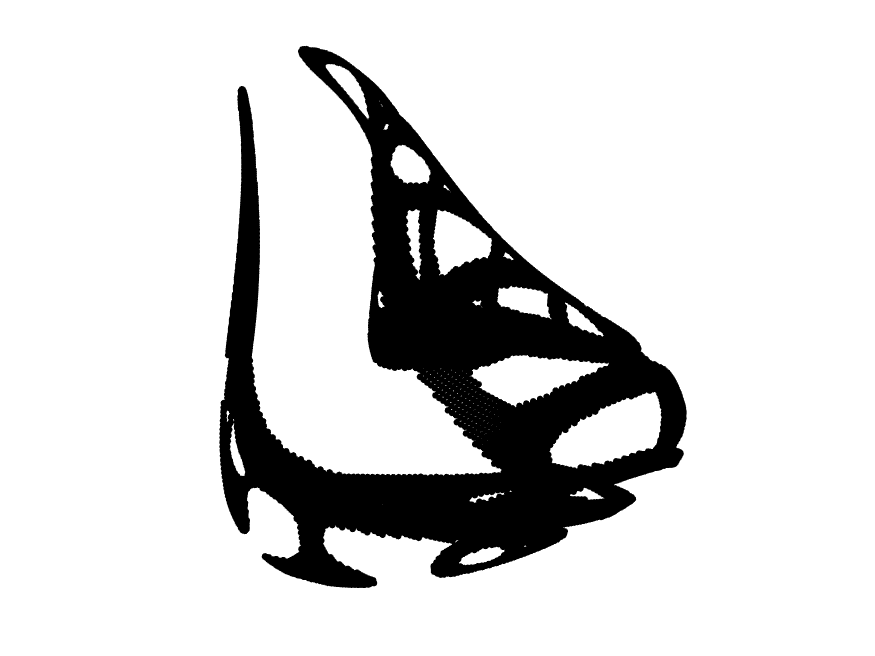}  & 

\includegraphics[width = .2\linewidth,height = .12\linewidth]{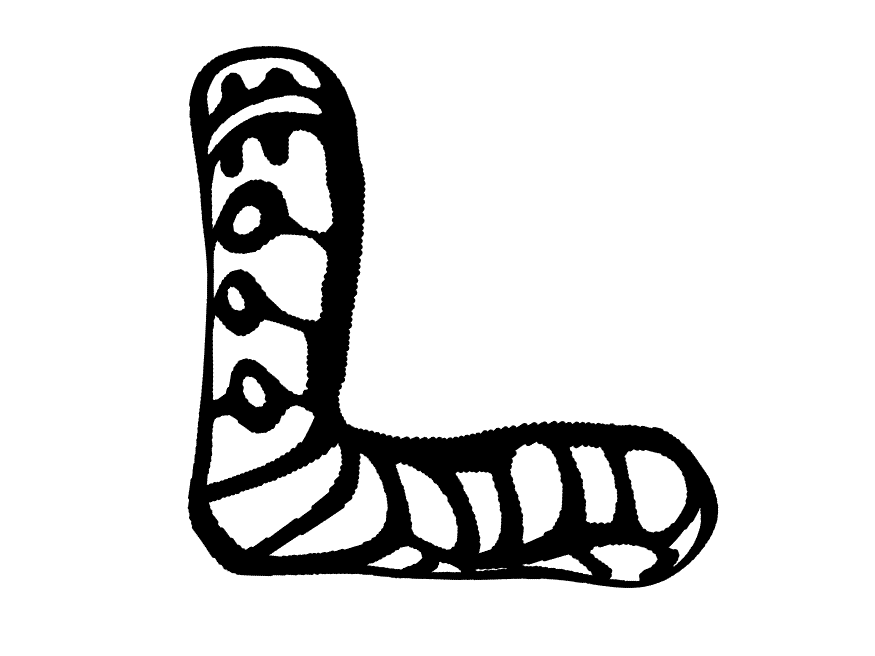} \\

\hline

\end{tabular}
\end{center}
\caption{In rows 1, 3, 5 and 7 we present the 50 points extracted along the external boundary of the model (col. 1) and target (col. 2) letters, and the transformation results estimated using $\mathcal{C}^{x}$ (col. 3) and $\mathcal{C}^{x,u}$ (col. 4). The connectivity information between points is shown, and is used when computing the normal vectors for $\mathcal{C}^{x,u}$. In rows 2, 4, 6 and 8 we show the patterned model letter (col. 1) and target letter (col. 2), and the transformed model letters estimated using $\mathcal{C}^{x}$ (col. 3) and $\mathcal{C}^{x,u}$ (col. 4). $\mathcal{C}^{x,u}$ outperforms $\mathcal{C}^{x}$ in all cases. Artifacts can emerge when transforming the entire model letter using a transformation estimated by considering only the boundary contour, eg. in row 6, col. 4, points inside the model letter `V' have been mapped outside the boundary after transformation to `N'.}
\label{fig:aboriginal1}
\end{figure}


\begin{figure*}[t]
\begin{center}
\begin{tabular}{ c c }

\includegraphics[width = .5\linewidth, height = .7\linewidth]{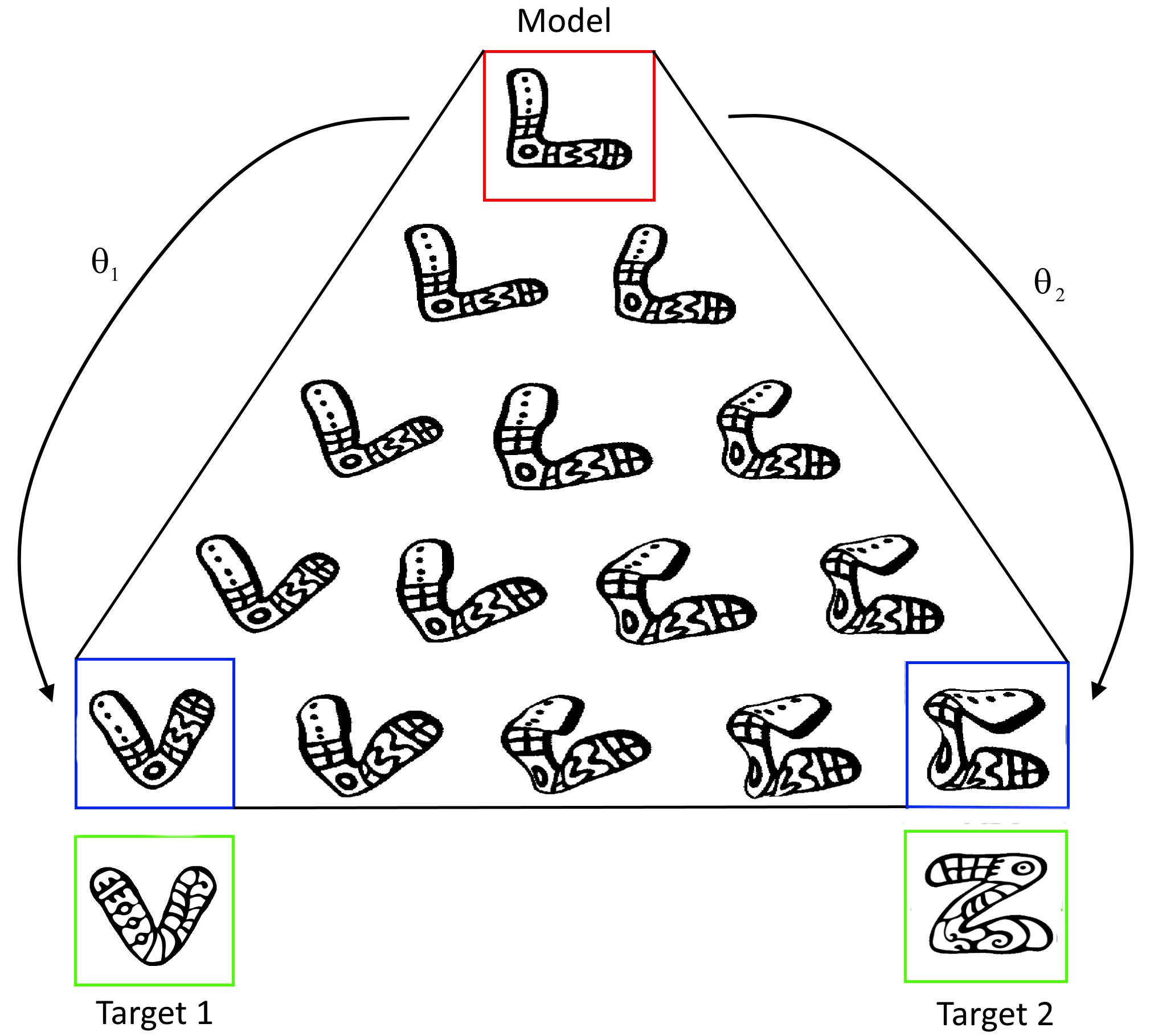}  & 

\includegraphics[width = .5\linewidth, height = .7\linewidth]{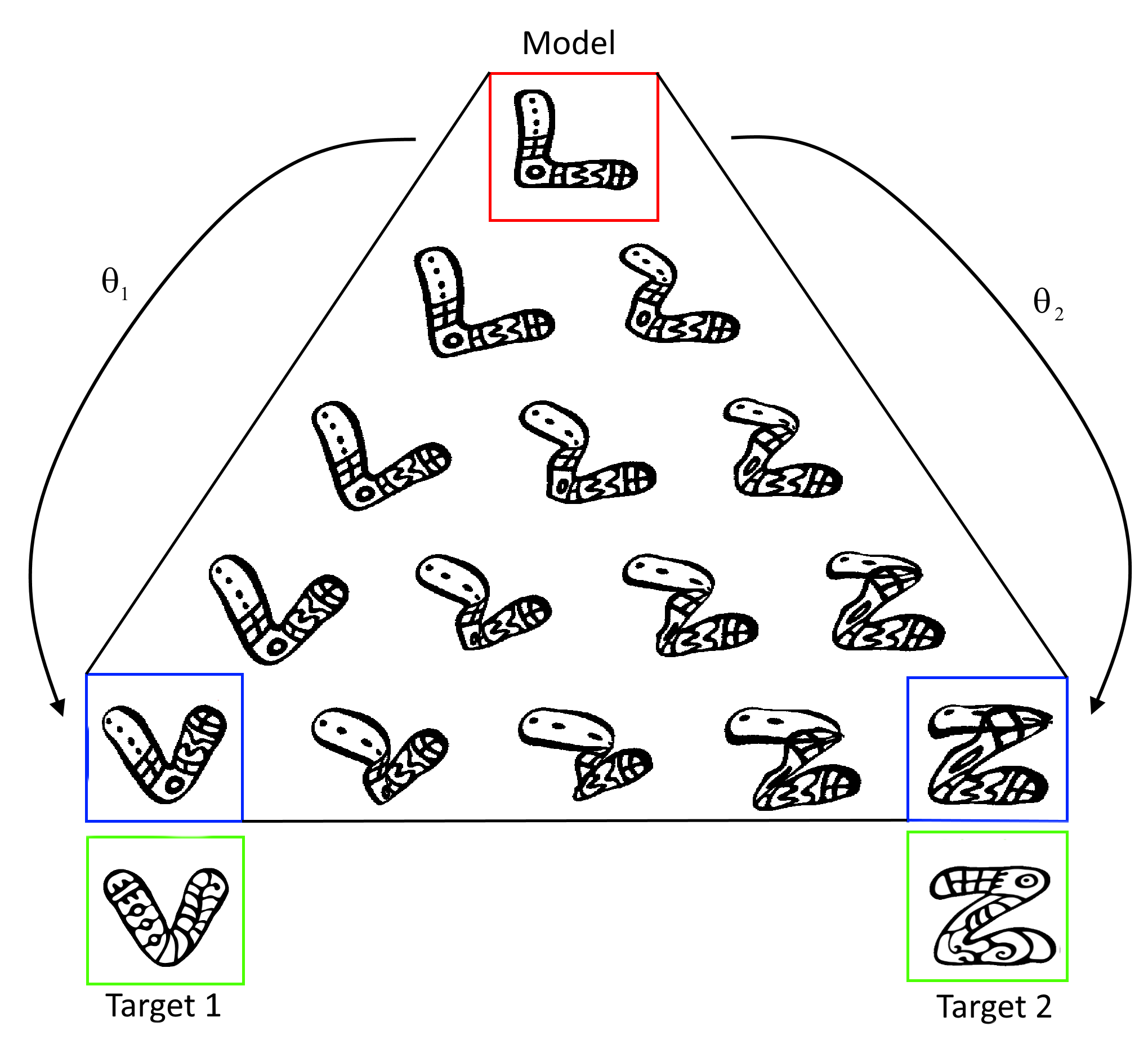} \\
(a) Using $\mathcal{C}^{x}$  & (b) Using $\mathcal{C}^{x,u}$\\

\end{tabular}
\end{center}
\caption{ Curve registration and interpolation results generated using (a) $\mathcal{C}^{x}$  and (b) $\mathcal{C}^{x,u}$. In both cases, $\mathcal{C}^{x}$  and $\mathcal{C}^{x,u}$ are used to register the model letter `L' (red) to target letters `V' and `Z' (green). The registration results after transformation using the estimated parameters $\theta_1$ and $\theta_2$ are outlined in blue, showing that $\mathcal{C}^{x,u}$ performs better than $\mathcal{C}^{x}$ when registering `L' to `Z'. In both cases, new shapes can be created by interpolating between the model shape `L' and its transformations into `V' and `Z'. These are shown in the pyramids.    }
\label{fig:pyramids}
\end{figure*}

We applied the cost functions $\mathcal{C}^{x,u}$ and $\mathcal{C}^{x}$ to the registration of curves extracted from images taken from a dataset provided by Lu et al. \cite{Lu2014}\footnote{Available at \url{http://gfx.cs.princeton.edu/pubs/Lu_2014_DDS/}}. Each image represents a patterned letter, and for a given pair of model and target letters we extracted 50 points along their external boundary contour (cf. Fig. \ref{fig:aboriginal1}, row 1,3,5,7). We then registered these point clouds, and applied the estimated transformation to all black pixels in the model letter, transforming it into the target letter (cf. Fig. \ref{fig:aboriginal1}, row 2,4,6,8). From Figure \ref{fig:aboriginal1} we can see that $\mathcal{C}^{x,u}$ outperforms $\mathcal{C}^{x}$ when registering the model and target point clouds (row 1,3,5,7). While the connectivity between the points is not taken into account when registering the point clouds using $\mathcal{C}^{x}$, with $\mathcal{C}^{x,u}$ the normal vectors are computed by fitting a spline to the ordered points, thus capturing some of the connectivity information between them, and giving a better registration result. When the estimated transformation is applied to all black pixels in the model letter, $\mathcal{C}^{x,u}$ again gives a better result. In row 6 and 8 we can see that artifacts can emerge during this step, even when the original point clouds have been registered almost exactly. This occurs when some points inside the boundary curve of the model letter, which are not taken into account during registration, get mapped outside the boundary contour by the TPS transformation, eg. in row 6, column 4, points inside the model letter `V' have been mapped outside the boundary contour when it is transformed to `N'. This could be resolved by considering points inside the boundary contour during registration. We also found that in some cases, neither cost function performed well as the model and target letters were too different, and an appropriate TPS transformation could not be estimated that would transform one shape into another.


As the transformations being estimated are parametric, we can create new transformations by interpolating between solutions. For example, given two solutions $\theta_1$ and $\theta_2$, estimated when registering a model letter to two different target letters, we can create interpolations between the three letters using the new transformation $\theta_{new}$:
\begin{equation}
\theta_{new} = \alpha_1 \theta_{Id} + \alpha_2  \theta_{1} + \alpha_3  \theta_{2} ,
\end{equation}
where $\theta_{Id}$ is the identity transformation and $\alpha_i$ are scalars. The pyramids in Figure \ref{fig:pyramids} display samples of shapes generated by interpolating between the estimated transformations, computed using different values of $\alpha_i$. Similar interpolation results have been presented when using optimal transport for colour transfer and shape registration. These methods use a discrete grid representation of shapes in 2D and 3D and do not explicitly take into account shape connectivity when estimating a registration solution  \cite{Solomon2015,Bonneel20162,Bonneel2015}.

\section{Conclusion}

We have proposed several cost functions to perform registration of shapes encoded with vertex and normal information. These were assessed experimentally for rigid (rotation) and non-rigid transformation for 2D contours and 3D  surfaces. 
We found that our new cost function  $\mathcal{C}^{x,u}$ combining normal and vertex information  overall outperform others:
\begin{itemize}
\item For  rotation estimation (2D \& 3D),  $\mathcal{C}^{x,u}$  performs best overall in terms of accuracy, outperforming Jian et al's cost function $\mathcal{C}^{x}$ \cite{Jian2011} as well as  CPD \cite{CPD2010} and Go ICP \cite{GoICP2013}. 
\item For 2D shapes differing by ONLY a non-rigid transformation we found that all techniques perform similarly.
\item For 2D shapes differing by  a non-rigid transformation AND a rotation,
 $\mathcal{C}^{x,u}$ outperforms $\mathcal{C}^{x}$ \cite{Jian2011} as well as CPD \cite{CPD2010}  and GLMD \cite{GLMD2015}. 
\item
When partial curves are registered and correspondences are used, $\mathcal{C}_{corr}^{x,u}$ also outperforms CPD \cite{CPD2010}  and $\mathcal{C}_{corr}^{x}$, giving similar results to GLMD \cite{GLMD2015}.
\end{itemize}
 However, in the case of 3D shapes differing by a non-rigid deformation we found that the high dimensional latent space and the small number of control points used reduced the accuracy of $\mathcal{C}_{corr}^{x,u}$ and $\mathcal{C}_{corr}^{x}$. As correspondences were incorporated into both cost functions to reduce computational cost, the accuracy of the results also depended on the quality of the correspondences estimated. The need to compute derivatives and normals vectors at each iteration when using $\mathcal{C}_{corr}^{x,u}$ also increased the computational cost of the algorithm.   
  Implementing a new optimisation technique which is less time consuming and could explore the latent space quickly would ensure that this type of cost functions could be used when the dimension of the space is high. Optimising a combination of these cost functions could also prove beneficial for robust registration, such as removing the rotational difference between shapes using normal information (cost function $\mathcal{C}^u$) before estimating the non-rigid transformation with $\mathcal{C}^{x,u}$.

\nocite{PhDGrogan17}

\section*{Acknowledgements}
\small 
This work has been supported by a Ussher scholarship from Trinity College
Dublin (Ireland), and partially supported by EU FP7-PEOPLE-2013-
IAPP GRAISearch grant (612334).
\normalsize

\bibliographystyle{IEEEtran}
\bibliography{AllRef}

\appendix

\section{Scalar product of vMF kernels}
\label{sec:scalar:vMF}

The product of two von Mises-Fisher distributions, $vMF_1=V_d(u;\mu_1,\kappa_1)$ and
$vMF_2=V_d(u;\mu_2,\kappa_2)$ can be written:
\begin{multline}
vMF_1 \times vMF_2=
C_d(\kappa_1)\ C_d(\kappa_2) \\
\times \exp\left( \|\kappa_1\mu_1+\kappa_2 \mu_2\| \  \frac{u^T(\kappa_1\mu_1+\kappa_2 \mu_2)}{\|\kappa_1\mu_1+\kappa_2 \mu_2\|}\right)
\end{multline}
In other words, the product $vMF_1 \times vMF_2$ is proportional to 
 $vMF=V_d(u;\mu,\kappa)$ such that:
\begin{equation}
vMF_1 \times vMF_2= \frac{C_d(\kappa_1)\ C_d(\kappa_2)}{C_d(\kappa)}\ vMF
\end{equation}
with $\kappa=\|\kappa_1\mu_1+\kappa_2 \mu_2\|$ and
$\mu=\frac{\kappa_1\mu_1+\kappa_2 \mu_2}{\|\kappa_1\mu_1+\kappa_2 \mu_2\|}$. Since $vMF$ integrates to 1,
the scalar product between $vMF_1$ and $vMF_2$ can be defined as:
\begin{multline}
\langle vMF_1 | vMF_2 \rangle= \int_{u\in\mathbb{S}^{d-1}} vMF_1 \times vMF_2 \ du\\
=  \frac{C_d(\kappa_1)\ C_d(\kappa_2)}{C_d(\kappa)}
\label{eq:vMFcrossVMF}
\end{multline}
The scalar product between two von Mises-Fisher distributions can therefore be easily
computed when an explicit expression for the function $C_d(\kappa)$ is
available (e.g. equation (\ref{eq:C3}) for $d=3$). Alternatively
numerical integration can be used as an approximation to 
equation (\ref{eq:Cd}) for any value $d>1$.

When modelling axial symmetric data the bimodal form of the von Mises-Fisher distribution could be used. The scalar product of two such distributions follows from Equation (\ref{eq:vMFcrossVMF}). 

\section{Cross products}

We propose modelling shape vertices and their normals using a combination of Dirac, von Mises-Fisher and Normal kernels. Computing the $\mathcal{L}_2$ distance between the proposed KDEs relies on the scalar products between their associated kernels. 
All of these scalar products are summarised in Table \ref{tab:scalarProducts}.   
\begin{table}[t]
\begin{center}
   \begin{tabular}{|c|c|c|}
 \hline
  & \multicolumn{2}{|c|}{$u\in \mathbb{S}^{d-1}$}\\
 \hline
 &\multicolumn{1}{|c|}{$ \delta(u-\mu_1)$} & \multicolumn{1}{|c|}{$vMF(\mu_1,\kappa_1)$}\\
  \hline 
  $ \delta(u-\mu_2)$&
    \multicolumn{1}{|c|}{\ding{55}}& \small{ $C_d(\kappa_1) \exp(\kappa_1 \  \mu_1^T \mu_2)$ }\\
 \hline
  \small{$ vMF(\mu_2,\kappa_2)$} &
\small{$C_d(\kappa_2) \exp(\kappa_2 \  \mu_2^T \mu_1)$}
&  \large{$\frac{C_d(\kappa_1)\ C_d(\kappa_2)}{C_d(\|\kappa_1\mu_1+\kappa_2 \mu_2\|)}$}\\
 \hline
   \hline
 \end{tabular}
  \begin{tabular}{|c|c|c|}
 \hline
  &\multicolumn{2}{|c|}{$x\in \mathbb{R}^d$}\\
  \hline
  &\multicolumn{1}{|c|}{$\delta(x-\mu_1)$} &  \multicolumn{1}{|c|}{\small{$\mathcal{N}(x;\mu_1,h_1^2)$ }}\\
  \hline
 \hline
  $\delta(x-\mu_2)$ &\multicolumn{1}{|c|}{\ding{55}}&
   \small{$\mathcal{N}(\mu_1; \mu_2,h_1^2)$}, \small{\cite{ScottTechnometrics2001}}\\
    \small{$ \mathcal{N}(x;\mu_2,h_2^2)$}&
  \small{$\mathcal{N}(\mu_1; \mu_2,h_2^2)$},
  \small{\cite{ScottTechnometrics2001}} & \small{$\mathcal{N}(\mu_1; \mu_2,h_1^2+h_2^2)$}
  \small{\cite{Jian2011}}\\
  \hline
 \end{tabular}
 
 \caption{Scalar products for Gaussian ($\mathcal{N}$), von Mises-Fisher ($vMF$) and Dirac ($\delta$) kernels. }
    \label{tab:scalarProducts}
\end{center}
\end{table}

\end{document}

%% file: rgb.tex
\definecolor{AliceBlue}{rgb}{0.94,0.97,1.00}
\definecolor{AntiqueWhite1}{rgb}{1.00,0.94,0.86}
\definecolor{AntiqueWhite2}{rgb}{0.93,0.87,0.80}
\definecolor{AntiqueWhite3}{rgb}{0.80,0.75,0.69}
\definecolor{AntiqueWhite4}{rgb}{0.55,0.51,0.47}
\definecolor{AntiqueWhite}{rgb}{0.98,0.92,0.84}
\definecolor{BlanchedAlmond}{rgb}{1.00,0.92,0.80}
\definecolor{BlueViolet}{rgb}{0.54,0.17,0.89}
\definecolor{CadetBlue1}{rgb}{0.60,0.96,1.00}
\definecolor{CadetBlue2}{rgb}{0.56,0.90,0.93}
\definecolor{CadetBlue3}{rgb}{0.48,0.77,0.80}
\definecolor{CadetBlue4}{rgb}{0.33,0.53,0.55}
\definecolor{CadetBlue}{rgb}{0.37,0.62,0.63}
\definecolor{CornflowerBlue}{rgb}{0.39,0.58,0.93}
\definecolor{DarkBlue}{rgb}{0.00,0.00,0.55}
\definecolor{DarkCyan}{rgb}{0.00,0.55,0.55}
\definecolor{DarkGoldenrod1}{rgb}{1.00,0.73,0.06}
\definecolor{DarkGoldenrod2}{rgb}{0.93,0.68,0.05}
\definecolor{DarkGoldenrod3}{rgb}{0.80,0.58,0.05}
\definecolor{DarkGoldenrod4}{rgb}{0.55,0.40,0.03}
\definecolor{DarkGoldenrod}{rgb}{0.72,0.53,0.04}
\definecolor{DarkGray}{rgb}{0.66,0.66,0.66}
\definecolor{DarkGreen}{rgb}{0.00,0.39,0.00}
\definecolor{DarkGrey}{rgb}{0.66,0.66,0.66}
\definecolor{DarkKhaki}{rgb}{0.74,0.72,0.42}
\definecolor{DarkMagenta}{rgb}{0.55,0.00,0.55}
\definecolor{DarkOliveGreen1}{rgb}{0.79,1.00,0.44}
\definecolor{DarkOliveGreen2}{rgb}{0.74,0.93,0.41}
\definecolor{DarkOliveGreen3}{rgb}{0.64,0.80,0.35}
\definecolor{DarkOliveGreen4}{rgb}{0.43,0.55,0.24}
\definecolor{DarkOliveGreen}{rgb}{0.33,0.42,0.18}
\definecolor{OliveGreen}{rgb}{0.33,0.42,0.18}
\definecolor{DarkOrange1}{rgb}{1.00,0.50,0.00}
\definecolor{DarkOrange2}{rgb}{0.93,0.46,0.00}
\definecolor{DarkOrange3}{rgb}{0.80,0.40,0.00}
\definecolor{DarkOrange4}{rgb}{0.55,0.27,0.00}
\definecolor{DarkOrange}{rgb}{1.00,0.55,0.00}
\definecolor{DarkOrchid1}{rgb}{0.75,0.24,1.00}
\definecolor{DarkOrchid2}{rgb}{0.70,0.23,0.93}
\definecolor{DarkOrchid3}{rgb}{0.60,0.20,0.80}
\definecolor{DarkOrchid4}{rgb}{0.41,0.13,0.55}
\definecolor{DarkOrchid}{rgb}{0.60,0.20,0.80}
\definecolor{DarkRed}{rgb}{0.55,0.00,0.00}
\definecolor{DarkSalmon}{rgb}{0.91,0.59,0.48}
\definecolor{DarkSeaGreen1}{rgb}{0.76,1.00,0.76}
\definecolor{DarkSeaGreen2}{rgb}{0.71,0.93,0.71}
\definecolor{DarkSeaGreen3}{rgb}{0.61,0.80,0.61}
\definecolor{DarkSeaGreen4}{rgb}{0.41,0.55,0.41}
\definecolor{DarkSeaGreen}{rgb}{0.56,0.74,0.56}
\definecolor{DarkSlateBlue}{rgb}{0.28,0.24,0.55}
\definecolor{DarkSlateGray1}{rgb}{0.59,1.00,1.00}
\definecolor{DarkSlateGray2}{rgb}{0.55,0.93,0.93}
\definecolor{DarkSlateGray3}{rgb}{0.47,0.80,0.80}
\definecolor{DarkSlateGray4}{rgb}{0.32,0.55,0.55}
\definecolor{DarkSlateGray}{rgb}{0.18,0.31,0.31}
\definecolor{DarkSlateGrey}{rgb}{0.18,0.31,0.31}
\definecolor{DarkTurquoise}{rgb}{0.00,0.81,0.82}
\definecolor{DarkViolet}{rgb}{0.58,0.00,0.83}
\definecolor{DeepPink1}{rgb}{1.00,0.08,0.58}
\definecolor{DeepPink2}{rgb}{0.93,0.07,0.54}
\definecolor{DeepPink3}{rgb}{0.80,0.06,0.46}
\definecolor{DeepPink4}{rgb}{0.55,0.04,0.31}
\definecolor{DeepPink}{rgb}{1.00,0.08,0.58}
\definecolor{DeepSkyBlue1}{rgb}{0.00,0.75,1.00}
\definecolor{DeepSkyBlue2}{rgb}{0.00,0.70,0.93}
\definecolor{DeepSkyBlue3}{rgb}{0.00,0.60,0.80}
\definecolor{DeepSkyBlue4}{rgb}{0.00,0.41,0.55}
\definecolor{DeepSkyBlue}{rgb}{0.00,0.75,1.00}
\definecolor{DimGray}{rgb}{0.41,0.41,0.41}
\definecolor{DimGrey}{rgb}{0.41,0.41,0.41}
\definecolor{DodgerBlue1}{rgb}{0.12,0.56,1.00}
\definecolor{DodgerBlue2}{rgb}{0.11,0.53,0.93}
\definecolor{DodgerBlue3}{rgb}{0.09,0.45,0.80}
\definecolor{DodgerBlue4}{rgb}{0.06,0.31,0.55}
\definecolor{DodgerBlue}{rgb}{0.12,0.56,1.00}
\definecolor{FloralWhite}{rgb}{1.00,0.98,0.94}
\definecolor{ForestGreen}{rgb}{0.13,0.55,0.13}
\definecolor{GhostWhite}{rgb}{0.97,0.97,1.00}
\definecolor{GreenYellow}{rgb}{0.68,1.00,0.18}
\definecolor{HotPink1}{rgb}{1.00,0.43,0.71}
\definecolor{HotPink2}{rgb}{0.93,0.42,0.65}
\definecolor{HotPink3}{rgb}{0.80,0.38,0.56}
\definecolor{HotPink4}{rgb}{0.55,0.23,0.38}
\definecolor{HotPink}{rgb}{1.00,0.41,0.71}
\definecolor{IndianRed1}{rgb}{1.00,0.42,0.42}
\definecolor{IndianRed2}{rgb}{0.93,0.39,0.39}
\definecolor{IndianRed3}{rgb}{0.80,0.33,0.33}
\definecolor{IndianRed4}{rgb}{0.55,0.23,0.23}
\definecolor{IndianRed}{rgb}{0.80,0.36,0.36}
\definecolor{LavenderBlush1}{rgb}{1.00,0.94,0.96}
\definecolor{LavenderBlush2}{rgb}{0.93,0.88,0.90}
\definecolor{LavenderBlush3}{rgb}{0.80,0.76,0.77}
\definecolor{LavenderBlush4}{rgb}{0.55,0.51,0.53}
\definecolor{LavenderBlush}{rgb}{1.00,0.94,0.96}
\definecolor{LawnGreen}{rgb}{0.49,0.99,0.00}
\definecolor{LemonChiffon1}{rgb}{1.00,0.98,0.80}
\definecolor{LemonChiffon2}{rgb}{0.93,0.91,0.75}
\definecolor{LemonChiffon3}{rgb}{0.80,0.79,0.65}
\definecolor{LemonChiffon4}{rgb}{0.55,0.54,0.44}
\definecolor{LemonChiffon}{rgb}{1.00,0.98,0.80}
\definecolor{LightBlue1}{rgb}{0.75,0.94,1.00}
\definecolor{LightBlue2}{rgb}{0.70,0.87,0.93}
\definecolor{LightBlue3}{rgb}{0.60,0.75,0.80}
\definecolor{LightBlue4}{rgb}{0.41,0.51,0.55}
\definecolor{LightBlue}{rgb}{0.68,0.85,0.90}
\definecolor{LightCoral}{rgb}{0.94,0.50,0.50}
\definecolor{LightCyan1}{rgb}{0.88,1.00,1.00}
\definecolor{LightCyan2}{rgb}{0.82,0.93,0.93}
\definecolor{LightCyan3}{rgb}{0.71,0.80,0.80}
\definecolor{LightCyan4}{rgb}{0.48,0.55,0.55}
\definecolor{LightCyan}{rgb}{0.88,1.00,1.00}
\definecolor{LightGoldenrod1}{rgb}{1.00,0.93,0.55}
\definecolor{LightGoldenrod2}{rgb}{0.93,0.86,0.51}
\definecolor{LightGoldenrod3}{rgb}{0.80,0.75,0.44}
\definecolor{LightGoldenrod4}{rgb}{0.55,0.51,0.30}
\definecolor{LightGoldenrodYellow}{rgb}{0.98,0.98,0.82}
\definecolor{LightGoldenrod}{rgb}{0.93,0.87,0.51}
\definecolor{LightGray}{rgb}{0.83,0.83,0.83}
\definecolor{LightGreen}{rgb}{0.56,0.93,0.56}
\definecolor{LightGrey}{rgb}{0.83,0.83,0.83}
\definecolor{LightPink1}{rgb}{1.00,0.68,0.73}
\definecolor{LightPink2}{rgb}{0.93,0.64,0.68}
\definecolor{LightPink3}{rgb}{0.80,0.55,0.58}
\definecolor{LightPink4}{rgb}{0.55,0.37,0.40}
\definecolor{LightPink}{rgb}{1.00,0.71,0.76}
\definecolor{LightSalmon1}{rgb}{1.00,0.63,0.48}
\definecolor{LightSalmon2}{rgb}{0.93,0.58,0.45}
\definecolor{LightSalmon3}{rgb}{0.80,0.51,0.38}
\definecolor{LightSalmon4}{rgb}{0.55,0.34,0.26}
\definecolor{LightSalmon}{rgb}{1.00,0.63,0.48}
\definecolor{LightSeaGreen}{rgb}{0.13,0.70,0.67}
\definecolor{LightSkyBlue1}{rgb}{0.69,0.89,1.00}
\definecolor{LightSkyBlue2}{rgb}{0.64,0.83,0.93}
\definecolor{LightSkyBlue3}{rgb}{0.55,0.71,0.80}
\definecolor{LightSkyBlue4}{rgb}{0.38,0.48,0.55}
\definecolor{LightSkyBlue}{rgb}{0.53,0.81,0.98}
\definecolor{LightSlateBlue}{rgb}{0.52,0.44,1.00}
\definecolor{LightSlateGray}{rgb}{0.47,0.53,0.60}
\definecolor{LightSlateGrey}{rgb}{0.47,0.53,0.60}
\definecolor{LightSteelBlue1}{rgb}{0.79,0.88,1.00}
\definecolor{LightSteelBlue2}{rgb}{0.74,0.82,0.93}
\definecolor{LightSteelBlue3}{rgb}{0.64,0.71,0.80}
\definecolor{LightSteelBlue4}{rgb}{0.43,0.48,0.55}
\definecolor{LightSteelBlue}{rgb}{0.69,0.77,0.87}
\definecolor{LightYellow1}{rgb}{1.00,1.00,0.88}
\definecolor{LightYellow2}{rgb}{0.93,0.93,0.82}
\definecolor{LightYellow3}{rgb}{0.80,0.80,0.71}
\definecolor{LightYellow4}{rgb}{0.55,0.55,0.48}
\definecolor{LightYellow}{rgb}{1.00,1.00,0.88}
\definecolor{LimeGreen}{rgb}{0.20,0.80,0.20}
\definecolor{MediumAquamarine}{rgb}{0.40,0.80,0.67}
\definecolor{MediumBlue}{rgb}{0.00,0.00,0.80}
\definecolor{MediumOrchid1}{rgb}{0.88,0.40,1.00}
\definecolor{MediumOrchid2}{rgb}{0.82,0.37,0.93}
\definecolor{MediumOrchid3}{rgb}{0.71,0.32,0.80}
\definecolor{MediumOrchid4}{rgb}{0.48,0.22,0.55}
\definecolor{MediumOrchid}{rgb}{0.73,0.33,0.83}
\definecolor{MediumPurple1}{rgb}{0.67,0.51,1.00}
\definecolor{MediumPurple2}{rgb}{0.62,0.47,0.93}
\definecolor{MediumPurple3}{rgb}{0.54,0.41,0.80}
\definecolor{MediumPurple4}{rgb}{0.36,0.28,0.55}
\definecolor{MediumPurple}{rgb}{0.58,0.44,0.86}
\definecolor{MediumSeaGreen}{rgb}{0.24,0.70,0.44}
\definecolor{MediumSlateBlue}{rgb}{0.48,0.41,0.93}
\definecolor{MediumSpringGreen}{rgb}{0.00,0.98,0.60}
\definecolor{MediumTurquoise}{rgb}{0.28,0.82,0.80}
\definecolor{MediumVioletRed}{rgb}{0.78,0.08,0.52}
\definecolor{MidnightBlue}{rgb}{0.10,0.10,0.44}
\definecolor{MintCream}{rgb}{0.96,1.00,0.98}
\definecolor{MistyRose1}{rgb}{1.00,0.89,0.88}
\definecolor{MistyRose2}{rgb}{0.93,0.84,0.82}
\definecolor{MistyRose3}{rgb}{0.80,0.72,0.71}
\definecolor{MistyRose4}{rgb}{0.55,0.49,0.48}
\definecolor{MistyRose}{rgb}{1.00,0.89,0.88}
\definecolor{NavajoWhite1}{rgb}{1.00,0.87,0.68}
\definecolor{NavajoWhite2}{rgb}{0.93,0.81,0.63}
\definecolor{NavajoWhite3}{rgb}{0.80,0.70,0.55}
\definecolor{NavajoWhite4}{rgb}{0.55,0.47,0.37}
\definecolor{NavajoWhite}{rgb}{1.00,0.87,0.68}
\definecolor{NavyBlue}{rgb}{0.00,0.00,0.50}
\definecolor{OldLace}{rgb}{0.99,0.96,0.90}
\definecolor{OliveDrab1}{rgb}{0.75,1.00,0.24}
\definecolor{OliveDrab2}{rgb}{0.70,0.93,0.23}
\definecolor{OliveDrab3}{rgb}{0.60,0.80,0.20}
\definecolor{OliveDrab4}{rgb}{0.41,0.55,0.13}
\definecolor{OliveDrab}{rgb}{0.42,0.56,0.14}
\definecolor{OrangeRed1}{rgb}{1.00,0.27,0.00}
\definecolor{OrangeRed2}{rgb}{0.93,0.25,0.00}
\definecolor{OrangeRed3}{rgb}{0.80,0.22,0.00}
\definecolor{OrangeRed4}{rgb}{0.55,0.15,0.00}
\definecolor{OrangeRed}{rgb}{1.00,0.27,0.00}
\definecolor{PaleGoldenrod}{rgb}{0.93,0.91,0.67}
\definecolor{PaleGreen1}{rgb}{0.60,1.00,0.60}
\definecolor{PaleGreen2}{rgb}{0.56,0.93,0.56}
\definecolor{PaleGreen3}{rgb}{0.49,0.80,0.49}
\definecolor{PaleGreen4}{rgb}{0.33,0.55,0.33}
\definecolor{PaleGreen}{rgb}{0.60,0.98,0.60}
\definecolor{PaleTurquoise1}{rgb}{0.73,1.00,1.00}
\definecolor{PaleTurquoise2}{rgb}{0.68,0.93,0.93}
\definecolor{PaleTurquoise3}{rgb}{0.59,0.80,0.80}
\definecolor{PaleTurquoise4}{rgb}{0.40,0.55,0.55}
\definecolor{PaleTurquoise}{rgb}{0.69,0.93,0.93}
\definecolor{PaleVioletRed1}{rgb}{1.00,0.51,0.67}
\definecolor{PaleVioletRed2}{rgb}{0.93,0.47,0.62}
\definecolor{PaleVioletRed3}{rgb}{0.80,0.41,0.54}
\definecolor{PaleVioletRed4}{rgb}{0.55,0.28,0.36}
\definecolor{PaleVioletRed}{rgb}{0.86,0.44,0.58}
\definecolor{PapayaWhip}{rgb}{1.00,0.94,0.84}
\definecolor{PeachPuff1}{rgb}{1.00,0.85,0.73}
\definecolor{PeachPuff2}{rgb}{0.93,0.80,0.68}
\definecolor{PeachPuff3}{rgb}{0.80,0.69,0.58}
\definecolor{PeachPuff4}{rgb}{0.55,0.47,0.40}
\definecolor{PeachPuff}{rgb}{1.00,0.85,0.73}
\definecolor{PowderBlue}{rgb}{0.69,0.88,0.90}
\definecolor{RosyBrown1}{rgb}{1.00,0.76,0.76}
\definecolor{RosyBrown2}{rgb}{0.93,0.71,0.71}
\definecolor{RosyBrown3}{rgb}{0.80,0.61,0.61}
\definecolor{RosyBrown4}{rgb}{0.55,0.41,0.41}
\definecolor{RosyBrown}{rgb}{0.74,0.56,0.56}
\definecolor{RoyalBlue1}{rgb}{0.28,0.46,1.00}
\definecolor{RoyalBlue2}{rgb}{0.26,0.43,0.93}
\definecolor{RoyalBlue3}{rgb}{0.23,0.37,0.80}
\definecolor{RoyalBlue4}{rgb}{0.15,0.25,0.55}
\definecolor{RoyalBlue}{rgb}{0.25,0.41,0.88}
\definecolor{SaddleBrown}{rgb}{0.55,0.27,0.07}
\definecolor{SandyBrown}{rgb}{0.96,0.64,0.38}
\definecolor{SeaGreen1}{rgb}{0.33,1.00,0.62}
\definecolor{SeaGreen2}{rgb}{0.31,0.93,0.58}
\definecolor{SeaGreen3}{rgb}{0.26,0.80,0.50}
\definecolor{SeaGreen4}{rgb}{0.18,0.55,0.34}
\definecolor{SeaGreen}{rgb}{0.18,0.55,0.34}
\definecolor{SkyBlue1}{rgb}{0.53,0.81,1.00}
\definecolor{SkyBlue2}{rgb}{0.49,0.75,0.93}
\definecolor{SkyBlue3}{rgb}{0.42,0.65,0.80}
\definecolor{SkyBlue4}{rgb}{0.29,0.44,0.55}
\definecolor{SkyBlue}{rgb}{0.53,0.81,0.92}
\definecolor{SlateBlue1}{rgb}{0.51,0.44,1.00}
\definecolor{SlateBlue2}{rgb}{0.48,0.40,0.93}
\definecolor{SlateBlue3}{rgb}{0.41,0.35,0.80}
\definecolor{SlateBlue4}{rgb}{0.28,0.24,0.55}
\definecolor{SlateBlue}{rgb}{0.42,0.35,0.80}
\definecolor{SlateGray1}{rgb}{0.78,0.89,1.00}
\definecolor{SlateGray2}{rgb}{0.73,0.83,0.93}
\definecolor{SlateGray3}{rgb}{0.62,0.71,0.80}
\definecolor{SlateGray4}{rgb}{0.42,0.48,0.55}
\definecolor{SlateGray}{rgb}{0.44,0.50,0.56}
\definecolor{SlateGrey}{rgb}{0.44,0.50,0.56}
\definecolor{SpringGreen1}{rgb}{0.00,1.00,0.50}
\definecolor{SpringGreen2}{rgb}{0.00,0.93,0.46}
\definecolor{SpringGreen3}{rgb}{0.00,0.80,0.40}
\definecolor{SpringGreen4}{rgb}{0.00,0.55,0.27}
\definecolor{SpringGreen}{rgb}{0.00,1.00,0.50}
\definecolor{SteelBlue1}{rgb}{0.39,0.72,1.00}
\definecolor{SteelBlue2}{rgb}{0.36,0.67,0.93}
\definecolor{SteelBlue3}{rgb}{0.31,0.58,0.80}
\definecolor{SteelBlue4}{rgb}{0.21,0.39,0.55}
\definecolor{SteelBlue}{rgb}{0.27,0.51,0.71}
\definecolor{VioletRed1}{rgb}{1.00,0.24,0.59}
\definecolor{VioletRed2}{rgb}{0.93,0.23,0.55}
\definecolor{VioletRed3}{rgb}{0.80,0.20,0.47}
\definecolor{VioletRed4}{rgb}{0.55,0.13,0.32}
\definecolor{VioletRed}{rgb}{0.82,0.13,0.56}
\definecolor{WhiteSmoke}{rgb}{0.96,0.96,0.96}
\definecolor{YellowGreen}{rgb}{0.60,0.80,0.20}
\definecolor{aliceblue}{rgb}{0.94,0.97,1.00}
\definecolor{antiquewhite}{rgb}{0.98,0.92,0.84}
\definecolor{aquamarine1}{rgb}{0.50,1.00,0.83}
\definecolor{aquamarine2}{rgb}{0.46,0.93,0.78}
\definecolor{aquamarine3}{rgb}{0.40,0.80,0.67}
\definecolor{aquamarine4}{rgb}{0.27,0.55,0.45}
\definecolor{aquamarine}{rgb}{0.50,1.00,0.83}
\definecolor{azure1}{rgb}{0.94,1.00,1.00}
\definecolor{azure2}{rgb}{0.88,0.93,0.93}
\definecolor{azure3}{rgb}{0.76,0.80,0.80}
\definecolor{azure4}{rgb}{0.51,0.55,0.55}
\definecolor{azure}{rgb}{0.94,1.00,1.00}
\definecolor{beige}{rgb}{0.96,0.96,0.86}
\definecolor{bisque1}{rgb}{1.00,0.89,0.77}
\definecolor{bisque2}{rgb}{0.93,0.84,0.72}
\definecolor{bisque3}{rgb}{0.80,0.72,0.62}
\definecolor{bisque4}{rgb}{0.55,0.49,0.42}
\definecolor{bisque}{rgb}{1.00,0.89,0.77}
\definecolor{black}{rgb}{0.00,0.00,0.00}
\definecolor{blanchedalmond}{rgb}{1.00,0.92,0.80}
\definecolor{blue1}{rgb}{0.00,0.00,1.00}
\definecolor{blue2}{rgb}{0.00,0.00,0.93}
\definecolor{blue3}{rgb}{0.00,0.00,0.80}
\definecolor{blue4}{rgb}{0.00,0.00,0.55}
\definecolor{blueviolet}{rgb}{0.54,0.17,0.89}
\definecolor{blue}{rgb}{0.00,0.00,1.00}
\definecolor{brown1}{rgb}{1.00,0.25,0.25}
\definecolor{brown2}{rgb}{0.93,0.23,0.23}
\definecolor{brown3}{rgb}{0.80,0.20,0.20}
\definecolor{brown4}{rgb}{0.55,0.14,0.14}
\definecolor{brown}{rgb}{0.65,0.16,0.16}
\definecolor{burlywood1}{rgb}{1.00,0.83,0.61}
\definecolor{burlywood2}{rgb}{0.93,0.77,0.57}
\definecolor{burlywood3}{rgb}{0.80,0.67,0.49}
\definecolor{burlywood4}{rgb}{0.55,0.45,0.33}
\definecolor{burlywood}{rgb}{0.87,0.72,0.53}
\definecolor{cadetblue}{rgb}{0.37,0.62,0.63}
\definecolor{chartreuse1}{rgb}{0.50,1.00,0.00}
\definecolor{chartreuse2}{rgb}{0.46,0.93,0.00}
\definecolor{chartreuse3}{rgb}{0.40,0.80,0.00}
\definecolor{chartreuse4}{rgb}{0.27,0.55,0.00}
\definecolor{chartreuse}{rgb}{0.50,1.00,0.00}
\definecolor{chocolate1}{rgb}{1.00,0.50,0.14}
\definecolor{chocolate2}{rgb}{0.93,0.46,0.13}
\definecolor{chocolate3}{rgb}{0.80,0.40,0.11}
\definecolor{chocolate4}{rgb}{0.55,0.27,0.07}
\definecolor{chocolate}{rgb}{0.82,0.41,0.12}
\definecolor{coral1}{rgb}{1.00,0.45,0.34}
\definecolor{coral2}{rgb}{0.93,0.42,0.31}
\definecolor{coral3}{rgb}{0.80,0.36,0.27}
\definecolor{coral4}{rgb}{0.55,0.24,0.18}
\definecolor{coral}{rgb}{1.00,0.50,0.31}
\definecolor{cornflowerblue}{rgb}{0.39,0.58,0.93}
\definecolor{cornsilk1}{rgb}{1.00,0.97,0.86}
\definecolor{cornsilk2}{rgb}{0.93,0.91,0.80}
\definecolor{cornsilk3}{rgb}{0.80,0.78,0.69}
\definecolor{cornsilk4}{rgb}{0.55,0.53,0.47}
\definecolor{cornsilk}{rgb}{1.00,0.97,0.86}
\definecolor{cyan1}{rgb}{0.00,1.00,1.00}
\definecolor{cyan2}{rgb}{0.00,0.93,0.93}
\definecolor{cyan3}{rgb}{0.00,0.80,0.80}
\definecolor{cyan4}{rgb}{0.00,0.55,0.55}
\definecolor{cyan}{rgb}{0.00,1.00,1.00}
\definecolor{darkblue}{rgb}{0.00,0.00,0.55}
\definecolor{darkcyan}{rgb}{0.00,0.55,0.55}
\definecolor{darkgoldenrod}{rgb}{0.72,0.53,0.04}
\definecolor{darkgray}{rgb}{0.66,0.66,0.66}
\definecolor{darkgreen}{rgb}{0.00,0.39,0.00}
\definecolor{darkgrey}{rgb}{0.66,0.66,0.66}
\definecolor{darkkhaki}{rgb}{0.74,0.72,0.42}
\definecolor{darkmagenta}{rgb}{0.55,0.00,0.55}
\definecolor{darkolive}{rgb}{0.33,0.42,0.18}
\definecolor{darkorange}{rgb}{1.00,0.55,0.00}
\definecolor{darkorchid}{rgb}{0.60,0.20,0.80}
\definecolor{darkred}{rgb}{0.55,0.00,0.00}
\definecolor{darksalmon}{rgb}{0.91,0.59,0.48}
\definecolor{darksea}{rgb}{0.56,0.74,0.56}
\definecolor{darkslate}{rgb}{0.18,0.31,0.31}
\definecolor{darkslate}{rgb}{0.18,0.31,0.31}
\definecolor{darkslate}{rgb}{0.28,0.24,0.55}
\definecolor{darkturquoise}{rgb}{0.00,0.81,0.82}
\definecolor{darkviolet}{rgb}{0.58,0.00,0.83}
\definecolor{deeppink}{rgb}{1.00,0.08,0.58}
\definecolor{deepsky}{rgb}{0.00,0.75,1.00}
\definecolor{dimgray}{rgb}{0.41,0.41,0.41}
\definecolor{dimgrey}{rgb}{0.41,0.41,0.41}
\definecolor{dodgerblue}{rgb}{0.12,0.56,1.00}
\definecolor{firebrick1}{rgb}{1.00,0.19,0.19}
\definecolor{firebrick2}{rgb}{0.93,0.17,0.17}
\definecolor{firebrick3}{rgb}{0.80,0.15,0.15}
\definecolor{firebrick4}{rgb}{0.55,0.10,0.10}
\definecolor{firebrick}{rgb}{0.70,0.13,0.13}
\definecolor{floralwhite}{rgb}{1.00,0.98,0.94}
\definecolor{forestgreen}{rgb}{0.13,0.55,0.13}
\definecolor{gainsboro}{rgb}{0.86,0.86,0.86}
\definecolor{ghostwhite}{rgb}{0.97,0.97,1.00}
\definecolor{gold1}{rgb}{1.00,0.84,0.00}
\definecolor{gold2}{rgb}{0.93,0.79,0.00}
\definecolor{gold3}{rgb}{0.80,0.68,0.00}
\definecolor{gold4}{rgb}{0.55,0.46,0.00}
\definecolor{goldenrod1}{rgb}{1.00,0.76,0.15}
\definecolor{goldenrod2}{rgb}{0.93,0.71,0.13}
\definecolor{goldenrod3}{rgb}{0.80,0.61,0.11}
\definecolor{goldenrod4}{rgb}{0.55,0.41,0.08}
\definecolor{goldenrod}{rgb}{0.85,0.65,0.13}
\definecolor{gold}{rgb}{1.00,0.84,0.00}
\definecolor{gray0}{rgb}{0.00,0.00,0.00}
\definecolor{gray100}{rgb}{1.00,1.00,1.00}
\definecolor{gray10}{rgb}{0.10,0.10,0.10}
\definecolor{gray11}{rgb}{0.11,0.11,0.11}
\definecolor{gray12}{rgb}{0.12,0.12,0.12}
\definecolor{gray13}{rgb}{0.13,0.13,0.13}
\definecolor{gray14}{rgb}{0.14,0.14,0.14}
\definecolor{gray15}{rgb}{0.15,0.15,0.15}
\definecolor{gray16}{rgb}{0.16,0.16,0.16}
\definecolor{gray17}{rgb}{0.17,0.17,0.17}
\definecolor{gray18}{rgb}{0.18,0.18,0.18}
\definecolor{gray19}{rgb}{0.19,0.19,0.19}
\definecolor{gray1}{rgb}{0.01,0.01,0.01}
\definecolor{gray20}{rgb}{0.20,0.20,0.20}
\definecolor{gray21}{rgb}{0.21,0.21,0.21}
\definecolor{gray22}{rgb}{0.22,0.22,0.22}
\definecolor{gray23}{rgb}{0.23,0.23,0.23}
\definecolor{gray24}{rgb}{0.24,0.24,0.24}
\definecolor{gray25}{rgb}{0.25,0.25,0.25}
\definecolor{gray26}{rgb}{0.26,0.26,0.26}
\definecolor{gray27}{rgb}{0.27,0.27,0.27}
\definecolor{gray28}{rgb}{0.28,0.28,0.28}
\definecolor{gray29}{rgb}{0.29,0.29,0.29}
\definecolor{gray2}{rgb}{0.02,0.02,0.02}
\definecolor{gray30}{rgb}{0.30,0.30,0.30}
\definecolor{gray31}{rgb}{0.31,0.31,0.31}
\definecolor{gray32}{rgb}{0.32,0.32,0.32}
\definecolor{gray33}{rgb}{0.33,0.33,0.33}
\definecolor{gray34}{rgb}{0.34,0.34,0.34}
\definecolor{gray35}{rgb}{0.35,0.35,0.35}
\definecolor{gray36}{rgb}{0.36,0.36,0.36}
\definecolor{gray37}{rgb}{0.37,0.37,0.37}
\definecolor{gray38}{rgb}{0.38,0.38,0.38}
\definecolor{gray39}{rgb}{0.39,0.39,0.39}
\definecolor{gray3}{rgb}{0.03,0.03,0.03}
\definecolor{gray40}{rgb}{0.40,0.40,0.40}
\definecolor{gray41}{rgb}{0.41,0.41,0.41}
\definecolor{gray42}{rgb}{0.42,0.42,0.42}
\definecolor{gray43}{rgb}{0.43,0.43,0.43}
\definecolor{gray44}{rgb}{0.44,0.44,0.44}
\definecolor{gray45}{rgb}{0.45,0.45,0.45}
\definecolor{gray46}{rgb}{0.46,0.46,0.46}
\definecolor{gray47}{rgb}{0.47,0.47,0.47}
\definecolor{gray48}{rgb}{0.48,0.48,0.48}
\definecolor{gray49}{rgb}{0.49,0.49,0.49}
\definecolor{gray4}{rgb}{0.04,0.04,0.04}
\definecolor{gray50}{rgb}{0.50,0.50,0.50}
\definecolor{gray51}{rgb}{0.51,0.51,0.51}
\definecolor{gray52}{rgb}{0.52,0.52,0.52}
\definecolor{gray53}{rgb}{0.53,0.53,0.53}
\definecolor{gray54}{rgb}{0.54,0.54,0.54}
\definecolor{gray55}{rgb}{0.55,0.55,0.55}
\definecolor{gray56}{rgb}{0.56,0.56,0.56}
\definecolor{gray57}{rgb}{0.57,0.57,0.57}
\definecolor{gray58}{rgb}{0.58,0.58,0.58}
\definecolor{gray59}{rgb}{0.59,0.59,0.59}
\definecolor{gray5}{rgb}{0.05,0.05,0.05}
\definecolor{gray60}{rgb}{0.60,0.60,0.60}
\definecolor{gray61}{rgb}{0.61,0.61,0.61}
\definecolor{gray62}{rgb}{0.62,0.62,0.62}
\definecolor{gray63}{rgb}{0.63,0.63,0.63}
\definecolor{gray64}{rgb}{0.64,0.64,0.64}
\definecolor{gray65}{rgb}{0.65,0.65,0.65}
\definecolor{gray66}{rgb}{0.66,0.66,0.66}
\definecolor{gray67}{rgb}{0.67,0.67,0.67}
\definecolor{gray68}{rgb}{0.68,0.68,0.68}
\definecolor{gray69}{rgb}{0.69,0.69,0.69}
\definecolor{gray6}{rgb}{0.06,0.06,0.06}
\definecolor{gray70}{rgb}{0.70,0.70,0.70}
\definecolor{gray71}{rgb}{0.71,0.71,0.71}
\definecolor{gray72}{rgb}{0.72,0.72,0.72}
\definecolor{gray73}{rgb}{0.73,0.73,0.73}
\definecolor{gray74}{rgb}{0.74,0.74,0.74}
\definecolor{gray75}{rgb}{0.75,0.75,0.75}
\definecolor{gray76}{rgb}{0.76,0.76,0.76}
\definecolor{gray77}{rgb}{0.77,0.77,0.77}
\definecolor{gray78}{rgb}{0.78,0.78,0.78}
\definecolor{gray79}{rgb}{0.79,0.79,0.79}
\definecolor{gray7}{rgb}{0.07,0.07,0.07}
\definecolor{gray80}{rgb}{0.80,0.80,0.80}
\definecolor{gray81}{rgb}{0.81,0.81,0.81}
\definecolor{gray82}{rgb}{0.82,0.82,0.82}
\definecolor{gray83}{rgb}{0.83,0.83,0.83}
\definecolor{gray84}{rgb}{0.84,0.84,0.84}
\definecolor{gray85}{rgb}{0.85,0.85,0.85}
\definecolor{gray86}{rgb}{0.86,0.86,0.86}
\definecolor{gray87}{rgb}{0.87,0.87,0.87}
\definecolor{gray88}{rgb}{0.88,0.88,0.88}
\definecolor{gray89}{rgb}{0.89,0.89,0.89}
\definecolor{gray8}{rgb}{0.08,0.08,0.08}
\definecolor{gray90}{rgb}{0.90,0.90,0.90}
\definecolor{gray91}{rgb}{0.91,0.91,0.91}
\definecolor{gray92}{rgb}{0.92,0.92,0.92}
\definecolor{gray93}{rgb}{0.93,0.93,0.93}
\definecolor{gray94}{rgb}{0.94,0.94,0.94}
\definecolor{gray95}{rgb}{0.95,0.95,0.95}
\definecolor{gray96}{rgb}{0.96,0.96,0.96}
\definecolor{gray97}{rgb}{0.97,0.97,0.97}
\definecolor{gray98}{rgb}{0.98,0.98,0.98}
\definecolor{gray99}{rgb}{0.99,0.99,0.99}
\definecolor{gray9}{rgb}{0.09,0.09,0.09}
\definecolor{gray}{rgb}{0.75,0.75,0.75}
\definecolor{green1}{rgb}{0.00,1.00,0.00}
\definecolor{green2}{rgb}{0.00,0.93,0.00}
\definecolor{green3}{rgb}{0.00,0.80,0.00}
\definecolor{green4}{rgb}{0.00,0.55,0.00}
\definecolor{greenyellow}{rgb}{0.68,1.00,0.18}
\definecolor{green}{rgb}{0.00,1.00,0.00}
\definecolor{grey0}{rgb}{0.00,0.00,0.00}
\definecolor{grey100}{rgb}{1.00,1.00,1.00}
\definecolor{grey10}{rgb}{0.10,0.10,0.10}
\definecolor{grey11}{rgb}{0.11,0.11,0.11}
\definecolor{grey12}{rgb}{0.12,0.12,0.12}
\definecolor{grey13}{rgb}{0.13,0.13,0.13}
\definecolor{grey14}{rgb}{0.14,0.14,0.14}
\definecolor{grey15}{rgb}{0.15,0.15,0.15}
\definecolor{grey16}{rgb}{0.16,0.16,0.16}
\definecolor{grey17}{rgb}{0.17,0.17,0.17}
\definecolor{grey18}{rgb}{0.18,0.18,0.18}
\definecolor{grey19}{rgb}{0.19,0.19,0.19}
\definecolor{grey1}{rgb}{0.01,0.01,0.01}
\definecolor{grey20}{rgb}{0.20,0.20,0.20}
\definecolor{grey21}{rgb}{0.21,0.21,0.21}
\definecolor{grey22}{rgb}{0.22,0.22,0.22}
\definecolor{grey23}{rgb}{0.23,0.23,0.23}
\definecolor{grey24}{rgb}{0.24,0.24,0.24}
\definecolor{grey25}{rgb}{0.25,0.25,0.25}
\definecolor{grey26}{rgb}{0.26,0.26,0.26}
\definecolor{grey27}{rgb}{0.27,0.27,0.27}
\definecolor{grey28}{rgb}{0.28,0.28,0.28}
\definecolor{grey29}{rgb}{0.29,0.29,0.29}
\definecolor{grey2}{rgb}{0.02,0.02,0.02}
\definecolor{grey30}{rgb}{0.30,0.30,0.30}
\definecolor{grey31}{rgb}{0.31,0.31,0.31}
\definecolor{grey32}{rgb}{0.32,0.32,0.32}
\definecolor{grey33}{rgb}{0.33,0.33,0.33}
\definecolor{grey34}{rgb}{0.34,0.34,0.34}
\definecolor{grey35}{rgb}{0.35,0.35,0.35}
\definecolor{grey36}{rgb}{0.36,0.36,0.36}
\definecolor{grey37}{rgb}{0.37,0.37,0.37}
\definecolor{grey38}{rgb}{0.38,0.38,0.38}
\definecolor{grey39}{rgb}{0.39,0.39,0.39}
\definecolor{grey3}{rgb}{0.03,0.03,0.03}
\definecolor{grey40}{rgb}{0.40,0.40,0.40}
\definecolor{grey41}{rgb}{0.41,0.41,0.41}
\definecolor{grey42}{rgb}{0.42,0.42,0.42}
\definecolor{grey43}{rgb}{0.43,0.43,0.43}
\definecolor{grey44}{rgb}{0.44,0.44,0.44}
\definecolor{grey45}{rgb}{0.45,0.45,0.45}
\definecolor{grey46}{rgb}{0.46,0.46,0.46}
\definecolor{grey47}{rgb}{0.47,0.47,0.47}
\definecolor{grey48}{rgb}{0.48,0.48,0.48}
\definecolor{grey49}{rgb}{0.49,0.49,0.49}
\definecolor{grey4}{rgb}{0.04,0.04,0.04}
\definecolor{grey50}{rgb}{0.50,0.50,0.50}
\definecolor{grey51}{rgb}{0.51,0.51,0.51}
\definecolor{grey52}{rgb}{0.52,0.52,0.52}
\definecolor{grey53}{rgb}{0.53,0.53,0.53}
\definecolor{grey54}{rgb}{0.54,0.54,0.54}
\definecolor{grey55}{rgb}{0.55,0.55,0.55}
\definecolor{grey56}{rgb}{0.56,0.56,0.56}
\definecolor{grey57}{rgb}{0.57,0.57,0.57}
\definecolor{grey58}{rgb}{0.58,0.58,0.58}
\definecolor{grey59}{rgb}{0.59,0.59,0.59}
\definecolor{grey5}{rgb}{0.05,0.05,0.05}
\definecolor{grey60}{rgb}{0.60,0.60,0.60}
\definecolor{grey61}{rgb}{0.61,0.61,0.61}
\definecolor{grey62}{rgb}{0.62,0.62,0.62}
\definecolor{grey63}{rgb}{0.63,0.63,0.63}
\definecolor{grey64}{rgb}{0.64,0.64,0.64}
\definecolor{grey65}{rgb}{0.65,0.65,0.65}
\definecolor{grey66}{rgb}{0.66,0.66,0.66}
\definecolor{grey67}{rgb}{0.67,0.67,0.67}
\definecolor{grey68}{rgb}{0.68,0.68,0.68}
\definecolor{grey69}{rgb}{0.69,0.69,0.69}
\definecolor{grey6}{rgb}{0.06,0.06,0.06}
\definecolor{grey70}{rgb}{0.70,0.70,0.70}
\definecolor{grey71}{rgb}{0.71,0.71,0.71}
\definecolor{grey72}{rgb}{0.72,0.72,0.72}
\definecolor{grey73}{rgb}{0.73,0.73,0.73}
\definecolor{grey74}{rgb}{0.74,0.74,0.74}
\definecolor{grey75}{rgb}{0.75,0.75,0.75}
\definecolor{grey76}{rgb}{0.76,0.76,0.76}
\definecolor{grey77}{rgb}{0.77,0.77,0.77}
\definecolor{grey78}{rgb}{0.78,0.78,0.78}
\definecolor{grey79}{rgb}{0.79,0.79,0.79}
\definecolor{grey7}{rgb}{0.07,0.07,0.07}
\definecolor{grey80}{rgb}{0.80,0.80,0.80}
\definecolor{grey81}{rgb}{0.81,0.81,0.81}
\definecolor{grey82}{rgb}{0.82,0.82,0.82}
\definecolor{grey83}{rgb}{0.83,0.83,0.83}
\definecolor{grey84}{rgb}{0.84,0.84,0.84}
\definecolor{grey85}{rgb}{0.85,0.85,0.85}
\definecolor{grey86}{rgb}{0.86,0.86,0.86}
\definecolor{grey87}{rgb}{0.87,0.87,0.87}
\definecolor{grey88}{rgb}{0.88,0.88,0.88}
\definecolor{grey89}{rgb}{0.89,0.89,0.89}
\definecolor{grey8}{rgb}{0.08,0.08,0.08}
\definecolor{grey90}{rgb}{0.90,0.90,0.90}
\definecolor{grey91}{rgb}{0.91,0.91,0.91}
\definecolor{grey92}{rgb}{0.92,0.92,0.92}
\definecolor{grey93}{rgb}{0.93,0.93,0.93}
\definecolor{grey94}{rgb}{0.94,0.94,0.94}
\definecolor{grey95}{rgb}{0.95,0.95,0.95}
\definecolor{grey96}{rgb}{0.96,0.96,0.96}
\definecolor{grey97}{rgb}{0.97,0.97,0.97}
\definecolor{grey98}{rgb}{0.98,0.98,0.98}
\definecolor{grey99}{rgb}{0.99,0.99,0.99}
\definecolor{grey9}{rgb}{0.09,0.09,0.09}
\definecolor{grey}{rgb}{0.75,0.75,0.75}
\definecolor{honeydew1}{rgb}{0.94,1.00,0.94}
\definecolor{honeydew2}{rgb}{0.88,0.93,0.88}
\definecolor{honeydew3}{rgb}{0.76,0.80,0.76}
\definecolor{honeydew4}{rgb}{0.51,0.55,0.51}
\definecolor{honeydew}{rgb}{0.94,1.00,0.94}
\definecolor{hotpink}{rgb}{1.00,0.41,0.71}
\definecolor{indianred}{rgb}{0.80,0.36,0.36}
\definecolor{ivory1}{rgb}{1.00,1.00,0.94}
\definecolor{ivory2}{rgb}{0.93,0.93,0.88}
\definecolor{ivory3}{rgb}{0.80,0.80,0.76}
\definecolor{ivory4}{rgb}{0.55,0.55,0.51}
\definecolor{ivory}{rgb}{1.00,1.00,0.94}
\definecolor{khaki1}{rgb}{1.00,0.96,0.56}
\definecolor{khaki2}{rgb}{0.93,0.90,0.52}
\definecolor{khaki3}{rgb}{0.80,0.78,0.45}
\definecolor{khaki4}{rgb}{0.55,0.53,0.31}
\definecolor{khaki}{rgb}{0.94,0.90,0.55}
\definecolor{lavenderblush}{rgb}{1.00,0.94,0.96}
\definecolor{lavender}{rgb}{0.90,0.90,0.98}
\definecolor{lawngreen}{rgb}{0.49,0.99,0.00}
\definecolor{lemonchiffon}{rgb}{1.00,0.98,0.80}
\definecolor{lightblue}{rgb}{0.68,0.85,0.90}
\definecolor{lightcoral}{rgb}{0.94,0.50,0.50}
\definecolor{lightcyan}{rgb}{0.88,1.00,1.00}
\definecolor{lightgoldenrod}{rgb}{0.93,0.87,0.51}
\definecolor{lightgoldenrod}{rgb}{0.98,0.98,0.82}
\definecolor{lightgray}{rgb}{0.83,0.83,0.83}
\definecolor{lightgreen}{rgb}{0.56,0.93,0.56}
\definecolor{lightgrey}{rgb}{0.83,0.83,0.83}
\definecolor{lightpink}{rgb}{1.00,0.71,0.76}
\definecolor{lightsalmon}{rgb}{1.00,0.63,0.48}
\definecolor{lightsea}{rgb}{0.13,0.70,0.67}
\definecolor{lightsky}{rgb}{0.53,0.81,0.98}
\definecolor{lightslate}{rgb}{0.47,0.53,0.60}
\definecolor{lightslate}{rgb}{0.47,0.53,0.60}
\definecolor{lightslate}{rgb}{0.52,0.44,1.00}
\definecolor{lightsteel}{rgb}{0.69,0.77,0.87}
\definecolor{lightyellow}{rgb}{1.00,1.00,0.88}
\definecolor{limegreen}{rgb}{0.20,0.80,0.20}
\definecolor{linen}{rgb}{0.98,0.94,0.90}
\definecolor{magenta1}{rgb}{1.00,0.00,1.00}
\definecolor{magenta2}{rgb}{0.93,0.00,0.93}
\definecolor{magenta3}{rgb}{0.80,0.00,0.80}
\definecolor{magenta4}{rgb}{0.55,0.00,0.55}
\definecolor{magenta}{rgb}{1.00,0.00,1.00}
\definecolor{maroon1}{rgb}{1.00,0.20,0.70}
\definecolor{maroon2}{rgb}{0.93,0.19,0.65}
\definecolor{maroon3}{rgb}{0.80,0.16,0.56}
\definecolor{maroon4}{rgb}{0.55,0.11,0.38}
\definecolor{maroon}{rgb}{0.69,0.19,0.38}
\definecolor{mediumaquamarine}{rgb}{0.40,0.80,0.67}
\definecolor{mediumblue}{rgb}{0.00,0.00,0.80}
\definecolor{mediumorchid}{rgb}{0.73,0.33,0.83}
\definecolor{mediumpurple}{rgb}{0.58,0.44,0.86}
\definecolor{mediumsea}{rgb}{0.24,0.70,0.44}
\definecolor{mediumslate}{rgb}{0.48,0.41,0.93}
\definecolor{mediumspring}{rgb}{0.00,0.98,0.60}
\definecolor{mediumturquoise}{rgb}{0.28,0.82,0.80}
\definecolor{mediumviolet}{rgb}{0.78,0.08,0.52}
\definecolor{midnightblue}{rgb}{0.10,0.10,0.44}
\definecolor{mintcream}{rgb}{0.96,1.00,0.98}
\definecolor{mistyrose}{rgb}{1.00,0.89,0.88}
\definecolor{moccasin}{rgb}{1.00,0.89,0.71}
\definecolor{navajowhite}{rgb}{1.00,0.87,0.68}
\definecolor{navyblue}{rgb}{0.00,0.00,0.50}
\definecolor{navy}{rgb}{0.00,0.00,0.50}
\definecolor{oldlace}{rgb}{0.99,0.96,0.90}
\definecolor{olivedrab}{rgb}{0.42,0.56,0.14}
\definecolor{orange1}{rgb}{1.00,0.65,0.00}
\definecolor{orange2}{rgb}{0.93,0.60,0.00}
\definecolor{orange3}{rgb}{0.80,0.52,0.00}
\definecolor{orange4}{rgb}{0.55,0.35,0.00}
\definecolor{orangered}{rgb}{1.00,0.27,0.00}
\definecolor{orange}{rgb}{1.00,0.65,0.00}
\definecolor{orchid1}{rgb}{1.00,0.51,0.98}
\definecolor{orchid2}{rgb}{0.93,0.48,0.91}
\definecolor{orchid3}{rgb}{0.80,0.41,0.79}
\definecolor{orchid4}{rgb}{0.55,0.28,0.54}
\definecolor{orchid}{rgb}{0.85,0.44,0.84}
\definecolor{palegoldenrod}{rgb}{0.93,0.91,0.67}
\definecolor{palegreen}{rgb}{0.60,0.98,0.60}
\definecolor{paleturquoise}{rgb}{0.69,0.93,0.93}
\definecolor{paleviolet}{rgb}{0.86,0.44,0.58}
\definecolor{papayawhip}{rgb}{1.00,0.94,0.84}
\definecolor{peachpuff}{rgb}{1.00,0.85,0.73}
\definecolor{peru}{rgb}{0.80,0.52,0.25}
\definecolor{pink1}{rgb}{1.00,0.71,0.77}
\definecolor{pink2}{rgb}{0.93,0.66,0.72}
\definecolor{pink3}{rgb}{0.80,0.57,0.62}
\definecolor{pink4}{rgb}{0.55,0.39,0.42}
\definecolor{pink}{rgb}{1.00,0.75,0.80}
\definecolor{plum1}{rgb}{1.00,0.73,1.00}
\definecolor{plum2}{rgb}{0.93,0.68,0.93}
\definecolor{plum3}{rgb}{0.80,0.59,0.80}
\definecolor{plum4}{rgb}{0.55,0.40,0.55}
\definecolor{plum}{rgb}{0.87,0.63,0.87}
\definecolor{powderblue}{rgb}{0.69,0.88,0.90}
\definecolor{purple1}{rgb}{0.61,0.19,1.00}
\definecolor{purple2}{rgb}{0.57,0.17,0.93}
\definecolor{purple3}{rgb}{0.49,0.15,0.80}
\definecolor{purple4}{rgb}{0.33,0.10,0.55}
\definecolor{purple}{rgb}{0.63,0.13,0.94}
\definecolor{red1}{rgb}{1.00,0.00,0.00}
\definecolor{red2}{rgb}{0.93,0.00,0.00}
\definecolor{red3}{rgb}{0.80,0.00,0.00}
\definecolor{red4}{rgb}{0.55,0.00,0.00}
\definecolor{red}{rgb}{1.00,0.00,0.00}
\definecolor{rosybrown}{rgb}{0.74,0.56,0.56}
\definecolor{royalblue}{rgb}{0.25,0.41,0.88}
\definecolor{saddlebrown}{rgb}{0.55,0.27,0.07}
\definecolor{salmon1}{rgb}{1.00,0.55,0.41}
\definecolor{salmon2}{rgb}{0.93,0.51,0.38}
\definecolor{salmon3}{rgb}{0.80,0.44,0.33}
\definecolor{salmon4}{rgb}{0.55,0.30,0.22}
\definecolor{salmon}{rgb}{0.98,0.50,0.45}
\definecolor{sandybrown}{rgb}{0.96,0.64,0.38}
\definecolor{seagreen}{rgb}{0.18,0.55,0.34}
\definecolor{seashell1}{rgb}{1.00,0.96,0.93}
\definecolor{seashell2}{rgb}{0.93,0.90,0.87}
\definecolor{seashell3}{rgb}{0.80,0.77,0.75}
\definecolor{seashell4}{rgb}{0.55,0.53,0.51}
\definecolor{seashell}{rgb}{1.00,0.96,0.93}
\definecolor{sienna1}{rgb}{1.00,0.51,0.28}
\definecolor{sienna2}{rgb}{0.93,0.47,0.26}
\definecolor{sienna3}{rgb}{0.80,0.41,0.22}
\definecolor{sienna4}{rgb}{0.55,0.28,0.15}
\definecolor{sienna}{rgb}{0.63,0.32,0.18}
\definecolor{skyblue}{rgb}{0.53,0.81,0.92}
\definecolor{slateblue}{rgb}{0.42,0.35,0.80}
\definecolor{slategray}{rgb}{0.44,0.50,0.56}
\definecolor{slategrey}{rgb}{0.44,0.50,0.56}
\definecolor{snow1}{rgb}{1.00,0.98,0.98}
\definecolor{snow2}{rgb}{0.93,0.91,0.91}
\definecolor{snow3}{rgb}{0.80,0.79,0.79}
\definecolor{snow4}{rgb}{0.55,0.54,0.54}
\definecolor{snow}{rgb}{1.00,0.98,0.98}
\definecolor{springgreen}{rgb}{0.00,1.00,0.50}
\definecolor{steelblue}{rgb}{0.27,0.51,0.71}
\definecolor{tan1}{rgb}{1.00,0.65,0.31}
\definecolor{tan2}{rgb}{0.93,0.60,0.29}
\definecolor{tan3}{rgb}{0.80,0.52,0.25}
\definecolor{tan4}{rgb}{0.55,0.35,0.17}
\definecolor{tan}{rgb}{0.82,0.71,0.55}
\definecolor{thistle1}{rgb}{1.00,0.88,1.00}
\definecolor{thistle2}{rgb}{0.93,0.82,0.93}
\definecolor{thistle3}{rgb}{0.80,0.71,0.80}
\definecolor{thistle4}{rgb}{0.55,0.48,0.55}
\definecolor{thistle}{rgb}{0.85,0.75,0.85}
\definecolor{tomato1}{rgb}{1.00,0.39,0.28}
\definecolor{tomato2}{rgb}{0.93,0.36,0.26}
\definecolor{tomato3}{rgb}{0.80,0.31,0.22}
\definecolor{tomato4}{rgb}{0.55,0.21,0.15}
\definecolor{tomato}{rgb}{1.00,0.39,0.28}
\definecolor{turquoise1}{rgb}{0.00,0.96,1.00}
\definecolor{turquoise2}{rgb}{0.00,0.90,0.93}
\definecolor{turquoise3}{rgb}{0.00,0.77,0.80}
\definecolor{turquoise4}{rgb}{0.00,0.53,0.55}
\definecolor{turquoise}{rgb}{0.25,0.88,0.82}
\definecolor{violetred}{rgb}{0.82,0.13,0.56}
\definecolor{violet}{rgb}{0.93,0.51,0.93}
\definecolor{wheat1}{rgb}{1.00,0.91,0.73}
\definecolor{wheat2}{rgb}{0.93,0.85,0.68}
\definecolor{wheat3}{rgb}{0.80,0.73,0.59}
\definecolor{wheat4}{rgb}{0.55,0.49,0.40}
\definecolor{wheat}{rgb}{0.96,0.87,0.70}
\definecolor{whitesmoke}{rgb}{0.96,0.96,0.96}
\definecolor{white}{rgb}{1.00,1.00,1.00}
\definecolor{yellow1}{rgb}{1.00,1.00,0.00}
\definecolor{yellow2}{rgb}{0.93,0.93,0.00}
\definecolor{yellow3}{rgb}{0.80,0.80,0.00}
\definecolor{yellow4}{rgb}{0.55,0.55,0.00}
\definecolor{yellowgreen}{rgb}{0.60,0.80,0.20}
\definecolor{yellow}{rgb}{1.00,1.00,0.00}